\pdfoutput=1
\documentclass[journal,transmag,twocolumn]{IEEEtran}
\usepackage{float}
\usepackage{makecell}

\usepackage{amsmath}
\usepackage{array}
\usepackage{graphicx}
\usepackage{booktabs}
\usepackage{colortbl}
\usepackage{multirow}
\usepackage{multicol}
\usepackage{array}
\usepackage{cite}
\usepackage{amsfonts,amssymb} 
\usepackage[colorlinks,linkcolor=red,anchorcolor=blue,citecolor=green]{hyperref}

\usepackage{graphicx}
\usepackage{subfig}
\usepackage{stfloats}
\usepackage{lipsum}
\usepackage{algorithm}
\usepackage{algorithmic}
\usepackage{url}

\ifCLASSINFOpdf
\else
\fi
\begin{document}
%

\title{Fast and Reliable Probabilistic Face Embeddings in the Wild}



\author{\IEEEauthorblockN{Kai~Chen\IEEEauthorrefmark{1},
Qi~Lv\IEEEauthorrefmark{2},Taihe~Yi\IEEEauthorrefmark{1}}
\IEEEauthorblockA{\IEEEauthorrefmark{1}College of Systems Engineering, National University of Defense Technology, Changsha, China}
\IEEEauthorblockA{\IEEEauthorrefmark{2}College of Meteorology and Oceanography, National University of Defense Technology, Changsha, China}
\thanks{Corresponding author: Q. Lv (email: lvqi@nudt.edu.cn).}}

\markboth{manuscript}%
{Shell \MakeLowercase{\textit{et al.}}: Bare Demo of IEEEtran.cls for IEEE Transactions on Magnetics Journals}
%



\IEEEtitleabstractindextext{%
\begin{abstract}
Probabilistic Face Embeddings (PFE) can improve face recognition performance in unconstrained scenarios by integrating data uncertainty into the feature representation.
However, existing PFE methods tend to be over-confident in estimating uncertainty and is too slow to apply to large-scale face matching.
This paper proposes a regularized probabilistic face embedding method to improve the robustness and speed of PFE.
Specifically, the mutual likelihood score (MLS) metric used in PFE is simplified to speedup the matching of face feature pairs. Then, an output-constraint loss is proposed to penalize the variance of the uncertainty output, which can regularize the output of the neural network.
In addition, an identification preserving loss is proposed to improve the discriminative of the MLS metric, and a multi-layer feature fusion module is proposed to improve the neural network's uncertainty estimation ability.
Comprehensive experiments show that the proposed method can achieve comparable or better results in 9 benchmarks than the state-of-the-art methods, and can improve the performance of risk-controlled face recognition.
The code of our work is publicly available in GitHub (\url{https://github.com/KaenChan/ProbFace}).

\end{abstract}

}

\maketitle

\IEEEdisplaynontitleabstractindextext

%
\IEEEpeerreviewmaketitle

\section{Introduction}

Face recognition is a classic computer vision task.
Compared with traditional face recognition algorithms, the performance of deep learning-based algorithms has been greatly boosted. The face recognition accuracy of the algorithm on the LFW dataset \cite{huang2008labeled} has exceeded the manual comparison accuracy (99.20\%) \cite{kumar2009attribute}.
This is mainly due to the development of following four aspects, including: 1) large-scale datasets, e.g. CASIA Webface \cite{yi2014learning}, VGGFace \cite{cao2018vggface2}, MS1M \cite{guo2016ms}, Glint360K \cite{an2020partial}, etc.; 2) alignment-based face preprocessing \cite{yi2014learning,deng2019retinaface}; 3) powerful backbone network, e.g. VGGNet \cite{simonyan2014very}, ResNet \cite{he2016deep}, SENet \cite{hu2018squeeze}, etc.; 4) suitable objective function, e.g. contrastive loss \cite{sun2015deepid3}, triplet loss  \cite{schroff2015facenet}, large-margin softmax \cite{wang2018cosface,wang2018additive,deng2019arcface}, etc.

However, in actual unconstrained scenes, there are still many challenges for the application of face recognition system \cite{masi2018deep}.
One of the most important challenges is that the quality of the input face image can affect the accuracy and robustness of the face recognition system.
The quality of a face image can be influenced by various factors such as illumination, age, pose, expression, occlusion, and motion blur et al. This makes it difficult to further improve the accuracy of face recognition, especially for the risk-sensitive systems, such as facial payment and facial access control systems.
To solve this problem, some related studies have been optimized from different aspects, such as face alignment with more accurate key-points \cite{deng2019retinaface}, frontal face generation \cite{tran2017disentangled,yin2017towards, qian2019unsupervised}, face image quality control \cite{hernandez2019faceqnet,terhorst2020ser}, and robust face feature extraction \cite{cao2018pose,masi2016pose,le2019illumination}.

Probabilistic face embeddings (PFE) \cite{shi2019probabilistic} can be seen as a method of combining quality control and feature representation optimization.
The face feature is defined as a Gaussian distribution in PFE, where the mean is the face feature and the variance is used to estimate the data uncertainty of the input face image.
Generally, uncertainty estimation is a very important part of the forecasting system. It can provide support for subsequent decision-making by estimating the uncertainty of the predicted results. For face recognition task, uncertainty estimation can be used to prevent mis-recognition of low-quality images or out-of-distribution (OOD) images.
There are some works that extend the PFE method.
DUL \cite{chang2020data} improves the robustness of the model by learning feature representation and estimating the uncertainty at the same time.
Shi et al. \cite{shi2020towards} proposes a probabilistic feature representation of multiple subspaces, which improves the model's robustness and interpretability.
However, there are still some problems in these methods:
1) there is no constraints of the uncertainty output, which makes the output range of the uncertainty estimation too broad and easy to over-fit.
2) the mutual likelihood score (MLS) metric is used to calculate the similarity score between two probabilistic features, which increases the amount of calculation for feature comparison.

\begin{figure}[tb]
	\centering
	\subfloat[]{%
		 \includegraphics[width=.23\textwidth]{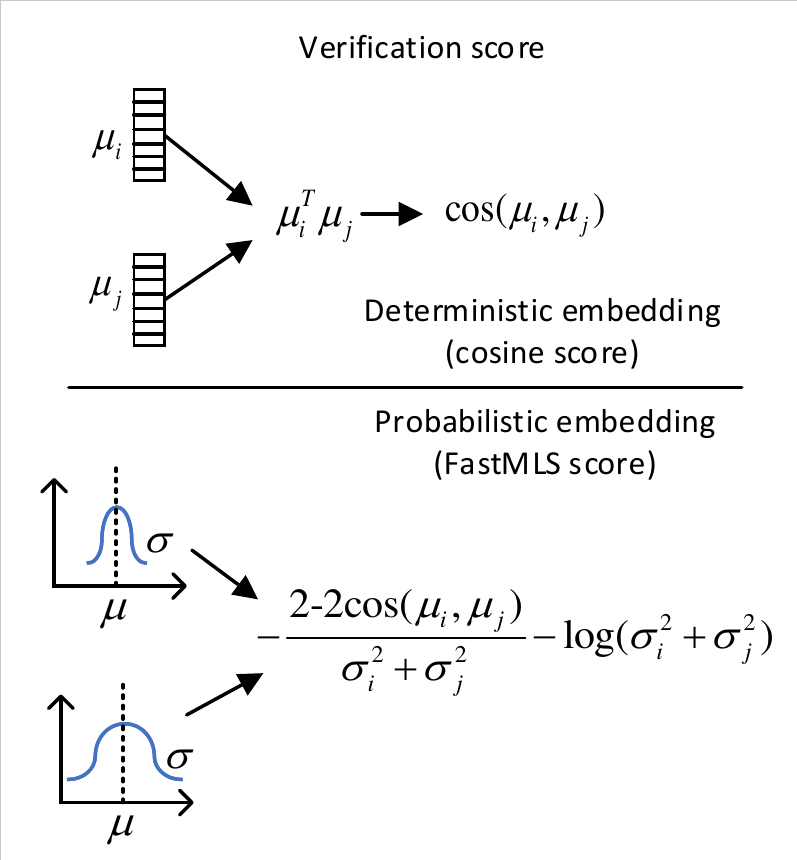}}\hfill
	\subfloat[]{%
		 \includegraphics[width=.23\textwidth]{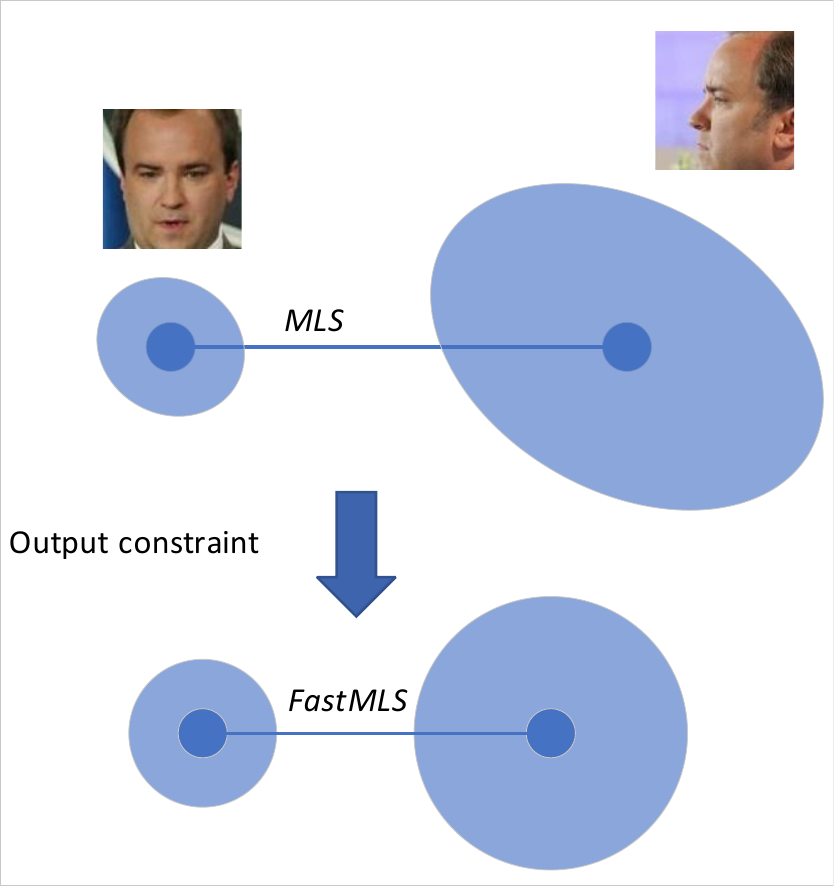}}\hfill
	\caption{~Demonstrations of our approach. (a) The FastMLS metric can be regarded as an uncertainty-based adjustment of the cosine metric. (b) In order to increase the robustness of MLS, the uncertainty is constrained to reduce the range of variance in FastMLS. }
	\label{fig:prob-example}
\end{figure}

In order to solve these problems, this paper proposes a robust probabilistic face embeddings method (ProbFace) to improve the recognition performance in the unconstrained environment.
To solve the problem that the range of the uncertain output is too large, a constraint term is added to penalize the variance of the uncertainty output, as shown in Figure \ref{fig:prob-example}(a). This is similar to the idea of confidence penalty in classification \cite{pereyra2017regularizing}, both of which constrain the output of deep neural networks.
Only positive sample pairs are considered in the training process of PFE, and negative sample pairs are not considered. To this end, we propose an uncertainty-aware loss function to preserve the identity information. The loss function can use the information of positive pairs and negative pairs at the same time, thus improving the discriminative of MLS metric.
For the time-consuming problem of MLS metric calculation, we reduce the output of uncertainty from $D$ dimension to 1 dimension, where $D$ is the length of face feature. The final calculation process is equivalent to an adjustment of cosine score, which can be calculated efficiently, as shown in Figure \ref{fig:prob-example}(b).
Finally, we use multi-scale feature fusion to utilize both low-level and high-level features to enhance the ability of uncertain predictions.
The contributions of the paper can be summarized as below:
\begin{itemize}
	\item Simplify the calculation of the MLS metric to an uncertainty-based adjustment of the cosine metric to speed up the calculation.
	\item A regularization term is proposed for uncertainty output to reduce the prediction range of uncertainty to prevent over-fitting.
	\item An identification preserving loss is proposed to improve the discriminative of the MLS metric.
	\item Fuse the features of different layers in the network to improve the ability of uncertainty estimation.
	\item Comprehensive experiments showing that the proposed method can achieve comparable or better performance in 9 benchmarks than SOTA methods, and can improve the performance of risk-control face recognition.
\end{itemize}

\section{Background and Related Works}

\subsection{Face Recognition in the Wild}

With the development of deep learning algorithms, the accuracy of face recognition continues to increase.
In 2014, DeepFace \cite{taigman2014deepface} used three-dimensional normalized alignment processing and classification loss to achieve a face comparison accuracy of 97.35\% on the LFW dataset.
DeepID3 \cite{sun2015deepid3} uses the verification loss to optimize the network, and the accuracy on LFW is increased to 99.53\%.
In 2015, FaceNet \cite{schroff2015facenet} used the Triplet loss function to achieve 99.63\% on LFW.
Margin-based loss functions, including L-softmax \cite{liu2016large}, A-softmax \cite{liu2017sphereface}, AMSoftmax \cite{wang2018additive} and ArcFace \cite{deng2019arcface}, etc., can recently reach 99.83\% on LFW.
These losses can enforce extra intra-class compactness and inter-class discrepancy simultaneously by importing angular/cosine margin into the softmax loss to enhance the model's discriminative strength.
From the findings above, it can be shown that the accuracy of the LFW dataset is already very high. But on larger datasets, the accuracy of these methods is still not high enough. For instance, on the Trillion Pairs \cite{trillionpairs} test set with 1.87 million face images, the current best recognition accuracy can only reach 89.80\%.

In unconstrained scenes \cite{masi2018deep}, the accuracy of face recognition will be affected by the quality of the face image.
To this end, three aspects can be optimized: image preprocessing, image quality control, and robust face feature extraction.
1) In image preprocessing, more accurate key-point detection algorithms can be used to get better aligned images, thereby improving the accuracy of face recognition \cite{deng2019retinaface}. Moreover, Generative adversarial network (GAN) can be used to generate clear frontal face from large poses or occluded face images with identity information preserved \cite{tran2017disentangled,yin2017towards,qian2019unsupervised}.
2) In the face image quality control methods, low-quality face images will be filtered out by the face quality prediction algorithms \cite{hernandez2019faceqnet,terhorst2020ser}.
3) In terms of feature extraction, the robustness of face recognition feature representation is improved through methods such as pose-invariant representation \cite{cao2018pose,masi2016pose}, illumination-invariant representation \cite{le2019illumination} and regularization \cite{xu2016new,zhao2019regularface}.

Therefore, the capabilities and accuracy of face recognition algorithms do need to be more enhanced as the recognition scale and scene complexity increase.

\subsection{Uncertainty in Deep Learning for Computer Vision}

There are two types of uncertainty in deep learning, model uncertainty and data uncertainty.
Model uncertainty captures the noise of the parameters in deep neural networks and can be reduced by increasing the size of the training data, which is also called epistemic uncertainty \cite{gal2016dropout}.
Data uncertainty captures the noise inherent in given training data and does not change with the increase of the amount of training data, which is also called aleatoric uncertainty \cite{kendall2017uncertainties}.

Uncertainty in deep learning can usually be estimated by Bayesian deep learning (BNNs) and ensemble learning methods \cite{abdar2020review}.
Because it is difficult to calculate exact posterior inferences for Bayesian deep learning, several approximate methods have been proposed, such as Monte Carlo (MC) dropout \cite{gal2015bayesian,nair2020exploring}, Markov chain Monte Carlo (MCMC) \cite{li2016learning}, Variational Inference (VI) \cite{posch2019variational}, etc.
The ensemble methods can estimate both model and data uncertainty by analyzing the diversity of each model's output \cite{lakshminarayanan2017simple, ashukha2020pitfalls}.

So far, there have been many studies on the uncertainty in deep learning for various computer vision applications.
\begin{itemize}
  \item \textbf{Semantic segmentation}. Monte Carlo (MC) dropout \cite{kendall2015bayesian} and Bayesian neural networks (BNNs) \cite{kendall2017uncertainties} are used to measure model uncertainty and data uncertainty to predict the uncertainty of pixel-wise class labels.
  \item \textbf{Human pose and localization}. Gundavarapu et al. \cite{gundavarapu2019structured} used data uncertainty to improve the robustness in human pose estimation. By considering the uncertainty of model and data at the same time, Bertoni et al. \cite{bertoni2019monoloco} addressed the challenges of the ill-posed problem of 3D human localization from monocular RGB images.
  \item \textbf{Object detection} To boost the detection efficiency, the uncertainty estimation method proposed in \cite{kendall2017uncertainties} was applied to the two-stage object detection network \cite{feng2018towards,wirges2019capturing} and the one-stage object detection network \cite{kraus2019uncertainty}.
  \item \textbf{Person Re-ID} Data uncertainty was used to minimize the negative impact of the noisy label and outlying samples in person Re-ID task \cite{yu2019robust}.
  \item \textbf{Face recognition} Model uncertainty can be used to analyze the capacity of the face representation \cite{gong2017capacity}, learn robust features \cite{khan2019striking}, and estimate the quality of face images \cite{terhorst2020ser}. For data uncertainty, probabilistic feature representation was proposed to improve the robustness and interpretability by representing each face image as one or more Gaussian distributions \cite{shi2019probabilistic,shi2020towards,chang2020data}. However, in these methods, there is no limit to the uncertainty estimation \cite{shi2019probabilistic,chang2020data} or only a simple limit \cite{shi2020towards}, which can easily lead to the problem of over-fitting.
\end{itemize}

\subsection{Probabilistic Face Embeddings and Mutual Likelihood Score} \label{sub:related_work_mls}
The probabilistic representation of data was introduced as early as 2014 for word embeddings \cite{Vilnis2015}, which can represent levels of specificity of word.
Then, it is extended to graph representation \cite{Bojchevski2017}, computer vision and other fields.
For computer vision, the probabilistic embeddings is used to improve performance and robustness in metric learning \cite{oh2018modeling}, pose estimation \cite{sun2020view}, prototype embeddings \cite{Scott2019} and face recognition \cite{shi2019probabilistic}.
Among them, the face probabilistic embeddings (PFE) \cite{shi2019probabilistic} is very similar to the word embedding \cite{Vilnis2015}. Both of them use Gaussian embeddings. And the mutual similarity score (MLS) of PFE is the same as the expected likelihood kernel (ELK) distributional distance used in word embedding \cite{Vilnis2015}.

According to \cite{shi2019probabilistic}, face feature can be defined as a Gaussian distribution:
$$p_D({\bf{z}}|{{\bf{x}}_i}) = {\cal N}({\bf{z}};{\mu _i},\sigma _i^2{\bf{I}})$$
where ${\bf{\mu }} \in {R^D}$, ${\bf{\sigma }} \in {R^D}$ and $D$ is the length of face embedding. ${\bf{\mu }}$ and ${\bf{\sigma }}$ represent the mean and variance of the face feature, both of which are the output of the neural network.
Assuming that ${\bf{z}}_i$ and ${\bf{z}}_j$ are two face distributions, mutual likelihood score can be used to measure the distance between ${\bf{z}}_i$ and ${\bf{z}}_j$, which is expressed as:
\begin{equation}\label{eq:mls-d}
\begin{aligned}
  S_D({{\bf{x}}_i},{{\bf{x}}_j}) = & - \frac{1}{2}\sum\limits_{l = 1}^D {\left( {\frac{{||\mu _i^{(l)} - \mu _j^{(l)}|{|^2}}}{{\sigma _i^{2(l)} + \sigma _j^{2(l)}}} + \log (\sigma _i^{2(l)} + \sigma _j^{2(l)})} \right)} \\
  & - const
\end{aligned}
\end{equation}
where ${const} = {\rm\frac{D}{2}{log 2\pi}}$.
The first term in the bracket in Equation (\ref{eq:mls-d}) can be regarded as a weighted distance, and the second term can be regarded as a penalty term. Only when the uncertainty of ${\bf{z}}_i$ and ${\bf{z}}_j$ are both small, can they get a higher $S_D$ score.
Therefore, $S_D$ is equivalent to fusing the face quality into the similarity score, which can reduce the similarity of low-quality face pairs, thus decreasing the recognition errors caused by low-quality images.

\section{Proposed Methods}

\begin{figure*}
	\centering
	\includegraphics[width=0.99\textwidth]{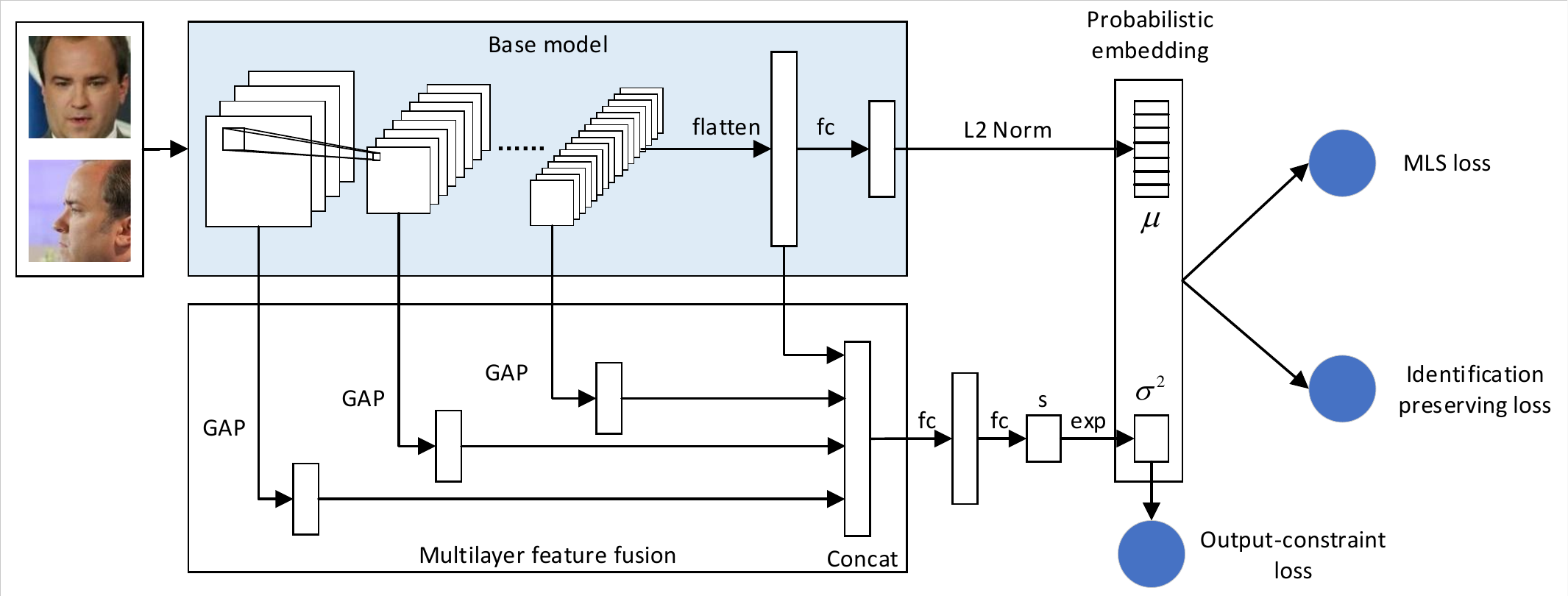}
	\caption{~Architecture of proposed ProbFace. The weights of the base model are fixed during training. Probability embedding is optimized by MLS loss and identification preserving loss, and the range of uncertainty $\sigma^2$ is regularized by output-constraint loss. Global average pooling (GAP) is used to extract the features of different layers to perform feature fusion. $\exp$ operation is performed on the output of the network to ensure that $\sigma^2$ is positive.}
	\label{fig:network-nn}
\end{figure*}

In this section, we introduce the proposed robust probabilistic face embedding method. In Section 3.1, we simplify the mutual likelihood score to speed up its calculation. In Section 3.2, we propose the output-constraint loss to penalize the variance of the predicted uncertainty. In Section 3.3, we propose identification preserving loss to optimize the discriminative of MLS metrics. Finally, in Section 3.4, we use the fusion of features at different layers to boost the ability of uncertainty estimation. The whole architecture is shown in Figure \ref{fig:network-nn}.

\subsection{Fast Mutual Likelihood Score}

From Section \ref{sub:related_work_mls}, it can be seen that each feature in PFE corresponds to a variance, leading to an increase in the size of face features storage. Furthermore, the calculation amount of the MLS metric $S_D$ is larger than the commonly used cosine metric.
Suppose that in the open-set identification task \cite{phillips2011evaluation}, the number of face images in gallery set is $N$ and the number of face images in probe set is $M$.
The number of comparisons between the gallery set and the probe set is $N\times M$.
From the perspective of computational complexity, the calculation amount of $S_D$ and cosine is not much different, both are $\Theta(N M D)$, and $S_D$ has one more division and one more log operation than the cosine metric.
However, the cosine metric can be quickly calculated using the optimized matrix multiplication library, such as OpenBLAS \cite{xianyi2012openblas}, etc. On the contrary, the calculation of $S_D$ cannot be written in the form of matrix multiplication, which makes it difficult to use the existing matrix calculation library. Therefore, the calculation speed of $S_D$ will be slow if the face database is relatively large.

In order to reduce the storage requirement and calculation time of $S_D$, we can change the output of uncertainty estimation from $D$ dimension to 1 dimension, so that the calculation of MLS can be converted to uncertainty-based adjustment of cosine metric. Specifically, the face features can be defined as a following Gaussian distribution:
\begin{equation}
  p_1({\bf{z}}|{{\bf{x}}_i}) = {\cal N}({\bf{z}};{\mu _i},\frac{{\sigma _i^2}}{D}{\bf{I}})
\end{equation}
where $\sigma \in R$. Similar to PFE, the ``likelihood" of ${\bf{z}}_{i}$ and ${\bf{z}}_{j}$ being the same person is as follows:
\begin{equation}
\begin{aligned}
  p_1&({{\bf{z}}_i} = {{\bf{z}}_j}|{{\rm{y}}_i} = {y_j}) = p_1(\Delta {{\bf{z}}_{ij}}= 0|{{\rm{y}}_i} = {y_j}) \\
  &= \frac{1}{{{{(2\pi (\sigma _i^2 + \sigma _j^2)/D)}^{D/2}}}}\exp \left( { - \frac{{D||{{\bf{\mu}} _i} - {{\bf{\mu}} _j}|{|^2}}}{{2(\sigma _i^2 + \sigma _j^2)}}} \right)
\end{aligned}
\end{equation}
where $y_i$ and $y_j$ are the identification labels of ${\bf{z}}_{i}$ and ${\bf{z}}_{j}$. Then, one-dimensional MLS can be obtained by the log likelihood:
\begin{equation}
\begin{aligned}
  S_1({{\bf{x}}_i},{{\bf{x}}_j}) &= \log p_1(\Delta {{\bf{z}}_{ij}}) \\
  &=  - \frac{D}{2}\left( {\frac{{||{{\bf{\mu}} _i} - {{\bf{\mu}} _j}|{|^2}}}{{\sigma _i^2 + \sigma _j^2}} + \log (\sigma _i^2 + \sigma _j^2)} \right) - const \\
  &= - \frac{D}{2}\left( {\frac{{2{\rm{ - 2cos(}}{\mu _i},{\mu _j})}}{{\sigma _i^2 + \sigma _j^2}} + \log (\sigma _i^2 + \sigma _j^2)} \right) - const
\end{aligned}
\end{equation}
Thus, the fast mutual likelihood score (FastMLS) can be obtained:
\begin{equation}\label{eq:mls1}
  S_{f}({{\bf{x}}_i},{{\bf{x}}_j}) \propto - \frac{{2 - {\rm{2cos}}({\mu _i},{\mu _j})}}{{\sigma _i^2 + \sigma _j^2}} - \log (\sigma _i^2 + \sigma _j^2)
\end{equation}
In Equation (\ref{eq:mls1}), the cosine score is scaled and punished by the sum of the uncertainties $\sigma_i$ and $\sigma_j$ of two inputs, so as to achieve the purpose of uncertainty-based cosine score correction.
If at least one of $\sigma_i$ and $\sigma_j$ is large, FastMLS score will be small.
Only when $\sigma_i$ and $\sigma_j$ are both small, can ${{\bf{x}}_i}$ and ${{\bf{x}}_j}$ get a higher FastMLS score.

As shown in the Figure \ref{fig:network-nn}, to ensure that $\sigma^2$ is greater than 0, we take the output of the network as $s\triangleq \log(\sigma^2)$ and we can get uncertainty by $\sigma^2=\exp(s)$.
The optimization goal of the network is to maximize the $S_{fast}$ of sample pairs with the same label. So we can get the loss function:
\begin{equation}\label{eq:loss-mls1}
  {L_{S}} = \frac{1}{N_{p}}\sum\limits_{i = 1}^M {\sum\limits_{j = i + 1}^M {{I_{{y_i} = {y_j}}}\left( {\frac{{{2 - {\rm{2cos}}({\mu _i},{\mu _j})}}}{{\sigma _i^2 + \sigma _j^2}} + \log (\sigma _i^2 + \sigma _j^2)} \right)} }
\end{equation}
where $M$ is the batch size, ${I_{( \cdot )}}$ is the indicator function and $N_{p}$ is the number of pairs satisfying $y_i=y_j$ in the mini-batch.

\subsection{Output-Constraint Loss}

PFE integrates face uncertainty into face feature, which can improve the robustness and accuracy of face recognition.
However, since the output of the neural network is easy to over-confident \cite{guo2017calibration}, the uncertainty of different images is likely to be too different, making the $\sigma$ value dominant in Equation (\ref{eq:mls1}).
When performing face recognition in the wild, the quality of the image changes greatly and the recognition performance based on MLS metric drops a lot.
Therefore, this paper constrains the range of the estimated uncertainty.
We can transform the problem into that the value of the uncertainty should not be too far from its average.
Suppose that $\sigma _{avg}^2$ is the average uncertainty of the all face images.
For the uncertainty $\sigma_i^2$ of input ${\bf{x}}_i$, we hope that the value of $\sigma _i^2/\sigma _{avg}^2$ is around 1.
By using the L1 regression loss, the output-constraint loss can be obtained:
\begin{equation}
  L_{C} = \frac{1}{M}\sum\limits_{i = 1}^M {|\frac{{\sigma _i^2}}{{\sigma _{avg}^2}} - 1|}
\end{equation}
where $M$ is the batch size. For the convenience of calculation, we use the average of $\sigma^2$ in a mini-batch to approximate the average of all data:
\begin{equation}
  \sigma _{avg}^2 = \frac{1}{M}\sum\limits_{i = 1}^M {\sigma _i^2}
\end{equation}

Constraining the output of neural networks is a commonly used regularization method, which has been used in many tasks.
In \cite{pereyra2017regularizing}, a maximum entropy based confidence penalty is used to regularize the output of large, deep neural networks on image classification, language modeling, etc. In fine-grained visual classification, pairwise confusion loss \cite{dubey2018pairwise} is used to bring class conditional probability distributions closer to each other.
Pairwise confusion loss is used to enhances the generalization of the learned representations for face anti-spoofing \cite{tu2020learning}.
Our proposed method constrains output of uncertainty to make the output variance more reasonable.

\subsection{Identification preserving loss}

In Equation (\ref{eq:loss-mls1}), only the influence of positive sample pairs on the uncertainty estimation is considered, and the influence of negative sample pairs is not considered.
The similarity of negative pairs should be smaller than the similarity of positive pairs. If there is a high score due to low quality, we can use data uncertainty to penalize it.
Therefore, we propose an identification preserving loss function that uses both positive sample pairs and negative sample pairs to improve the discriminative of MLS.
By referring to the triplet loss \cite{weinberger2009distance}, we get the uncertainty-aware triplet loss as:
\cite{weinberger2009distance}:
\begin{equation}
{L_{Id}} = \frac{1}{{|\mathcal{T}|}}\sum\limits_{(a,p,n) \in \mathcal{T}} {{{\left[ {\frac{{||{\mu _a} - {\mu _p}|{|^2}}}{{\sigma _a^2 + \sigma _p^2}} - \frac{{||{\mu _a} - {\mu _n}|{|^2}}}{{\sigma _a^2 + \sigma _n^2}} + m} \right]}_ + }}
\end{equation}
where $\mathcal{T}$ is triplet set in the mini-batch, $|\mathcal{T}|$ is the number of triplets and margin $m$ is set to 3.

The total loss can be written as follow:
\begin{equation}\label{eq:totalloss}
  L = {L_{S}} + {\lambda _{C}}{L_{C}} + {\lambda _{Id}}{L_{Id}}
\end{equation}
where $\lambda_{C}$ and $\lambda_{Id}$ are the weights of output-constraint loss and identification preserving loss respectively.

\subsection{Multi-layer Feature Fusion}
The uncertainty of a face image is influenced by many factors, including high-level semantic information (such as angle, occlusion) and low-level image details (such as blur, lighting).
In deep neural network, the low-level layers focus on local texture information and high-level layers can learn global semantic information.
As a consequence, fusing the features of multiple layers of the network can improve the ability of uncertainty estimation.
We use global average pooling (GAP) to transform feature maps of different level layers into vectors, and then concatenate them with the flattened vector of the last convolution, as illustrated in Figure \ref{fig:network-nn}.
Specifically,we use the ResNet network as the base model. Suppose $\mathbf{t}_{i}$ ($i=1,2,3,4$) is the output feature map of Conv1, Conv2, Conv3 and Conv4 modules in ResNet. By using GAP, the feature vector of each layer is obtained as $\mathbf{g}_{i}= Flatten(GAP(\mathbf{t}_{i}))$. The feature vector of the last layer of convolution is $\mathbf{g}_{last} = Flatten(\mathbf{t}_{i})$.
Finally, these feature vectors are concatenated together to get the fusion feature $\mathbf{G}=[\mathbf{g}_{1}, \mathbf{g}_{2}, \mathbf{g}_{3}, \mathbf{g}_{4 }, \mathbf{g}_{last}]$.

\section{Experiments}

In this section, we first introduce the datasets and implementation details. Then we conduct detailed ablation study over the proposed losses and modules.
We evaluate different types of test datasets and compared the results with state-of-the-art methods.
Further, we evaluate the performance of the proposed method on noisy data and risk-controlled scenario respectively.
We compare the running time of the original MLS and proposed fast MLS.
Finally, we provide the visualization results of the effect of uncertainty on the feature map.

\subsection{Datasets and Implementation Details}

We describe the public datasets used and our implementation details.

\begin{table}[htbp]
  \footnotesize
	\centering
	\caption{Description of the datasets}
	\begin{tabular}{clrrl}
		\hline
		& Datasets & \multicolumn{1}{l}{\#Identitiy} & \#Image &  \\
		\hline
		\multicolumn{1}{l}{Train} & MS-Celeb-1M-v2 \cite{deng2019arcface} & \multicolumn{1}{r}{85K} & 5.8M &  Variation\\
		\hline
		\multirow{8}[0]{*}{Test} & LFW \cite{huang2008labeled}  & 5,749 & \multicolumn{1}{r}{13,233} & limited \\
		& CFP-FF \cite{sengupta2016frontal} & 500   & \multicolumn{1}{r}{7,000} &  limited \\
		& CALFW \cite{huang2008labeled} & 5,749 & \multicolumn{1}{r}{12,174} & large-age\\
		& AgeDB30 \cite{moschoglou2017agedb} & 568   & \multicolumn{1}{r}{16,488} & large-age \\
		& CPLFW \cite{zheng2018cross} & 5,749 & \multicolumn{1}{r}{11,652} & large-pose \\
		& CFP-FP \cite{sengupta2016frontal} & 500   & \multicolumn{1}{r}{7,000} & large-pose \\
		& Vgg2FP \cite{cao2018vggface2} & 300   & \multicolumn{1}{r}{10,000} & large-pose \\
		& IJB-B \cite{whitelam2017iarpa} & 1,845 & 76.8K & full \\
		& IJB-C \cite{maze2018iarpa} & 3,531 & 148.8K & full \\
		\hline
	\end{tabular}%
	\label{tab:dataset}%
\end{table}%

\textbf{Datasets}. As shown in the Table \ref{tab:dataset}, we use MS-Celeb-1M-v2 dataset \cite{deng2019arcface} with 5.8 million
images of 85k subjects as training set, which is a clean version of the MS-Celeb-1M dataset \cite{guo2016ms}.
There are 8 testing sets, including 2 datasets with limited changes (LFW \cite{huang2008labeled} and CFP-FF \cite{sengupta2016frontal}), 2 large-age datasets (CALFW \cite{huang2008labeled}, AgeDB30 \cite{moschoglou2017agedb}), 3 large-pose datasets (CPLFW \cite{zheng2018cross}, CFP-FP \cite{sengupta2016frontal}, Vgg2FP \cite{cao2018vggface2}), and a large-scale image dataset with various variations (IJB-B \cite{whitelam2017iarpa}, IJB-C \cite{maze2018iarpa}).
Similar to \cite{deng2019arcface}, all training and test images are aligned by affine transformation according to the key points of the face, and resized to $96\times 96$.

\textbf{Implementation Details}. In the experiment, we used three networks as the base models, ResFace64(0.5), ResFace64 and ResFace100, as shown in the Table \ref{tab:resfacenetwork}. The smaller network ResFace64(0.5) has half the channel number of ResFace64.
The structure of the base model is similar to the SphereFace \cite{liu2017sphereface}. The difference is that we changed the activation function of the network from PRELU to RELU, and set the embedding size of the face feature to 256. This change can improve the training speed of the network.
The uncertainty module uses a two-layer fully connected neural network, where the input is the fusion of the multi-layer features of base mode, and the dimension of the middle layer is 128.
We first train base model using the ArcFace Loss \cite{deng2019arcface} on MS-Celeb-1M-v2 dataset.
Then the parameters of the base model is fixed and the uncertainty module is trained by SGD optimizer with momentum of 0.9 and weight decay of $5e-4$.
In the training process of uncertainty module, we set the batch size as 128 using a computer with one GTX1080Ti GPU.
In each batch, 8 persons are sampled, each with 16 pictures.
The learning rate start at 0.01, and then decreased to 0.001 and 0.0001 at 32K and 48K steps, and finish at 64K steps.

\begin{table}[htbp]
  \scriptsize
  \centering
  \caption{Network architecture of base model and uncertainty model.  S1 and S2 denote the stride of the convolutional layer is 1 and 2 respectively. The input of the uncertainty module is output of the multi-layer feature fusion module. The activation function used is RELU.}
    \begin{tabular}{cccc}
    \toprule
    \multicolumn{4}{c}{Face Embedding Network (Base Model)} \\
    \midrule
    Layer & Resface64(0.5) & Resface64 & Resface100 \\
    \midrule
    Input & 96x96x3 & 96x96x3 & 96x96x3 \\
    \midrule
    \multirow{2}[2]{*}{Conv1x} & \multicolumn{1}{l}{[3$\times $3,32]$\times $1,S2} & \multicolumn{1}{l}{[3$\times $3,64]$\times $1,S2} & \multicolumn{1}{l}{[3$\times $3,64]$\times $1,S2} \\
          & \multicolumn{1}{l}{[3$\times $3,32]$\times $3,S1} & \multicolumn{1}{l}{[3$\times $3,64]$\times $3,S1} & \multicolumn{1}{l}{[3$\times $3,64]$\times $3,S1} \\
    \midrule
    \multirow{2}[2]{*}{Conv2x} & \multicolumn{1}{l}{[3$\times $3,64]$\times $1,S2} & \multicolumn{1}{l}{[3$\times $3,128]$\times $1, S2} & \multicolumn{1}{l}{[3$\times $3,128]$\times $1,S2} \\
          & \multicolumn{1}{l}{[3$\times $3,64]$\times $8,S1} & \multicolumn{1}{l}{[3$\times $3,128]$\times $8,S1} & \multicolumn{1}{l}{[3$\times $3,128]$\times $10,S1} \\
    \midrule
    \multirow{2}[2]{*}{Conv3x} & \multicolumn{1}{l}{[3$\times $3,128]$\times $1,S2} & \multicolumn{1}{l}{[3$\times $3,256]$\times $1,S2} & \multicolumn{1}{l}{[3$\times $3,256]$\times $1,S2} \\
          & \multicolumn{1}{l}{[3$\times $3,128]$\times $16,S1} & \multicolumn{1}{l}{[3$\times $3,256]$\times $16,S1} & \multicolumn{1}{l}{[3$\times $3,256]$\times $32,S1} \\
    \midrule
    \multirow{2}[2]{*}{Conv4x} & \multicolumn{1}{l}{[3$\times $3,256]$\times $1,S2} & \multicolumn{1}{l}{[3$\times $3,512]$\times $1,S2} & \multicolumn{1}{l}{[3$\times $3,512]$\times $1,S2} \\
          & \multicolumn{1}{l}{[3$\times $3,256]$\times $3,S1} & \multicolumn{1}{l}{[3$\times $3,512]$\times $3,S1} & \multicolumn{1}{l}{[3$\times $3,512]$\times $3,S1} \\
    \midrule
    Flatten & 9216  & 18432 & 18432 \\
    \midrule
    FC    & 256   & 256   & 256 \\
    \midrule
    \#Params & 11.68 M & 41.94 M & 61.44 M \\
    FLOPs & 2.52 G & 10.46 G & 16.58 G \\
    \midrule
    \midrule
    \multicolumn{4}{c}{Uncertainty Network} \\
    \midrule
    Input & 9696  & 19392 & 19392 \\
    FC1   & 128   & 128   & 128 \\
    FC2   & 1     & 1     & 1 \\
    \midrule
    \#Params & 1.24 M & 2.48 M & 2.48 M \\
    FLOPs & 2 M   & 5 M   & 5 M \\
    \bottomrule
    \end{tabular}%
  \label{tab:resfacenetwork}%
\end{table}%

\subsection{Ablation experiment}

\subsubsection{Effect of Output-constraint Loss and Identification Preserving Loss}

\begin{figure}[htbp]
	\centering
	\subfloat[LFW and CPLFW]{%
		 \includegraphics[width=.23\textwidth]{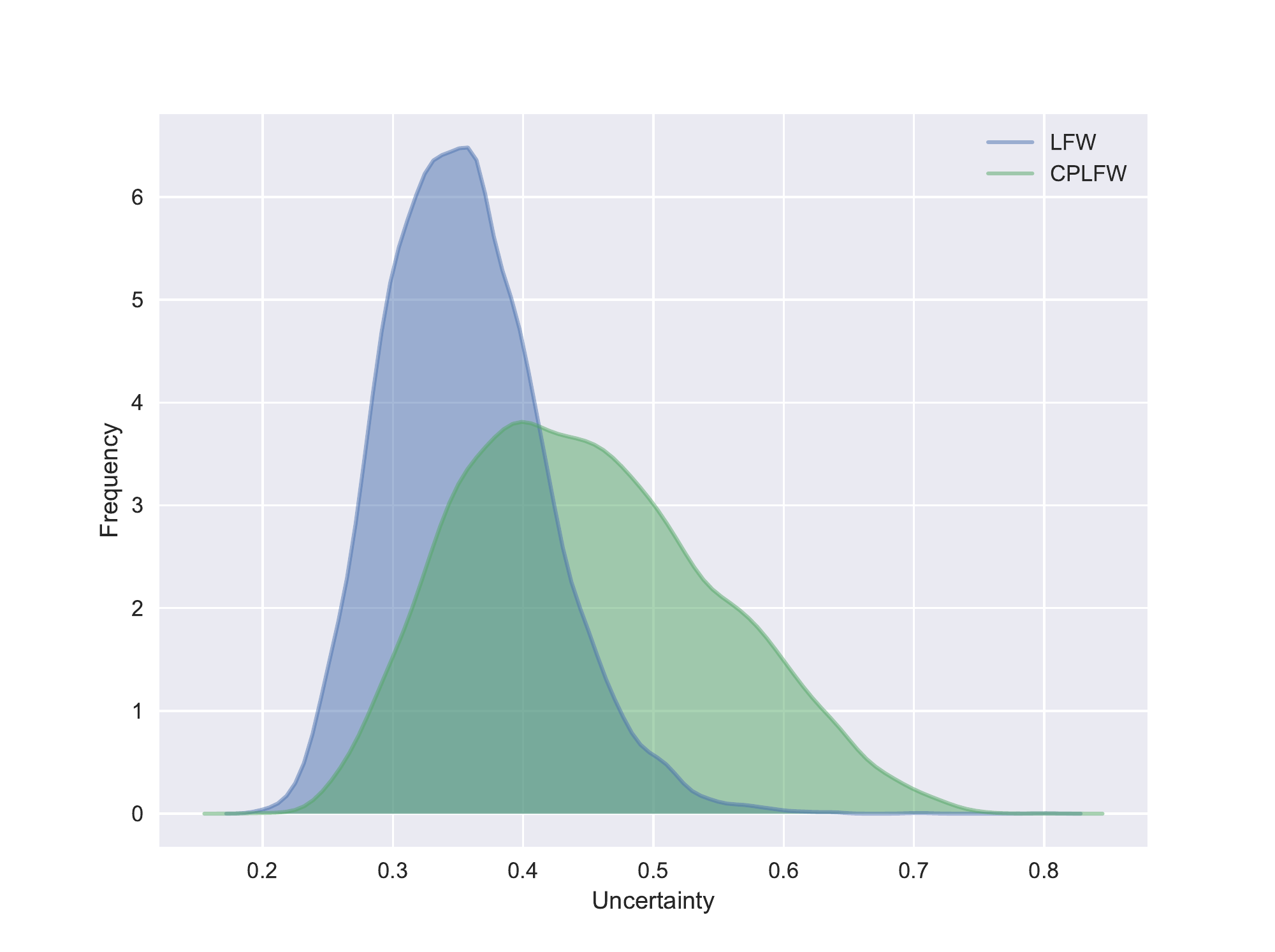}}\hfill
	\subfloat[CFP-FF and CFP-FP]{%
		 \includegraphics[width=.23\textwidth]{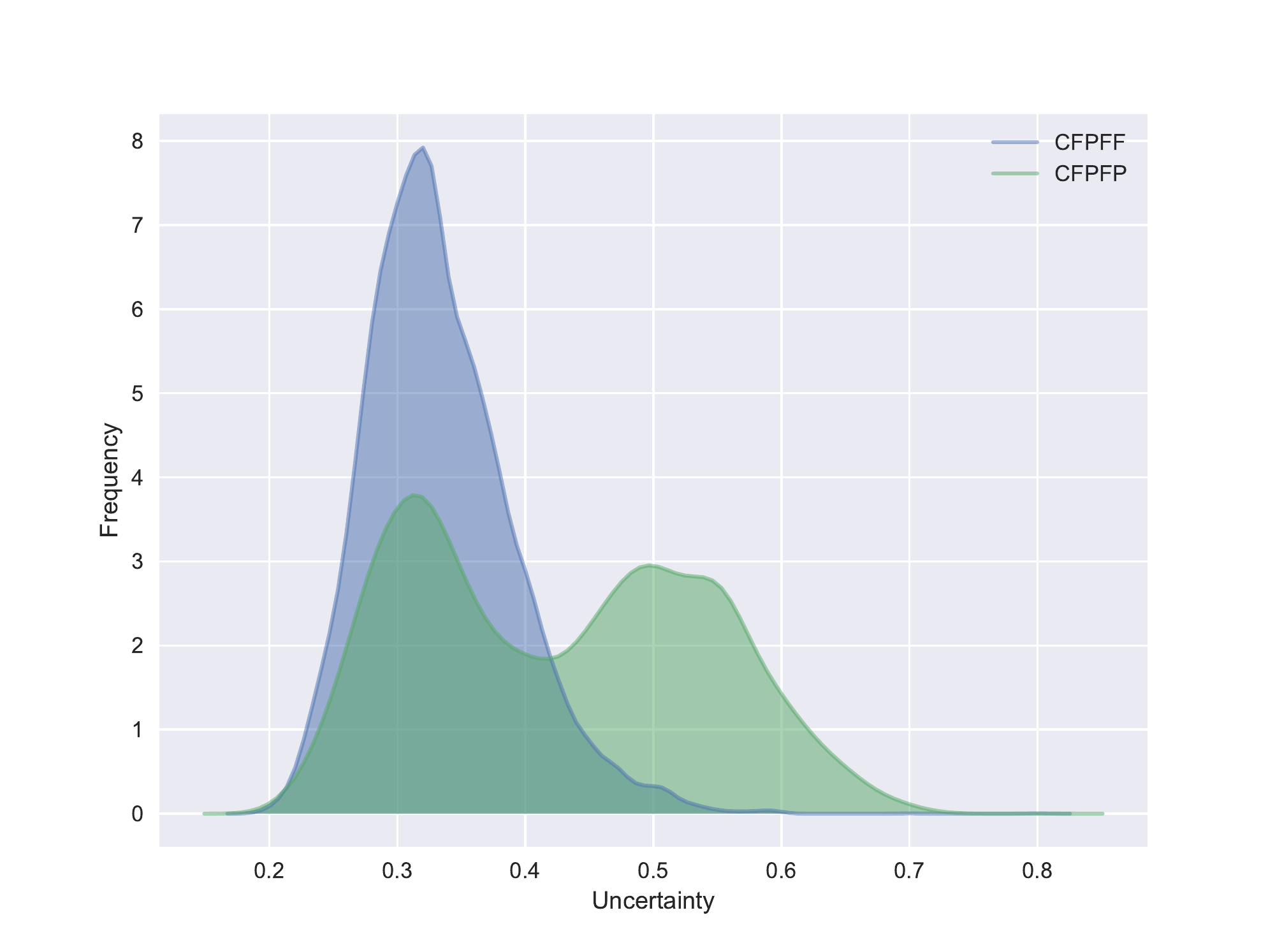}}\hfill
	\subfloat[CALFW and AgeDB30]{%
		 \includegraphics[width=.23\textwidth]{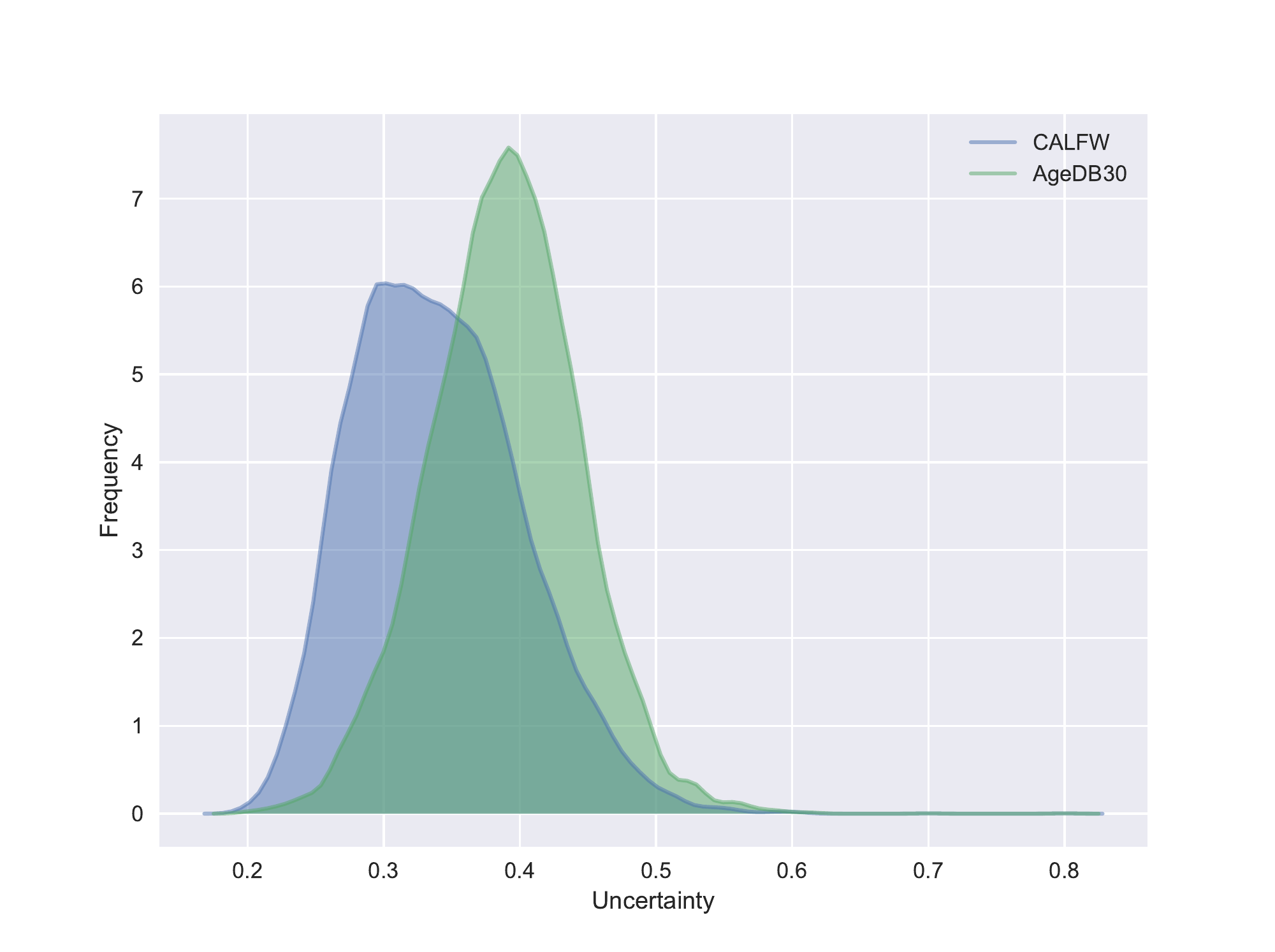}}\hfill
	\subfloat[IJB-B and IJB-C]{%
		 \includegraphics[width=.23\textwidth]{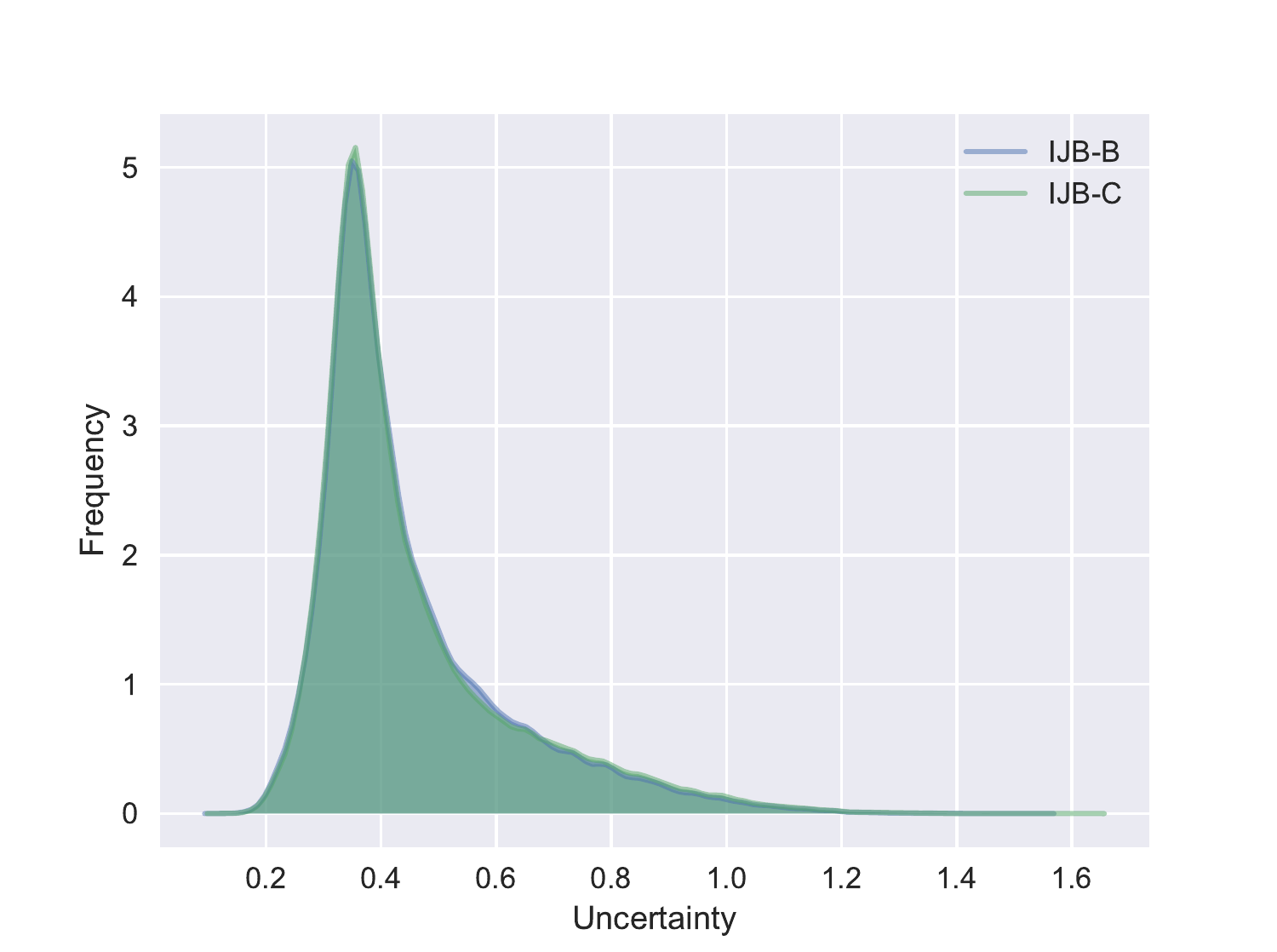}}\hfill
	\caption{~Distribution of estimated uncertainty ($\sigma^2$) on different datasets.}
	\label{fig:HistSigmaDatasets}
\end{figure}

\begin{figure}[htbp]
	\centering
	\subfloat[$L_{S} + L_{C}$]{%
		\includegraphics[width=.23\textwidth]{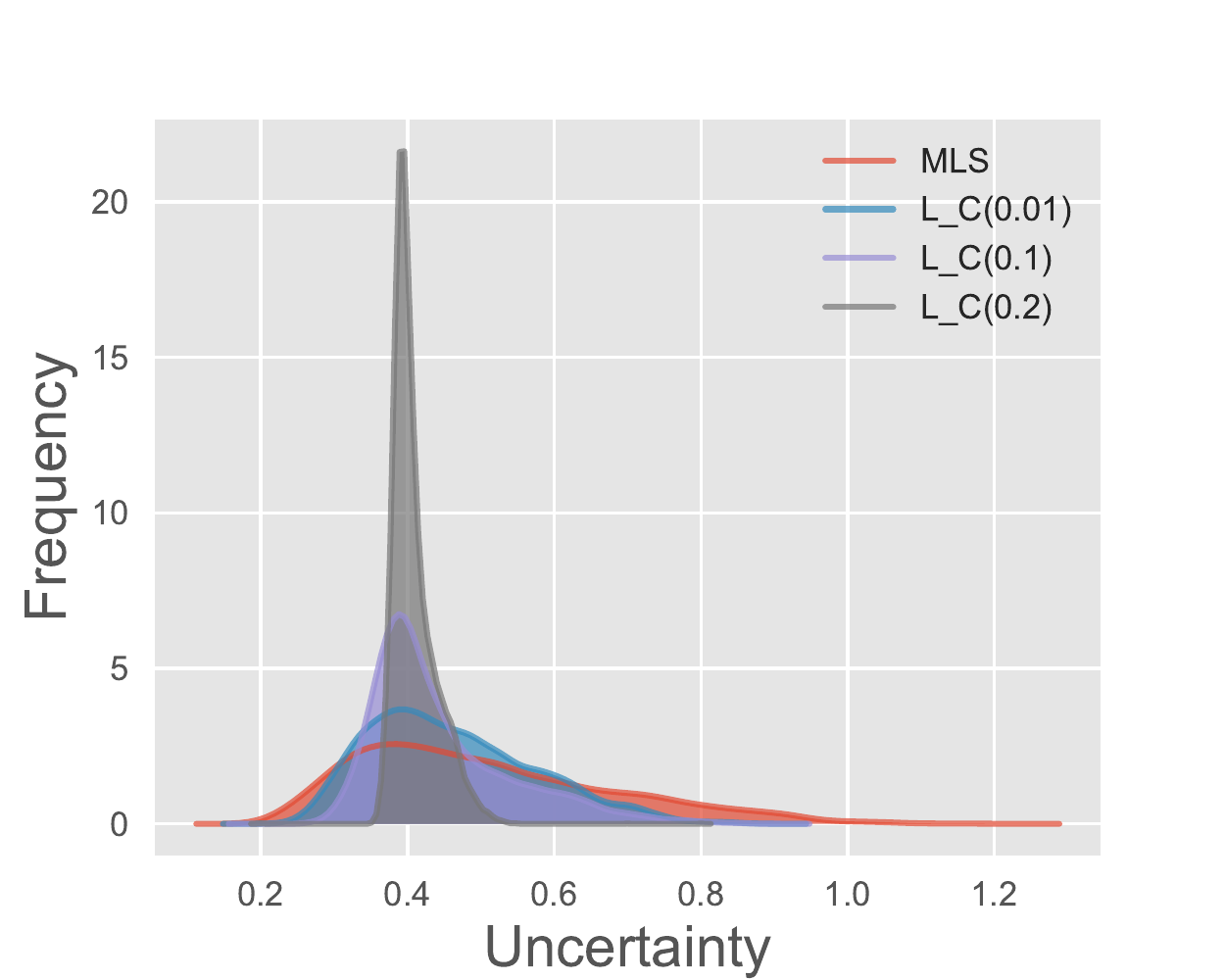}}\hfill
	\subfloat[$L_{S} + L_{Id}$]{%
		\includegraphics[width=.23\textwidth]{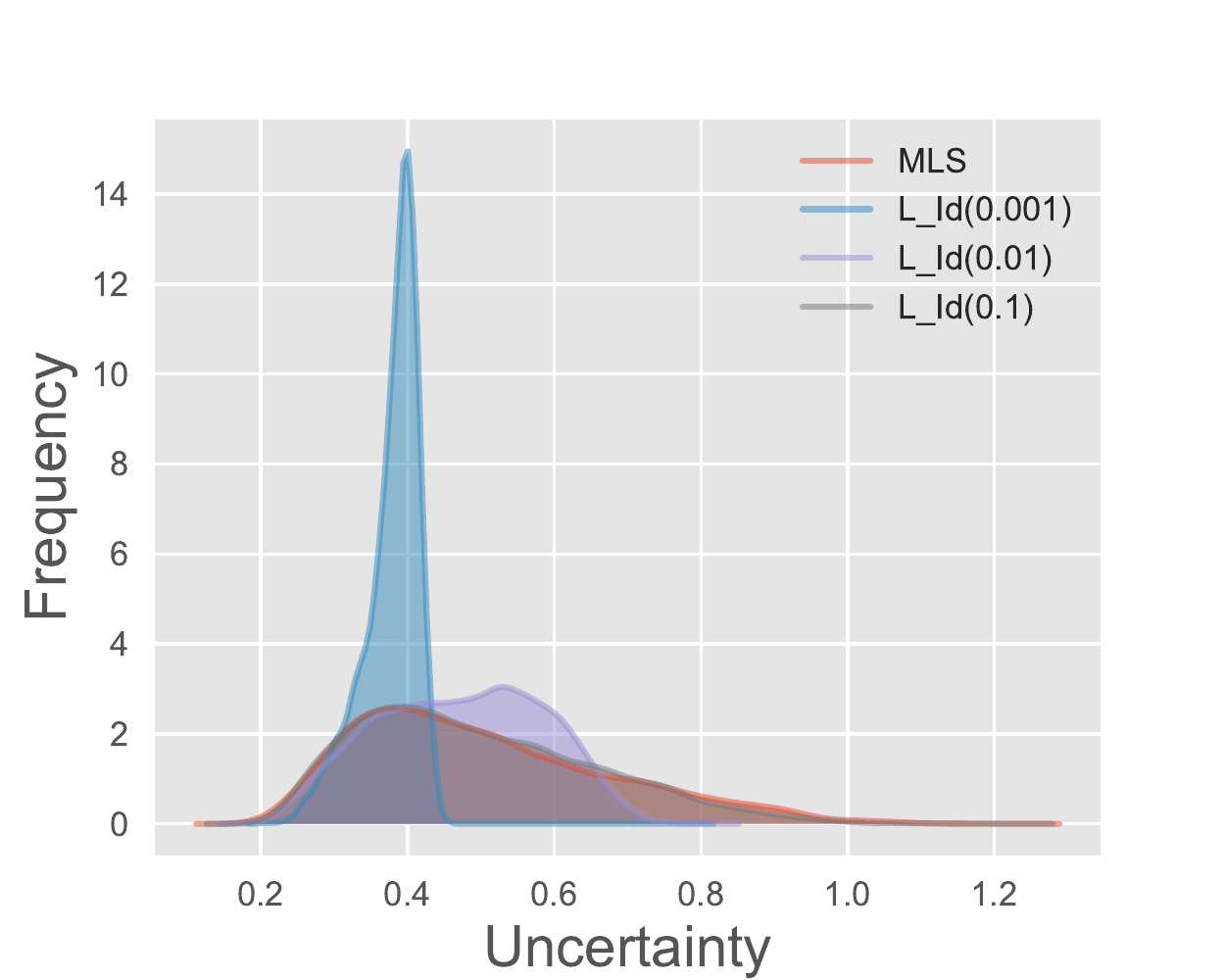}}\hfill
	\caption{~Distribution of estimated uncertainty ($\sigma^2$) on IJB-B with different losses. The weight of loss is given in brackets.
 (a) output-constraint loss. (b) identification preserving loss.}
	\label{fig:HistSigmaPrior}
\end{figure}

\begin{figure}[htbp]
	\centering
	\includegraphics[width=0.48\textwidth]{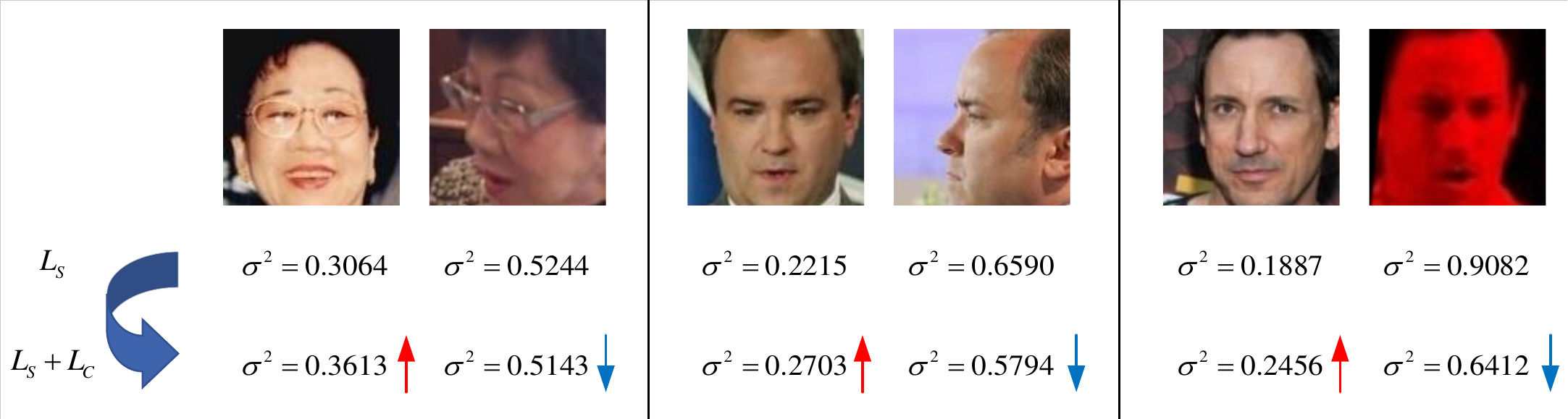}
	\caption{~Examples of the uncertainty score changes after adding output-constraint loss.}
	\label{fig:ExampleCplfw}
\end{figure}

In order to analyze the effect of the proposed losses on the uncertainty estimation, we use a histogram to show the distribution of uncertainty scores.
Figure \ref{fig:HistSigmaDatasets} shows the distribution of estimated uncertainty on testing datasets.
We can see that the uncertainty of LFW and CFP-FF with limited variation is smaller and more concentrated.
The uncertainties of large-pose face datasets CPLFW and CFP-FP are larger than LFW and CFP-FF. The uncertainty of CFP-FP has two obvious peaks, which are frontal and profile pictures respectively.
The large-age data sets, CALFW and AgeDB30, have large age changes, but the uncertainties remain small.
Figure \ref{fig:HistSigmaPrior}(a) output-constraint losses can concentrate the sigma distribution. With the increase of $\lambda_{C}$, its distribution will become more and more concentrated.
Figure \ref{fig:HistSigmaPrior}(b) show the effect of identification preserving loss. It can been seen that the effect of uncertainty-aware contractive loss and uncertainty-aware triplet loss on uncertainty output is not very obvious.
Figure \ref{fig:ExampleCplfw} demonstrates example images of the uncertainty score changes after adding output-constraint loss. We can see that the score of the front face is significantly lower than that of the profile face. After the output constraint is applied, the range of score changes decreases, as the front face score increases and the profile score decreases.

\begin{figure}[tb]
	\centering
	\subfloat[Different $\lambda_C$ with fixed $s=16$ and $m=0.4$]{%
		\includegraphics[width=.45\textwidth]{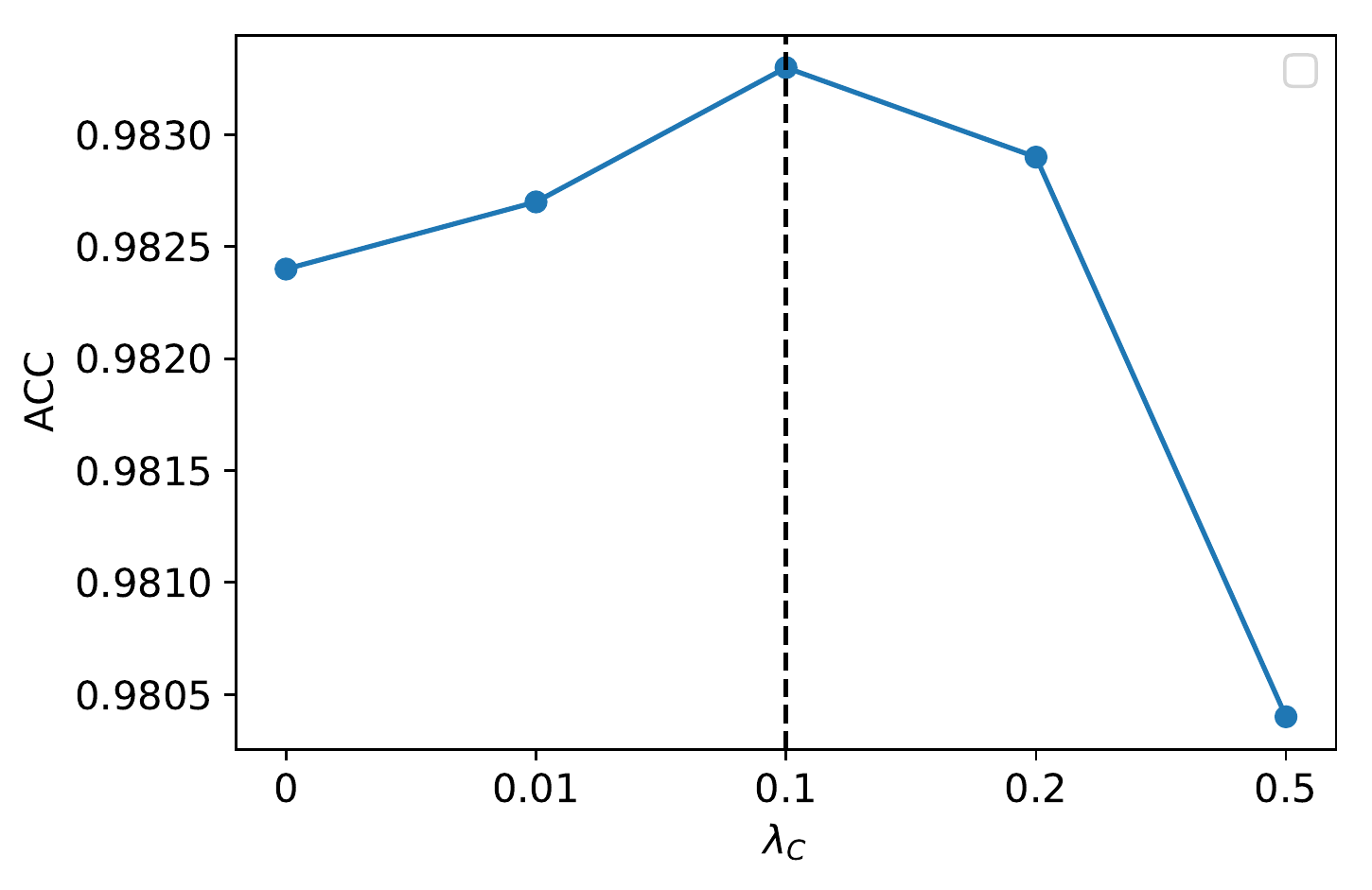}}\hfill
	\subfloat[Different $\lambda_{Id}$]{%
		\includegraphics[width=.45\textwidth]{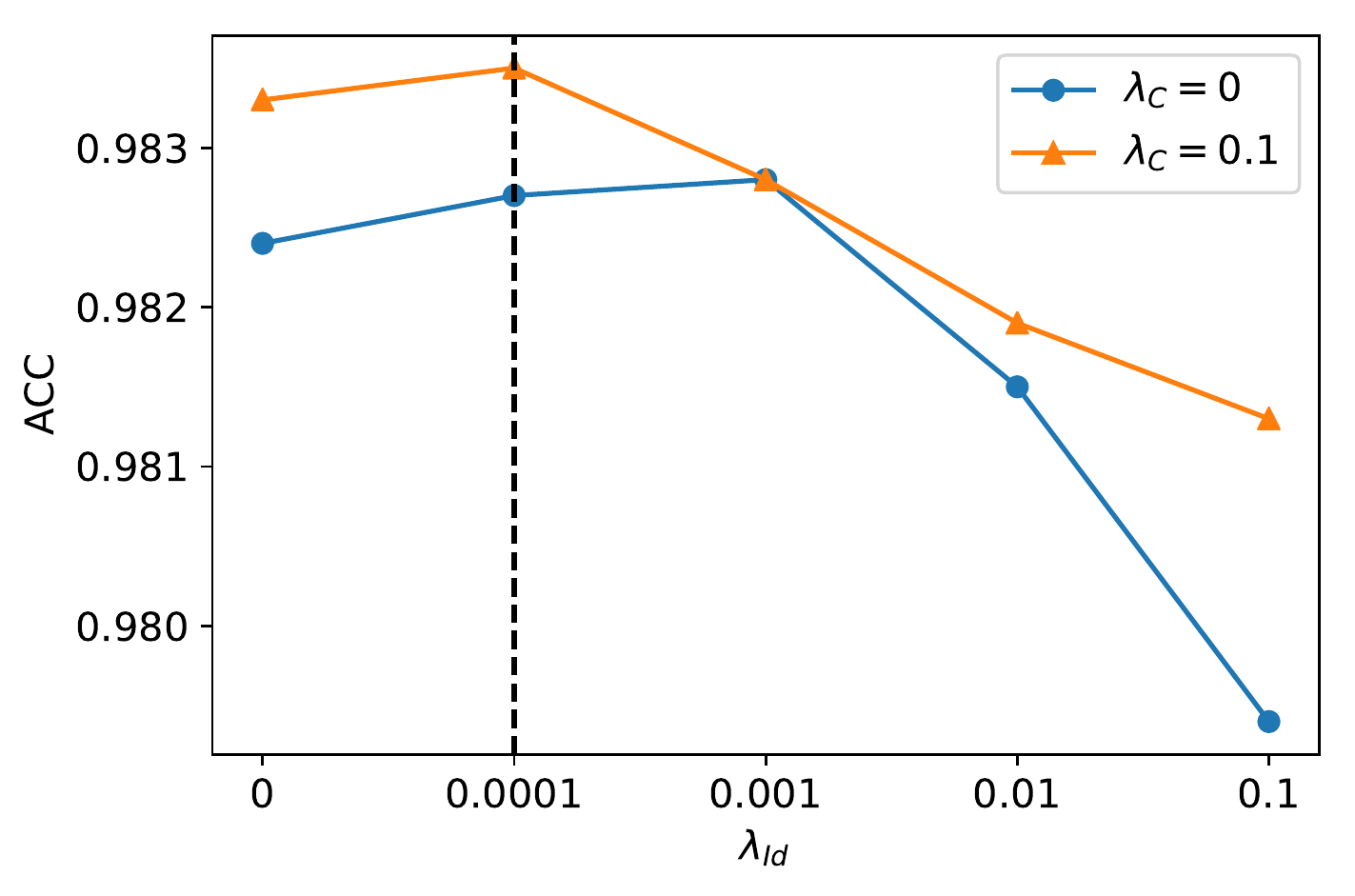}}\hfill
	\caption{~Parameters sensitivity results on the CFP-FP dataset.}
	\label{fig:param-sensitive}
\end{figure}

We analyze the sensitivity of our approach to the parameters $\lambda_{C}$ and $\lambda_{Id}$ on CFP-FP dataset based on ResFace64 in Figure \ref{fig:param-sensitive}.
The face verification accuracy increases to a peak and subsequently falls with increasing $\lambda_C$.
If the output constraint is too strong, the curves will drop because the range of uncertainty is too short and the capacity is insufficient, as demonstrated in Figure \ref{fig:HistSigmaPrior}(a).
At $\lambda_C=0.1$, the recognition performance reaches its peak.
Then, we change the $\lambda_{Id}$ with fixed $\lambda_{C}=0$ and $\lambda_{C}=0.1$.
The best performance obtained at the position of $\lambda_{Id}=0.0001$ in the curve of $\lambda_{C}=0.1$.
In subsequent experiments, unless otherwise specified, the hyper-parameter $\lambda_{C}$ is set to 0.1, and $\lambda_{Id}$ is set to 0.0001.

\subsubsection{Ablation on All Modules}

\begin{table}[htbp]
  \footnotesize
  \centering
  \caption{Ablation experiment (\%). The base model is ResFace64. MF denotes multi-layer feature fusion. }
    \begin{tabular}{l|ccc|c}
    \toprule
    Method & LFW   & CPLFW & CFP-FP & Avg \\
    \midrule
    Baseline & 99.80 & 92.53 & 98.04 & 96.79 \\
    \midrule
    PFE   & 99.82 & 92.80 & 98.33 & 96.98 \\
    \midrule
    $L_S$  & 99.80 & 93.01 & 98.29 & 97.03 \\
    $L_S$ + $L_C$ & \textbf{99.85} & 93.12 & 98.21 & 97.06 \\
    $L_S$ + $L_{Id}$ & 99.80 & 93.08 & 98.29 & 97.06 \\
    $L_S$ + $L_C$ + $L_{Id}$ & \textbf{99.85} & 93.17 & 98.39 & 97.13 \\
    $L_S$ + $L_C$ + $L_{Id}$ + MF & \textbf{99.85} & \textbf{93.53} & \textbf{98.41} & \textbf{97.27} \\
    \bottomrule
    \end{tabular}%
  \label{tab:ablation}%
\end{table}%

To verify the effectiveness of the proposed losses and modules,
we conduct an ablation study on all modules in Table \ref{tab:ablation}.
Compared with the baseline, both PFE and $L_S$ using the probabilistic embeddings can improve the performance on LFW, CPLFW and CFP-FP datasets.
The performance of $L_S$ is slightly higher than that of PFE, indicating that FastMLS can achieve comparable performance to MLS while increasing the calculation speed.
Adding $L_C$ and $L_{Id}$ alone, as well as adding $L_C$ and $L_{Id}$ at the same time, can improve performance, indicating that the proposed losses are effective.
At the end, we further add multi-layer feature fusion, which can get the best result.

\subsection{Runtime comparison of MLS and FastMLS}

\begin{figure}[htbp]
	\centering
	\subfloat[1:1 matching]{%
		\includegraphics[width=.49\textwidth]{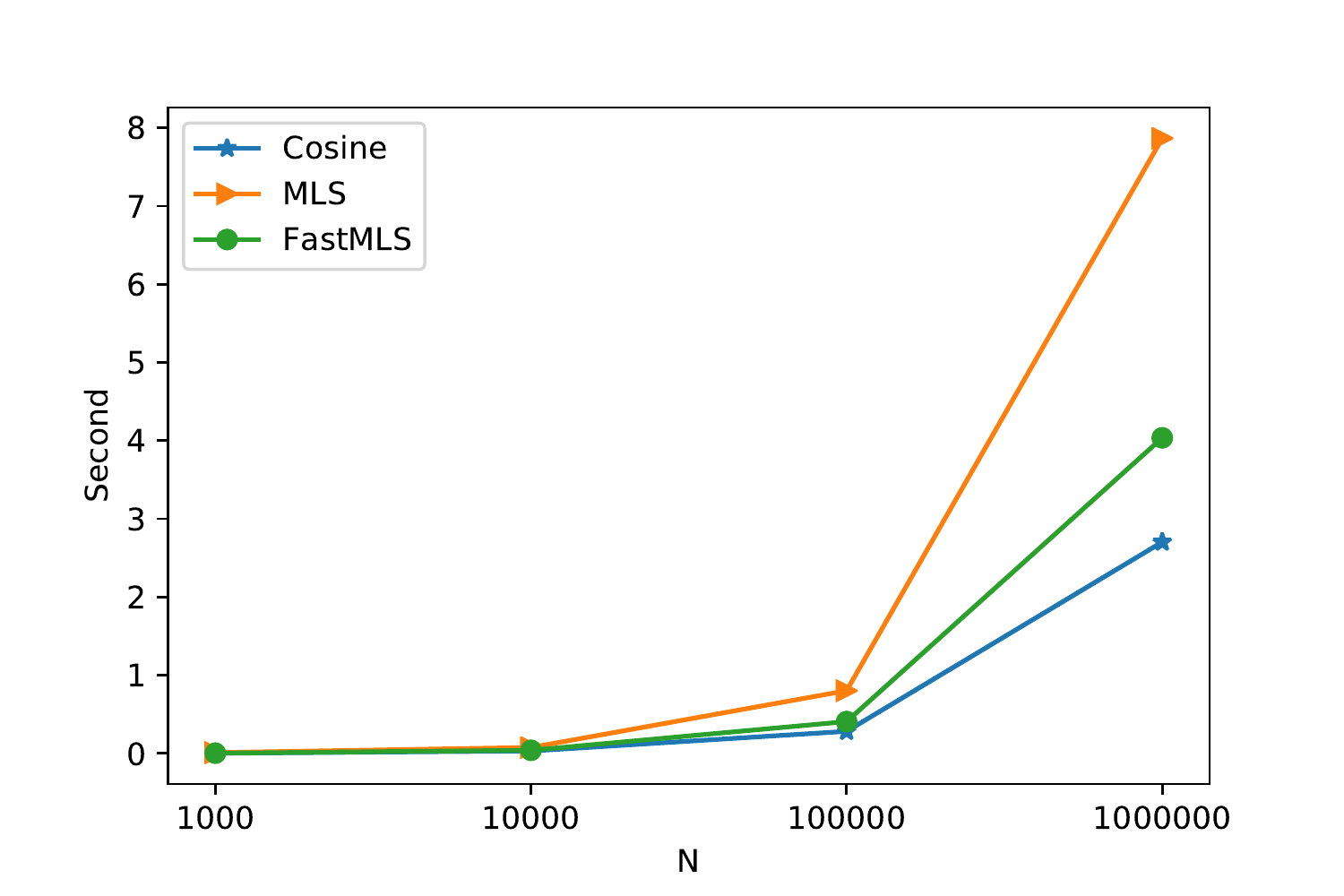}}\hfill
	\subfloat[N:N matching]{%
		\includegraphics[width=.49\textwidth]{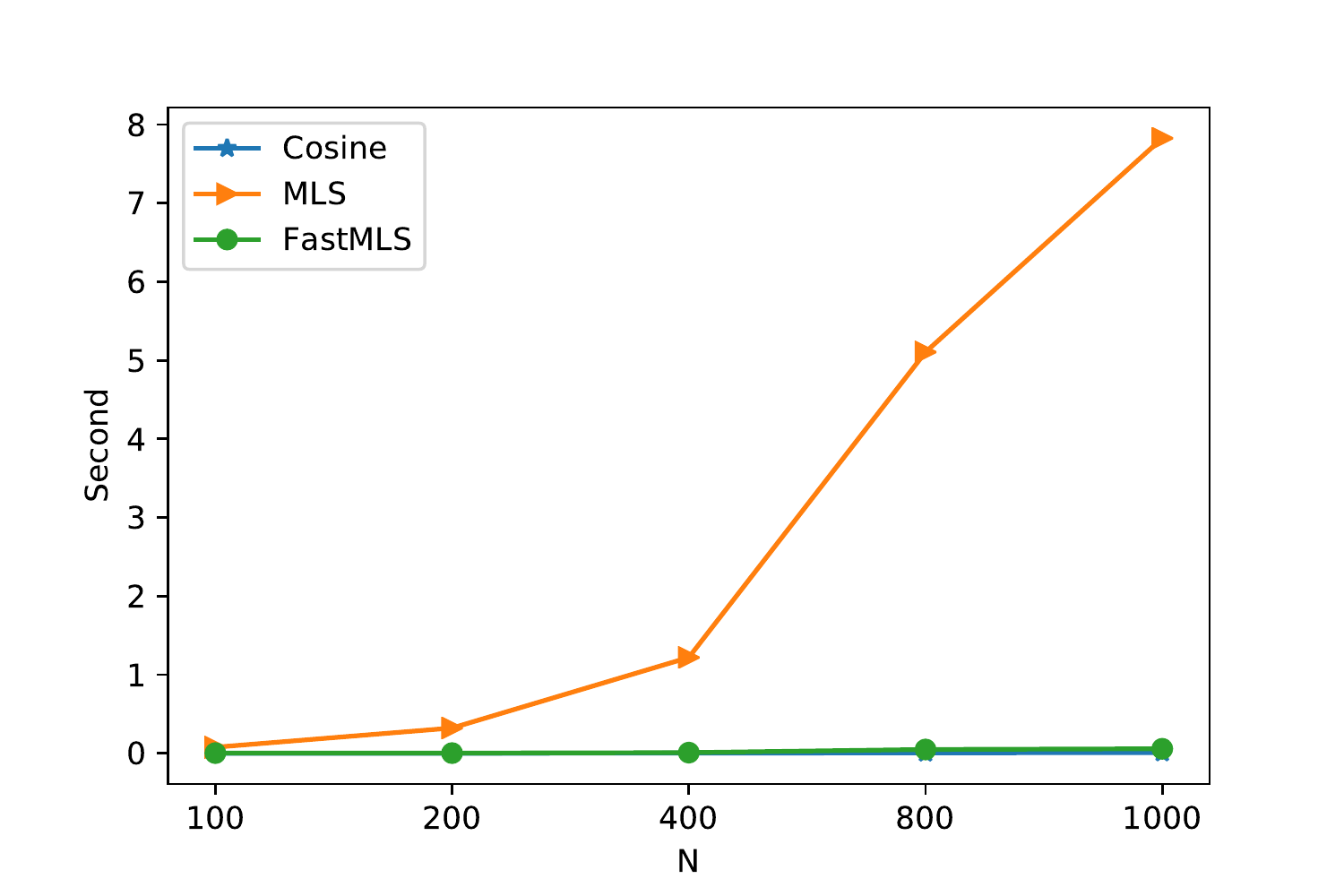}}\hfill
	\caption{~Runtime comparison of cosine, MLS and FastMLS metrics on generated 256-D features under 1:1 matching and N:N matching.}
	\label{fig:cmp_time}
\end{figure}

\begin{table}[htbp]
  \footnotesize
  \centering
  \caption{The results of runtime comparison of Cosine, MLS and FastMLS metrics on LFW dataset with 1:1 and 1:N protocal.}
    \begin{tabular}{ccclr}
    \hline
    \multicolumn{1}{l}{Protocol} & \multicolumn{1}{c}{\#Images} & \multicolumn{1}{l}{\#Matches} & Method & \multicolumn{1}{l}{Time(s)} \\
    \hline
    \multirow{3}[0]{*}{LFW 1:1} & \multirow{3}[0]{*}{12000} & \multirow{3}[0]{*}{6000} & Cosine   & 0.016 \\
          &       &       & MLS & 0.039 \\
          &       &       & FastMLS & 0.017 \\
    \hline
    \multirow{3}[0]{*}{LFW 1:N} &  Gallery: 596 & \multirow{3}[0]{*}{6,013,640} & Cosine   & 0.024 \\
          &   Probe: 10,090    &   & MLS & 26.619 \\
          &       &       & FastMLS & 0.251 \\
    \hline
    \end{tabular}%
  \label{tab:runtimecmp}%
\end{table}%

The use of MLS metric can boost the recognition performance, but also brings more calculations.
Table \ref{tab:runtimecmp} gives the comparison results of runtime by cosine, MLS and our FastMLS under 1:1 protocol and 1:N protocol of LFW.
The code for the calculation of cosine and MLS is written using Python Numpy library.
All experiments are conducted on a computer with 3.7GHz 6-cores Intel i7-8700K CPU and 32GB memory.
In the 1:1 protocol, the similarity scores between the given face pairs are calculated.
From the results, we can see that there are only 6000 pairs in LFW 1:1 protocol, so that the runtime of three methods are all short, where FastMLS is close to cosine, and MLS is the longest.
In the 1:N protocol, face images are separated into gallery set and probe set, and then the similarity scores between the images of the two sets are calculated.
We can see that the number of comparisons in the 1:N protocol is comparatively large. Cosine metric and our FastMLS metric can use matrix multiplication to get the result within 1 second, while the MLS takes 26 seconds.

Figure \ref{fig:cmp_time} shows the runtime results of generated 256-D features under 1:1 matching and N:N matching.
The 1:1 matching is the same as 1:1 protocol of LFW, in which $N$ pairs of features are matched $N$ times.
The N:N matching is a match between all the features in pairs, in which $N$ features produce $N^2$ matches.
As the number of comparisons increases, the runtime of MLS will increase more, while the runtime of FastMLS is slightly higher than cosine and will not increase too much.

\subsection{Comparison with State-Of-The-Art Face Recognition Methods}

\begin{table*}[htbp]
  \footnotesize
	\centering
	\caption{Comparison with the state-of-the-art methods on limited, large-age and large pose datasets. TT-Flip denotes test-time flip which uses the aggregation features of image and flipped image during the test time.}
    \resizebox{\textwidth}{!}{\begin{tabular}{l|ccc|cc|cc|ccc|c}
    \toprule
    \multicolumn{1}{c|}{\multirow{2}[2]{*}{Methods}} & \multicolumn{1}{c}{\multirow{2}[2]{*}{Metric}} & \multicolumn{1}{c}{\multirow{2}[2]{*}{TT-Flip}} & \multicolumn{1}{c|}{\multirow{2}[2]{*}{\#Image}} & \multicolumn{2}{c|}{Limited} & \multicolumn{2}{c|}{Large-age} & \multicolumn{3}{c|}{Large-pose} & \multirow{2}[2]{*}{Avg} \\
          &       &       &       & LFW   & CFP-FF & CALFW & AgeDB & CPLFW & CFP-FP & Vgg2FP &  \\
    \midrule
    FaceNet \cite{schroff2015facenet} & Cosine & N     & 200M  & 99.63 & -     & -     & -     & -     & -     & -     & - \\
    CenterFace \cite{wen2016discriminative} & Cosine & Y     & 0.7M  & 99.28 & -     & -     & -     & -     & -     & -     & - \\
    SphereFace \cite{liu2017sphereface} & Cosine & Y     & 0.5M  & 99.42 & -     & -     & -     & -     & -     & -     & - \\
    CosFace \cite{wang2018cosface} & Cosine & Y     & 5M    & 99.73 & -     & -     & -     & -     & -     & -     & - \\
    MobileFaceNet,ArcFace \cite{deng2019arcface} & Cosine & Y     & 5.8M  & 99.50 & -     & -     & 95.91 & -     & 88.94 & -     & - \\
    LResNet50E-IR,ArcFace \cite{deng2019arcface} & Cosine & Y     & 5.8M  & 99.80 & -     & -     & 97.76 & -     & 92.74 & -     & - \\
    LResNet100E-IR,ArcFace \cite{deng2019arcface} & Cosine & Y     & 5.8M  & 99.77 & -     & -     & 98.28 & -     & 98.27 & -     & - \\
    IR-50(Arcface) \cite{faceevoLVePyTorch} & Cosine & Y     & 5.8M  & 99.78 & 99.69 & 95.87 & 97.53 & 92.45 & 98.14 & 95.22 & 96.95 \\
    IR-152(Arcface) \cite{faceevoLVePyTorch} & Cosine & Y     & 5.8M  & 99.82 & \textbf{99.83} & 96.03 & 98.07 & 93.05 & 98.37 & \textbf{95.50} & \textbf{97.24} \\
    GroupFace \cite{kim2020groupface}  & Cosine & Y     & 5.8M  & \textbf{99.85} & -     & \textbf{96.20} & 98.28 & \textbf{93.17} & 98.63 & -     & - \\
    CurricularFace \cite{huang2020curricularface} & Cosine & -     & 5.9M  & 99.80 & -     & \textbf{96.20} & \textbf{98.32} & 93.13 & 98.37 & -     & - \\
    DUL-cls \cite{chang2020data} & Cosine & -     & 3.6M  & 99.78 & -     & -     & -     & -     & 98.67 & -     & - \\
    DUL-rgs \cite{chang2020data} & Cosine & -     & 3.6M  & 99.83 & -     & -     & -     & -     & \textbf{98.78} & -     & - \\
    Sphereface-PFE \cite{shi2019probabilistic} & MLS   & N     & 4.4M  & 99.82 & 99.70 & 95.85 & 96.93 & 91.78 & 97.56 & 94.94 & 96.65 \\
    \midrule
    Resface64s2 & Cosine & N     & 5.8M  & 99.72 & 99.76 & 95.67 & \textbf{97.73} & 91.90 & 97.50 & 94.24 & 96.64 \\
    Resface64s2-PFE & MLS   & N     & 5.8M  & \textbf{99.82} & 99.74 & 95.77 & 97.43 & 92.47 & 97.90 & 94.84 & 96.85 \\
    Resface64s2-ProbFace & FastMLS   & N     & 5.8M  & \textbf{99.82} & \textbf{99.83} & \textbf{95.82} & 97.57 & \textbf{92.92} & \textbf{97.93} & \textbf{95.00} & \textbf{96.98} \\
    \midrule
    Resface64 & Cosine & N     & 5.8M  & 99.80 & \textbf{99.80} & 95.93 & \textbf{97.93} & 92.53 & 98.04 & 94.92 & 96.99 \\
    Resface64-PFE & MLS   & N     & 5.8M  & 99.82 & 99.77 & 95.85 & 97.52 & 92.80 & 98.33 & 95.22 & 97.04 \\
    Resface64-ProbFace & FastMLS   & N     & 5.8M  & \textbf{99.85} & \textbf{99.80} & \textbf{96.02} & 97.90 & \textbf{93.53} & \textbf{98.41} & \textbf{95.34} & \textbf{97.26} \\
    \midrule
    Resface100 & Cosine & N     & 5.8M  & 99.82 & \textbf{99.84} & 95.97 & 98.00 & 93.38 & 98.71 & 95.50 & 97.32 \\
    Resface100-PFE & MLS   & N     & 5.8M  & 99.80 & \textbf{99.84} & 95.88 & 97.85 & 93.82 & 98.49 & 95.12 & 97.26 \\
    Resface100-ProbFace & FastMLS   & N     & 5.8M  & \textbf{99.83} & \textbf{99.84} & \textbf{96.02} & \textbf{98.15} & \textbf{93.97} & \textbf{98.81} & \textbf{95.34} & \textbf{97.42} \\
    \bottomrule
    \end{tabular}}
  \label{tab:cmp_sota_lfw}%
\end{table*}%

\begin{table*}[htbp]
  \footnotesize
	\centering
	\caption{Comparison with the state-of-the-art methods on IJB-B and IJB-C with 1:1 verification protocol. TT-Flip denotes test-time flip which uses the aggregation features of image and flipped image during the test time.}
    \resizebox{\textwidth}{!}{\begin{tabular}{l|ccc|cccc|cccc}
    \toprule
    \multicolumn{1}{c|}{\multirow{2}[2]{*}{Methods}} & \multicolumn{1}{c}{\multirow{2}[2]{*}{Metric}} & \multicolumn{1}{c}{\multirow{2}[2]{*}{TT-Flip}} & \multicolumn{1}{c|}{\multirow{2}[2]{*}{\#Image}} & \multicolumn{4}{c|}{IJB-B (TPR@FPR)} & \multicolumn{4}{c}{IJB-C (TPR@FPR)} \\
          &       &       &       & 1e-5  & 1e-4  & 1e-3  & 1e-2  & 1e-5  & 1e-4  & 1e-3  & 1e-2 \\
    \midrule
    VGG2-ResNet50-ArcFace \cite{deng2019arcface} & Cosine & Y     & 5.8M  & 80.38 & 89.76 & 94.37 & 97.55 & 86.12 & 92.14 & 95.95 & \textbf{98.23} \\
    MS1MV2-ResNet100-ArcFace \cite{deng2019arcface} & Cosine & Y     & 5.8M  & 90.42 & 94.67 & 96.20 & \textbf{97.61} & 93.15 & 95.65 & \textbf{97.20} & 98.18 \\
    Multicolumn \cite{xie2018multicolumn} & Cosine & -     & 3.3M  & 70.80 & 83.10 & 90.90 & 95.80 & 77.10 & 86.20 & 92.70 & 96.80 \\
    GroupFace \cite{kim2020groupface} & Cosine & Y     & 5.8M  & \textbf{91.24} & \textbf{94.93} & -     & -     & \textbf{94.53} & \textbf{96.26} & -     & - \\
    CurricularFace \cite{huang2020curricularface} & Cosine & -     & 5.9M  & -     & 94.80 & -     & -     &       & 96.10 & -     & - \\
    DUL-cls \cite{chang2020data} & Cosine & -     & 3.6M  & -     & -     & -     & -     & 88.18 & 94.61 & 96.70 & - \\
    DUL-rgs \cite{chang2020data} & Cosine & -     & 3.6M  & -     & -     & -     & -     & 90.23 & 94.21 & 96.32 & - \\
    Sphereface-PFE \cite{shi2019probabilistic} & MLS   & N     & 4.4M  & 87.71 & 93.11 & \textbf{95.90} & 97.44 & 89.64 & 93.25 & 95.49 &  \\
    \midrule
    Resface64(0.5) & Cosine & N     & 5.8M  & 88.12 & 93.31 & 95.61 & 97.14 & 92.52 & 95.05 & 96.66 & 97.86 \\
    Resface64(0.5)+PFE & MLS   & N     & 5.8M  & 89.35 & 93.89 & 95.99 & 97.27 & 93.26 & 95.56 & 97.16 & 98.25 \\
    Resface64(0.5)+ProbFace & FastMLS   & N     & 5.8M  & \textbf{89.53} & \textbf{94.01} & \textbf{96.20} & \textbf{97.60} & \textbf{93.37} & \textbf{95.72} & \textbf{97.31} & \textbf{98.43} \\
    \midrule
    Resface64 & Cosine & N     & 5.8M  & 89.92 & 93.89 & 96.16 & 97.57 & 93.24 & 95.42 & 97.16 & 98.27 \\
    Resface64+PFE & MLS   & N     & 5.8M  & 90.58 & 94.38 & 96.12 & 97.42 & 93.90 & 95.98 & 97.36 & 98.34 \\
    Resface64+ProbFace & FastMLS   & N     & 5.8M  & \textbf{90.86} & \textbf{94.52} & \textbf{96.51} & \textbf{97.71} & \textbf{94.12} & \textbf{96.22} & \textbf{97.65} & \textbf{98.59} \\
    \midrule
    Resface100 & Cosine & N     & 5.8M  & 88.92 & 94.59 & 96.62 & 97.61 & 93.39 & 96.09 & 97.55 & 98.49 \\
    Resface100-PFE & MLS   & N     & 5.8M  & \textbf{91.48} & 94.91 & 96.34 & 97.40 & \textbf{94.72} & 96.48 & 97.65 & 98.44 \\
    Resface100-ProbFace & FastMLS   & N     & 5.8M  & 91.13 & \textbf{95.17} & \textbf{96.63} & \textbf{97.69} & 94.57 & \textbf{96.65} & \textbf{97.78} & \textbf{98.59} \\
    \bottomrule
    \end{tabular}}
  \label{tab:cmp_sota_ijbb}%
\end{table*}%

In Table \ref{tab:cmp_sota_lfw}, we compare the results with state-of-the-art methods on 7 datasets with limited, large-age and large pose images.
The base models are resface64(0.5), resface64 and resface100. The training data set is MS-Celba-1M-v2 with 5.8M images \cite{deng2019arcface} .
The results of the comparison algorithms in the table come from published papers and models.
Among them, FaceNet \cite{schroff2015facenet} is one of the important early ones. It is trained with Triplet loss and has an accuracy of 99.63\% in LFW.
CenterFace \cite{wen2016discriminative} and SphereFace \cite{liu2017sphereface} have the lower results since they are trained on small dataset CASIA Webface \cite{yi2014learning}.
In the results of ArcFace method \cite{deng2019arcface} includes MobileFaceNet, LResNet50E-IR and LResNet100E-IR.
We also compare the re-implementation results of ArcFace based on Pytorch \cite{faceevoLVePyTorch}, including two backbone networks, IR-50 and IR-152.
In addition, we also compare recent methods such as GroupFace \cite{kim2020groupface}, CurricularFace \cite{huang2020curricularface}, DUL \cite{chang2020data}, Universal \cite{shi2020towards}.
Compared with these methods, our method can get comparable or better results on all 7 datasets without test-time flip.
Compared with the same type of method PFE, our proposed ProbFace method has better results for most datasets on all 3 base models.
In the case of the small base model ResFace(0.5), the improvement in the use of ProbFace is more obvious, and the average accuracy can be increased from 96.64\% to 96.98\%.

Table \ref{tab:cmp_sota_ijbb} reports the comparison results with state-of-the-art methods on set-to-set video face recognition dataset IJB-B and IJB-C.
These datasets contain a variety of face images, resulting in wide range of quality variations.
As in \cite{shi2019probabilistic}, the uncertainty estimated is used to perform a weighted average of the face features for each video frame set.
So that, for three base models, both datasets can achieve good performance when using probabilistic embedding methods, including PFE and proposed ProbFace.
In comparison, the improvement of ProbFace is higher than that of PFE in most results.
At the same time, compared to the large model ResFace64 and ResFace100, the improvement effect of using ProbFace on the small model ResFace64(0.5) is more obvious since it is more difficult to handle low-quality images for small model.
ROC curves of IJB-B and IJB-C can refer to the \ref{sub:appendix_ijbb}.

\subsection{Face Recognition with Data Noise}

\begin{figure}[htbp]
	\centering
	\includegraphics[width=0.48\textwidth]{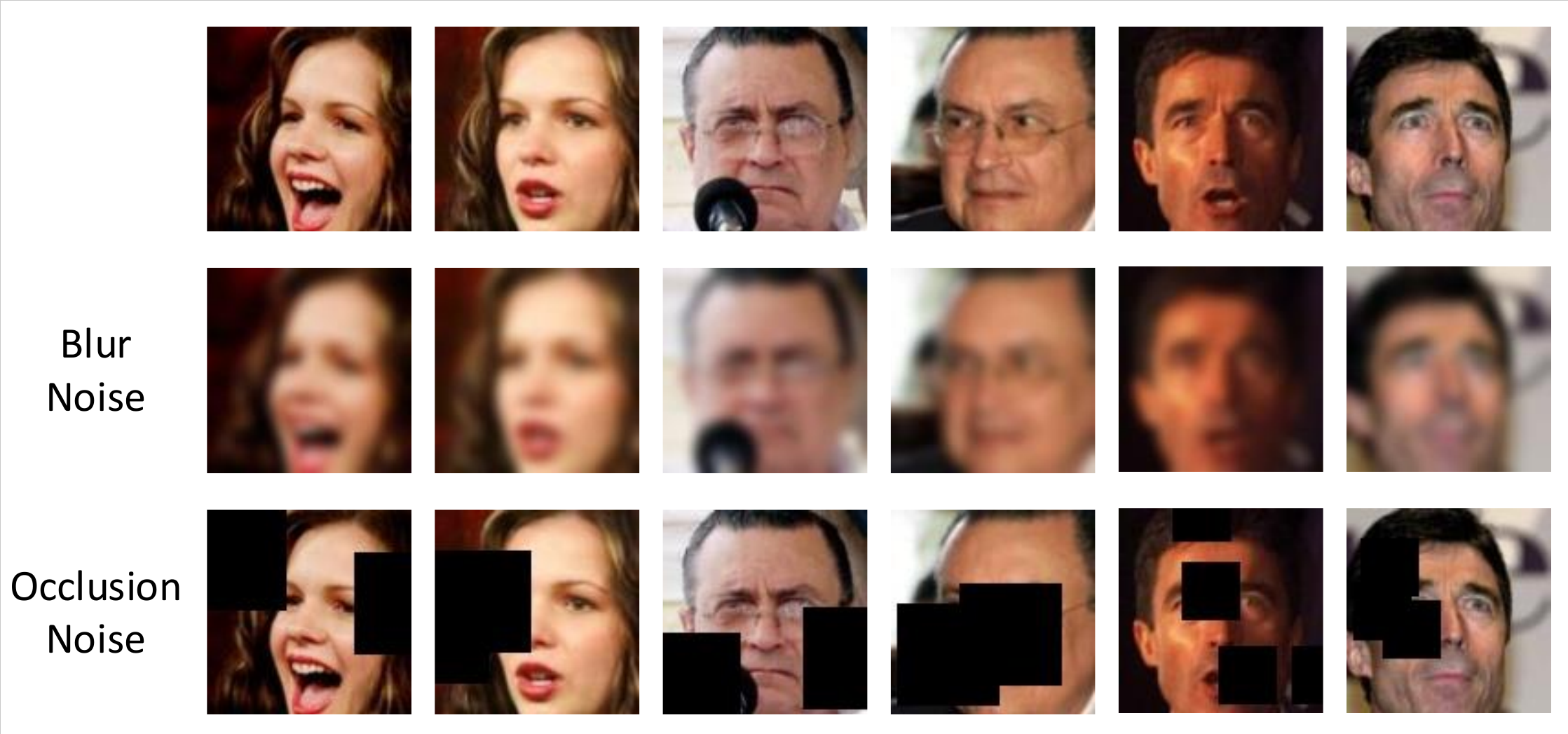}
	\caption{~Images with blur and occlusion noise.}
	\label{fig:ExampleNoise}
\end{figure}

\begin{figure*}[htbp]
	\centering
	\subfloat[LFW with blur noise]{%
		\includegraphics[width=.33\textwidth]{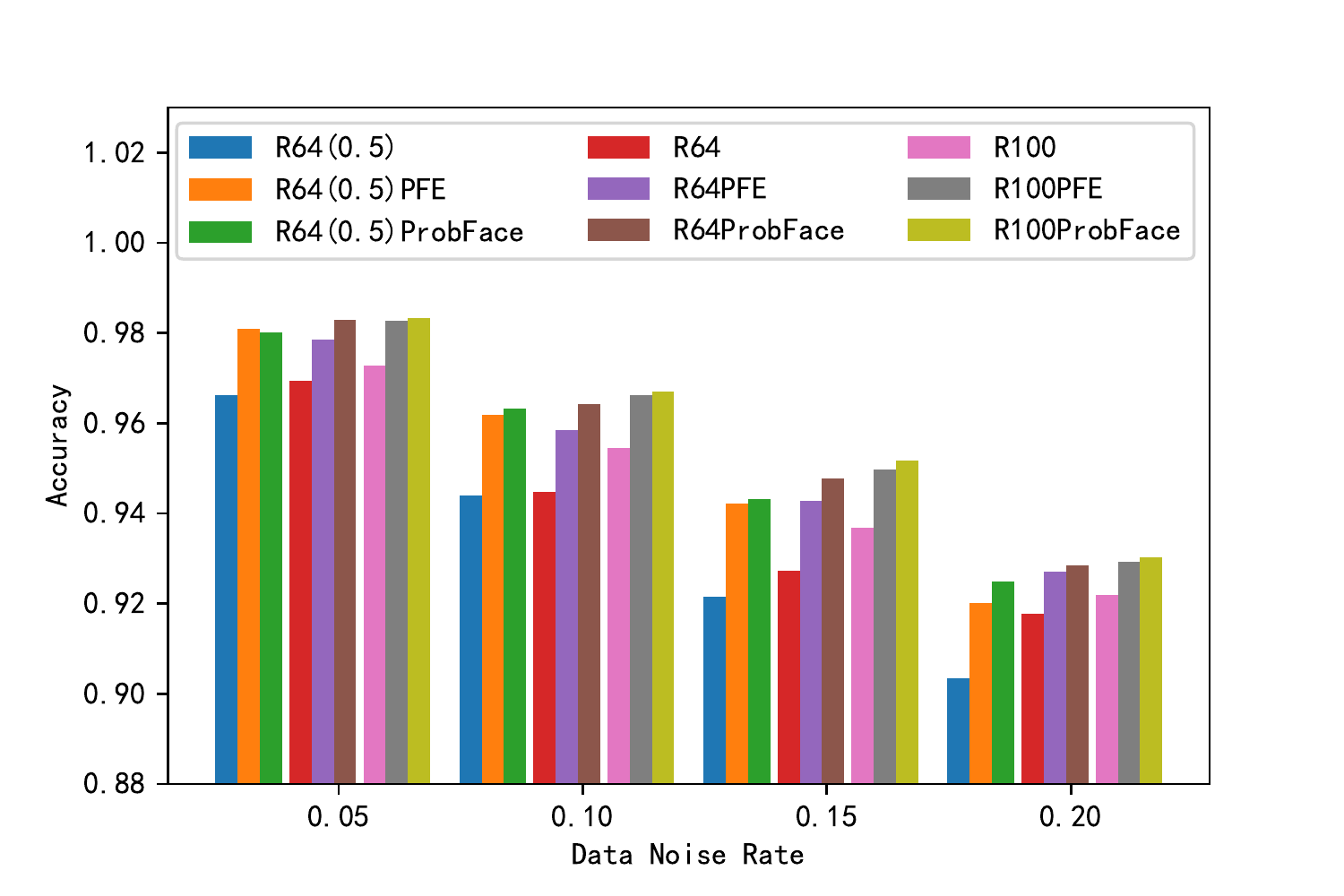}}\hfill
	\subfloat[CFP-FP with blur noise]{%
		\includegraphics[width=.33\textwidth]{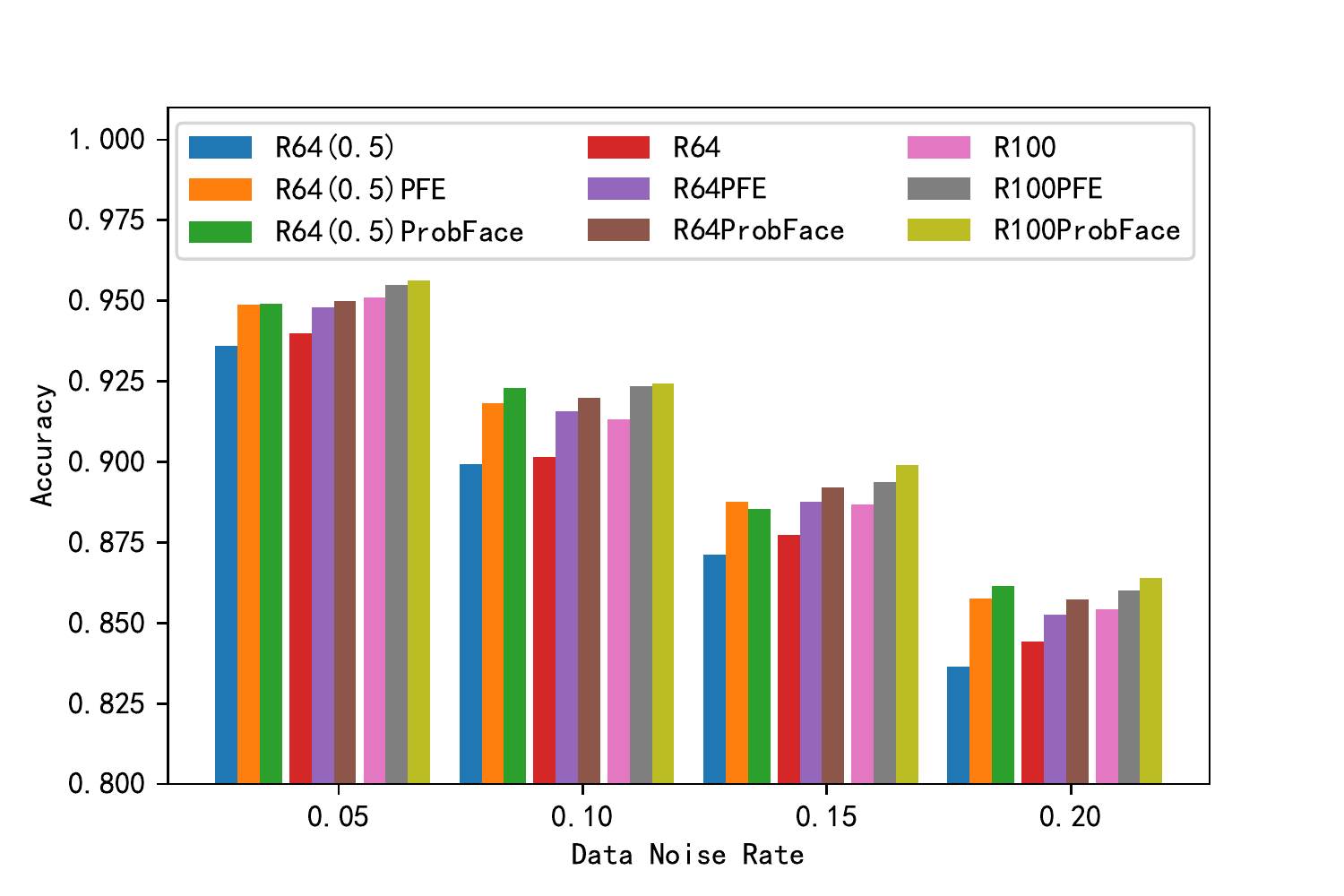}}\hfill
	\subfloat[AgeDB with blur noise]{%
		\includegraphics[width=.33\textwidth]{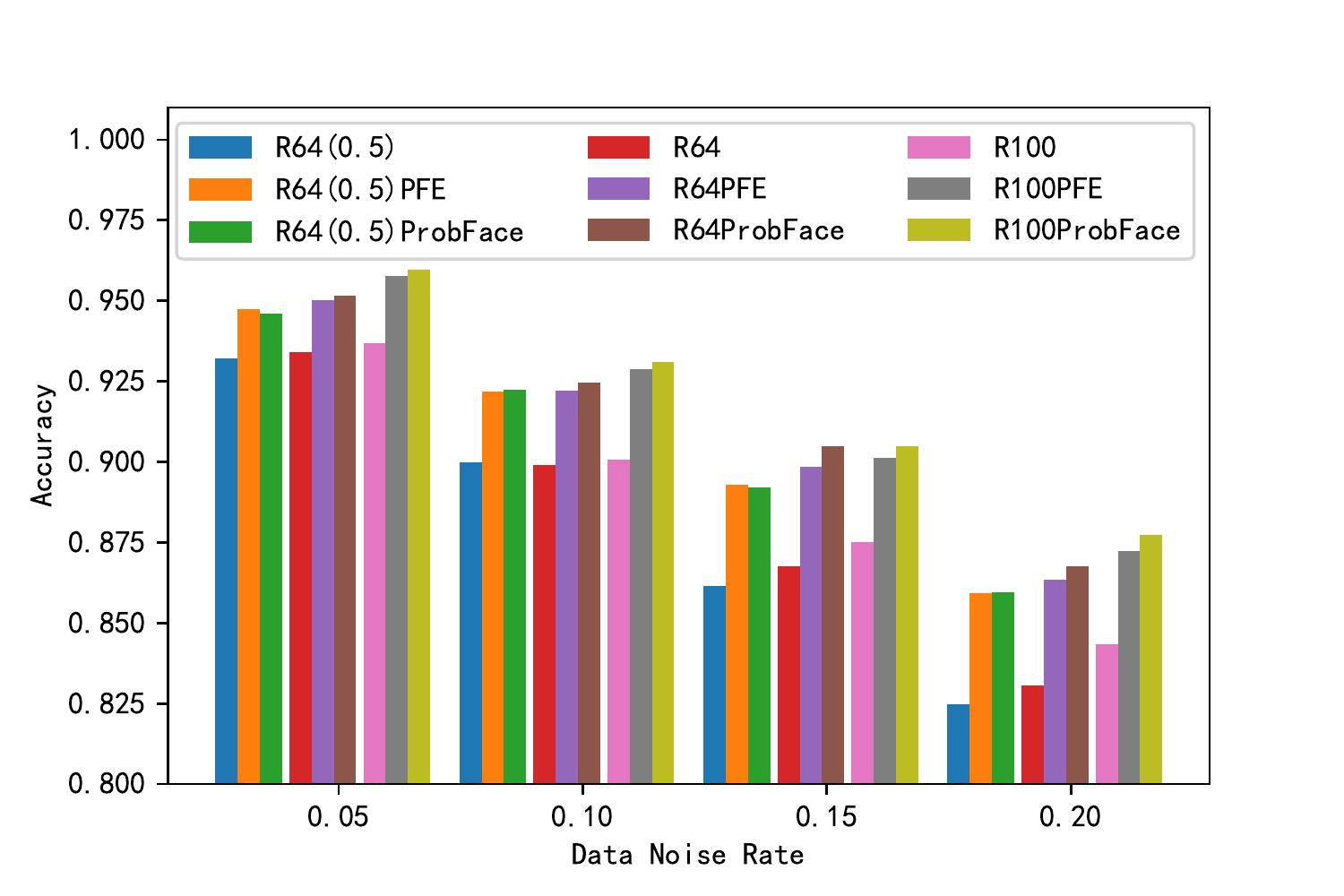}}\hfill
	\subfloat[LFW with occlusion noise]{%
		\includegraphics[width=.33\textwidth]{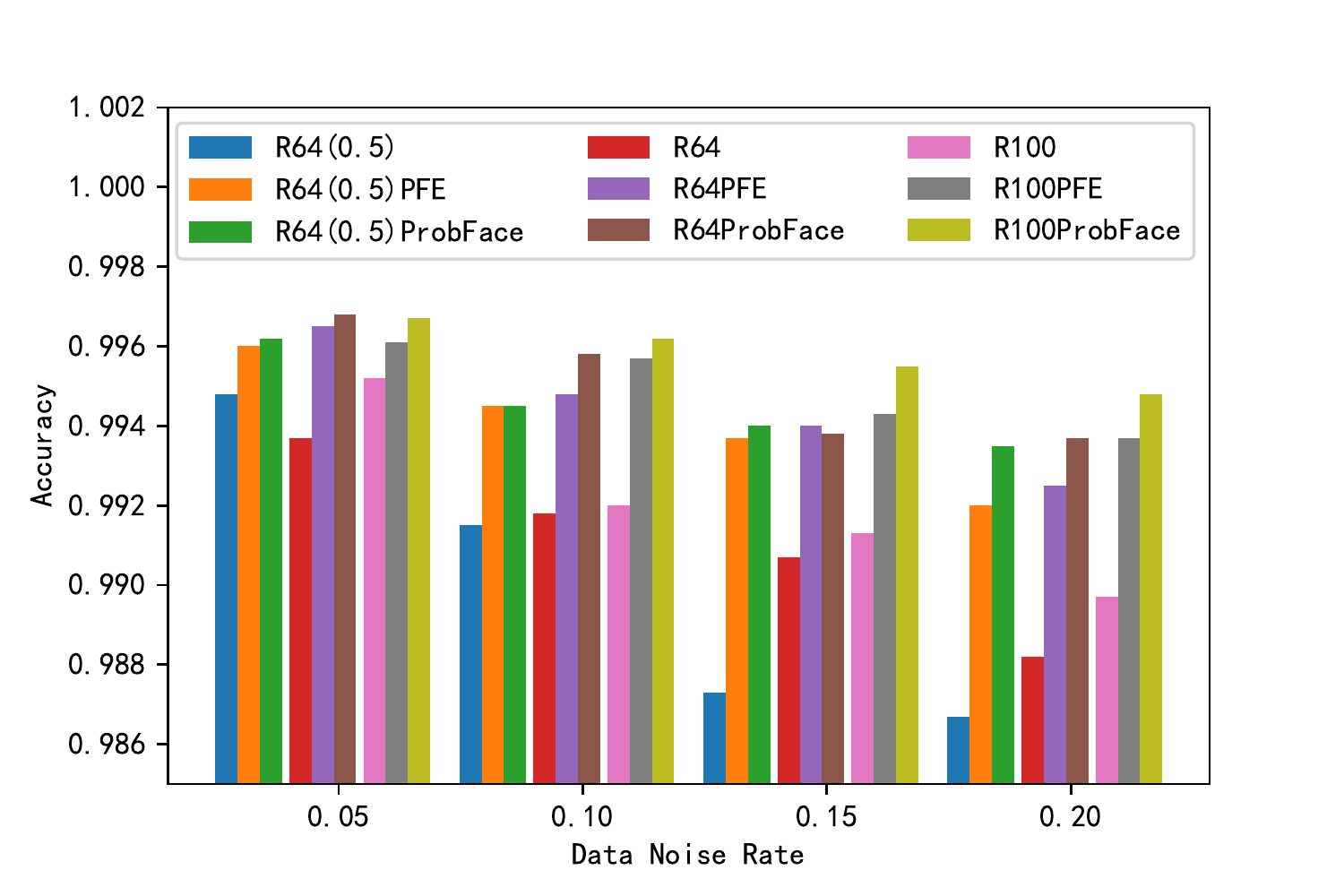}}\hfill
	\subfloat[CFP-FP with occlusion noise]{%
		\includegraphics[width=.33\textwidth]{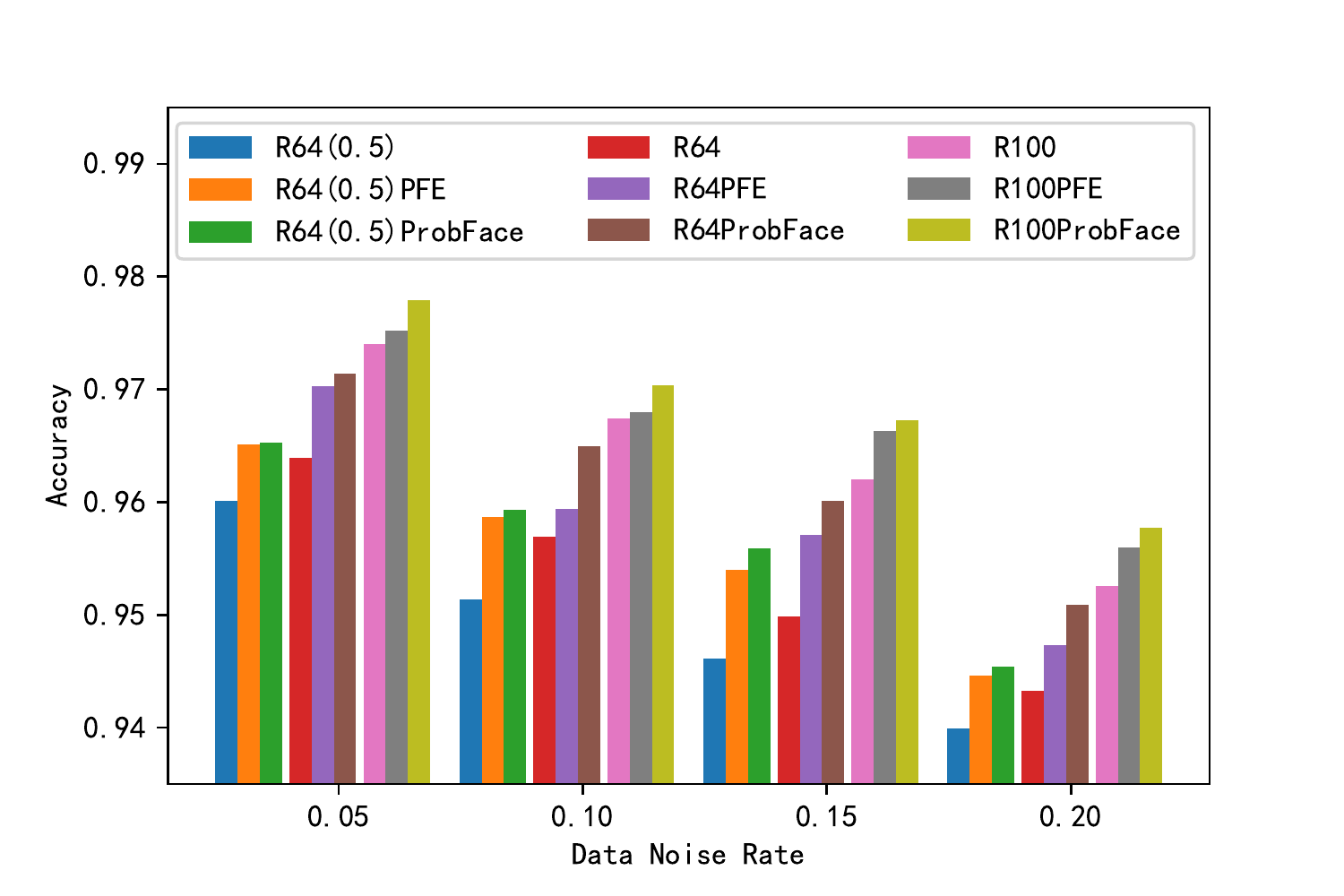}}\hfill
	\subfloat[AgeDB with occlusion noise]{%
		\includegraphics[width=.33\textwidth]{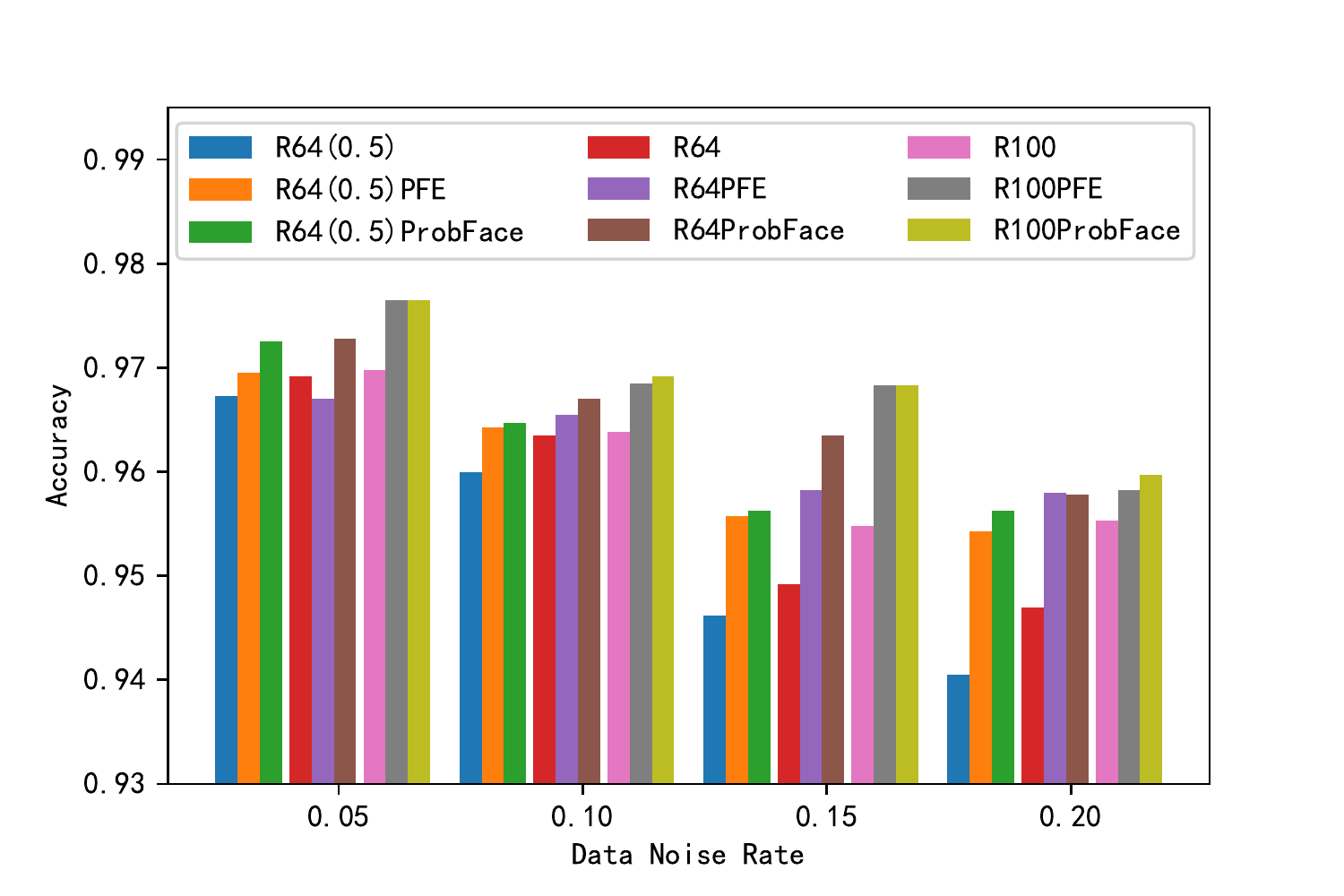}}\hfill
	\caption{~Verification accuracy of face recognition on datasets with blur and occlusion noise.}
	\label{fig:hist_noise}
\end{figure*}

In this section, we analyze the impact of data noise on face recognition. We take blur and occlusion noise as examples for experiments, as shown in the Figure \ref{fig:ExampleNoise}.
Among them, blur noise uses Gaussian blur with a Gaussian radius of 5.
Face occlusion uses random erasure of part of the rectangular area.
We add noise to the LFW, CFP-FP and AgeDB30 datasets in proportions of 0.05, 0.1, 0.15, 0.2.
Figure \ref{fig:hist_noise} shows the face recognition results of the noise dataset based on LFW, CFP-FP and AgeDB30.
Both PFE and ProbFace can improve the accuracy of face recognition under noisy data for all datasets and base models. As the proportion of the noise data increases, the improvement becomes more obvious. On most datasets and base models, ProbFace has the best results.
Therefore, the results show the superiority of ProbFace in face recognition under noisy data. Detailed results can refer to the \ref{sub:appendix_noise}.

\subsection{Uncertainty Estimation Comparison}

\begin{figure*}
	\centering
	\subfloat[LFW]{%
		\includegraphics[width=.45\textwidth]{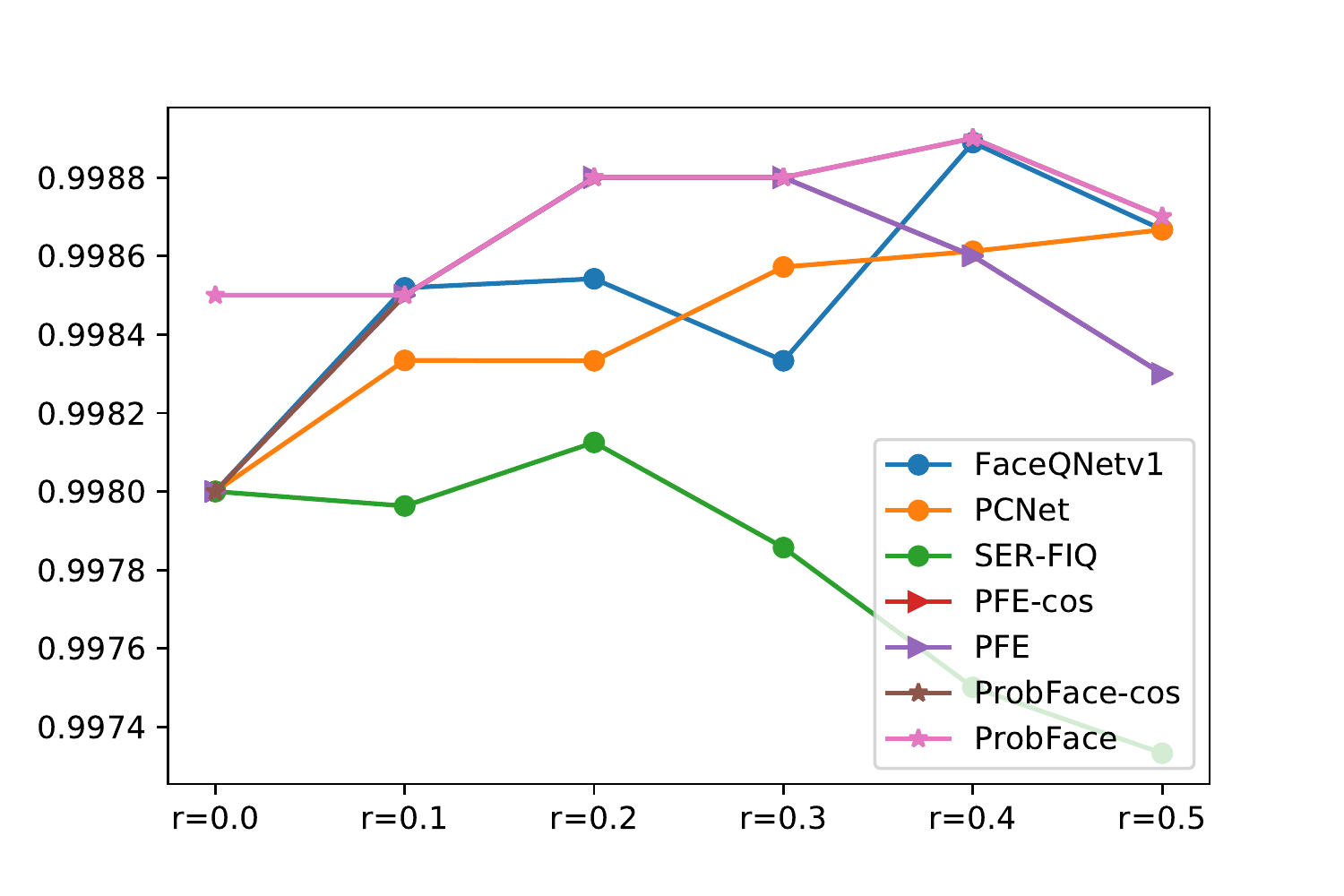}}\hfill
	\subfloat[CALFW]{%
        \includegraphics[width=.45\textwidth]{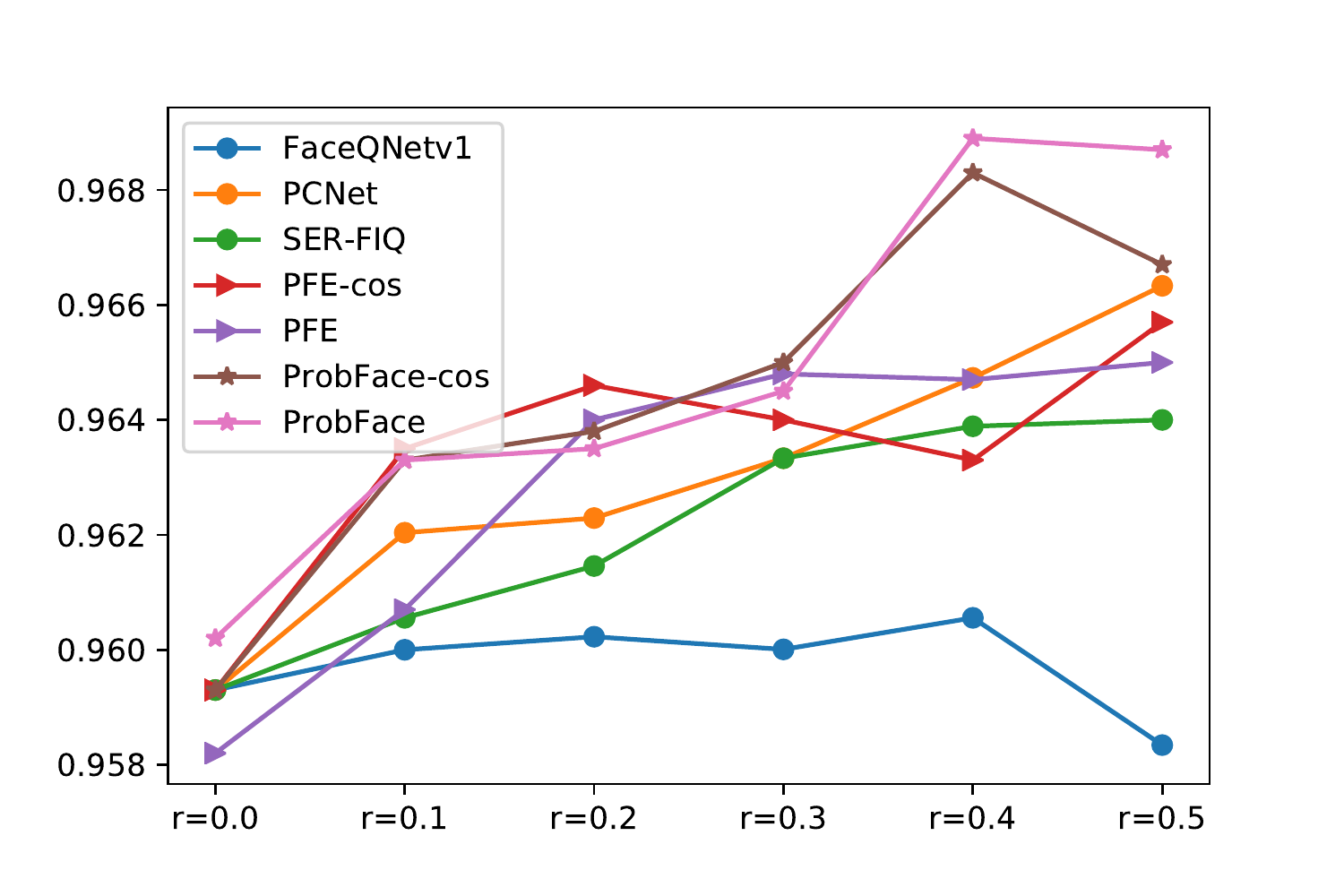}}\hfill
	\subfloat[CFP-FP]{%
		\includegraphics[width=.45\textwidth]{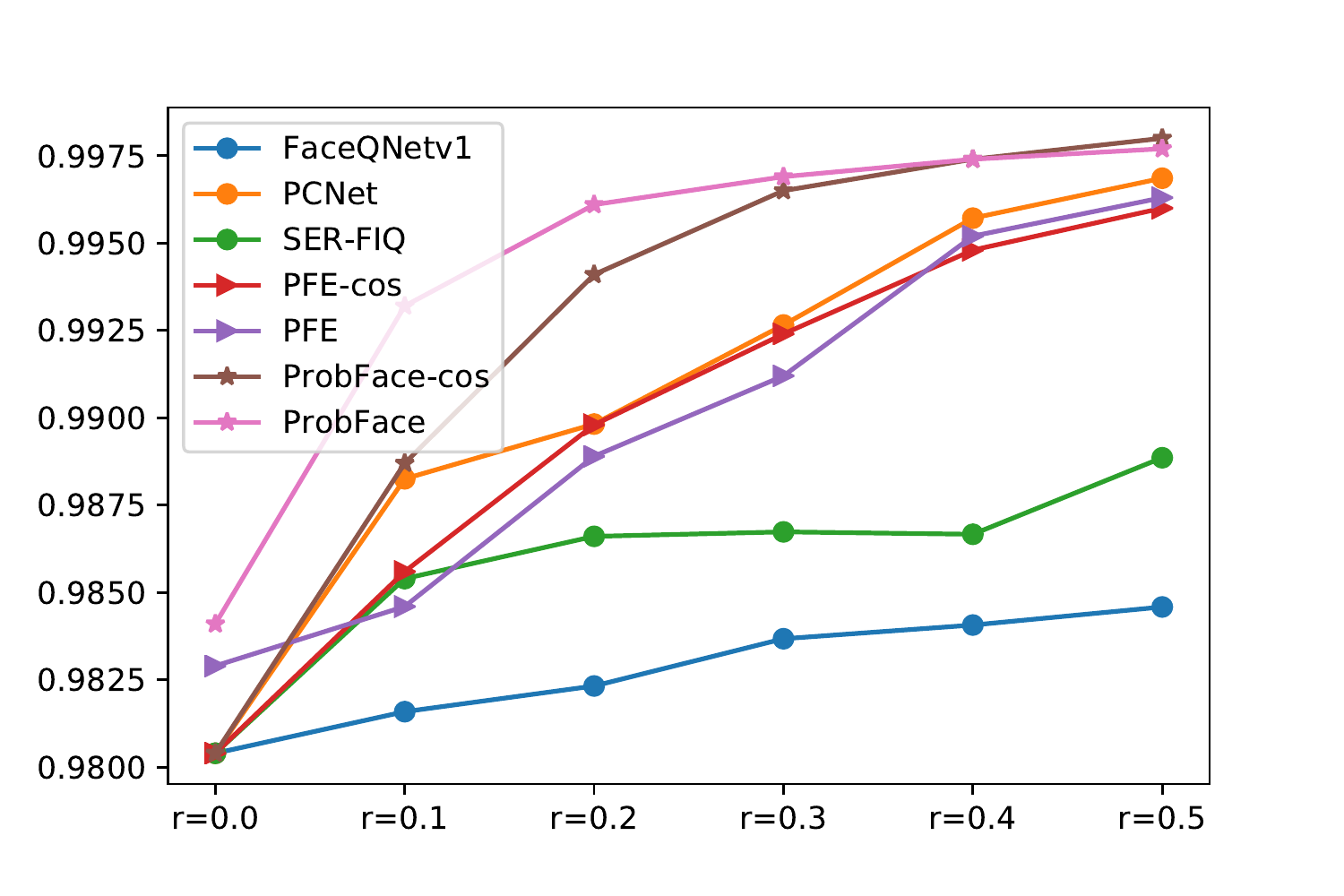}}\hfill
	\subfloat[Vgg2FP]{%
		\includegraphics[width=.45\textwidth]{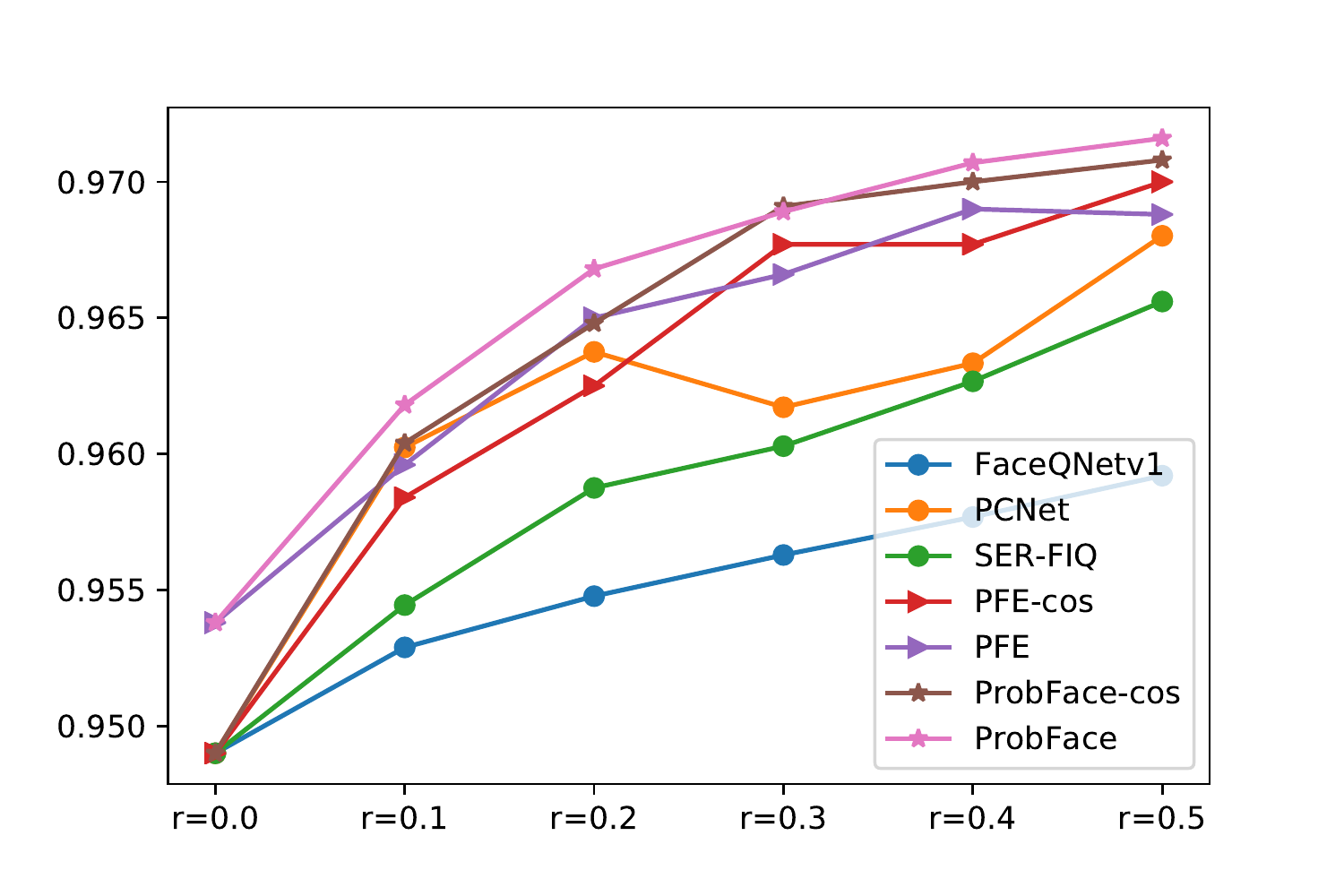}}\hfill
	\caption{~Accuracy-versus-reject curves on LFW, CALFW, CFP-FP and Vgg2FP datasets. The base model is ResFace64. PFE-cos and ProbFace-cos denote the use of cosine metric instead of MLS and FastMLS.}
	\label{fig:RiskResFace100}
\end{figure*}

In a face recognition system with controllable risks, the algorithm will reject the input image if the model is unsure, ensuring that the performance of the system can be controlled when faced with complex recognition scenarios.
Figure \ref{fig:RiskResFace100} shows the accuracy-versus-reject curve based on ResFace64 model and LFW, CALFW, CPF-FP and Vgg2FP datasets.
The curve is obtained by removing the image pairs with highest uncertainty scores dependent on the rejection ratio.
For each image pair ($\mathbf{x}_1, \mathbf{x}_1$), we use the larger of the two uncertainty scores of the images as the uncertainty score of the pair, denoted as $\max(\sigma^2_1, \sigma^2_2)$.
Methods for comparison include FaceQNet \cite{hernandez2019faceqnet}, PCNet \cite{xie2020inducing} and SER-FIQ \cite{terhorst2020ser}.
Among them, FaceQNet and PCNet are performance-based face quality assessments tool which are trained on the dataset with generated quality labels. We use the the v1 version model of FaceQNet in the experiments\footnote{\url{https://github.com/uam-biometrics/FaceQnet}}.
SER-FIQ is based on the test-time dropout of the face recognition model to get the quality score. We inference each image 100 times to get the variance based on ResFace64 model.
As can be seen from the figures, ProbFace and PFE have obtained comparable results on the better-quality LFW and CALFW dataset.
On the large-pose CFP-FP and Vgg2FP dataset, ProbFace achieves the best results.
Experiments show that ProbFace has good performance in predicting face quality and can be applied to risk-controlled face recognition scenarios.
More experimental results can refer to the \ref{sub:appendix_risk}.

\subsection{Visualization}

\begin{figure*}
	\centering
	\includegraphics[width=0.9\textwidth]{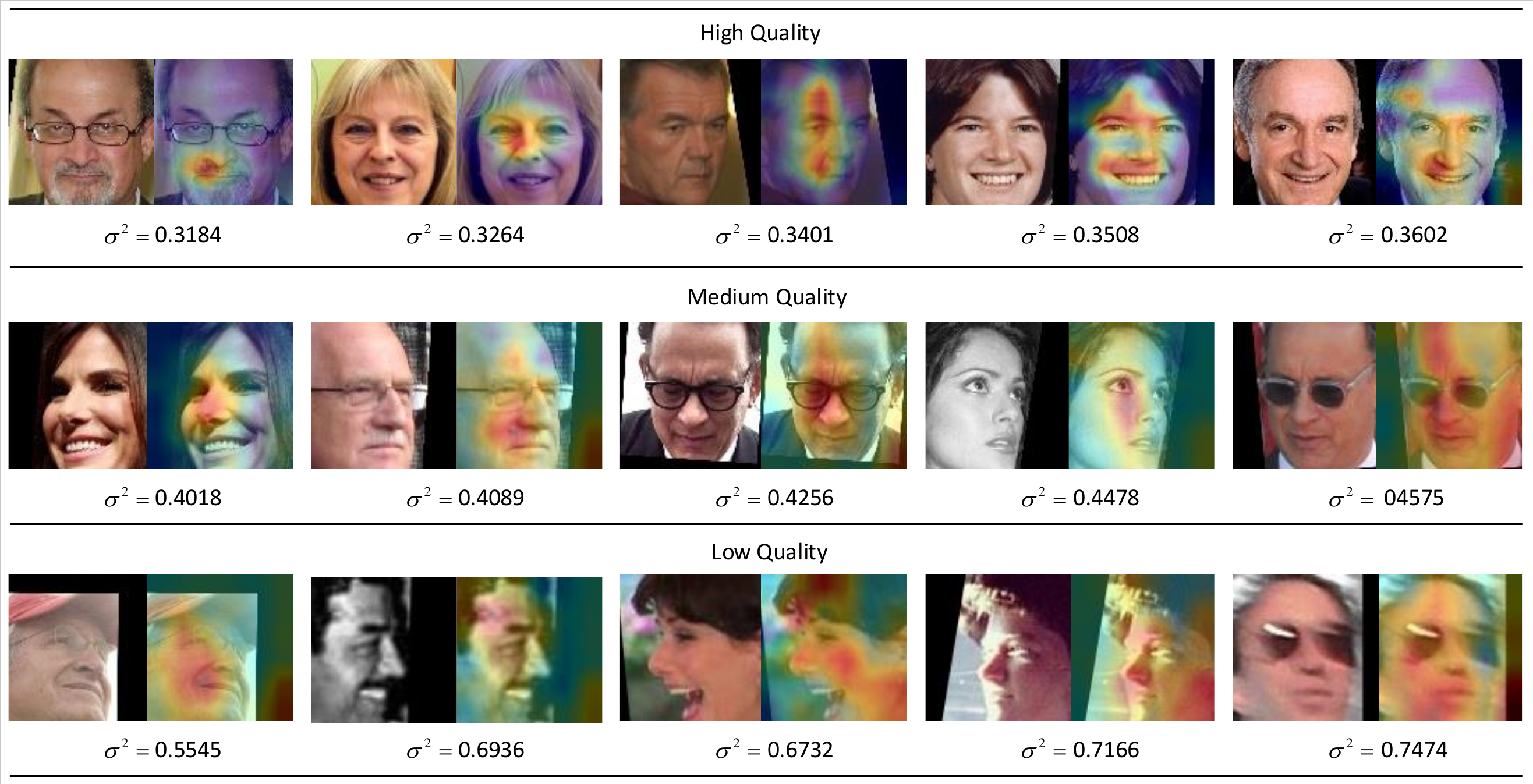}
	\caption{~Grad-CAM visualization of IJB-B dataset}
	\label{fig:GradCam}
\end{figure*}

Figure \ref{fig:GradCam} gives the visualization results of the effect of uncertainty on the feature map base on Grad-CAM \cite{selvaraju2016grad}.
Grad-CAM is designed for classification tasks and cannot be directly used in our uncertainty estimation task. Therefore, we modify the output part of grad-cam and use the gradient of $-\sigma^2$ to perform  back-propagation to obtain a visual image.
In this way, the highlighted parts in the feature map indicate which areas can reduce the uncertainty ($\sigma^2$) of the input face image.
It can be seen from the figure that the activation of high-quality face images is concentrated on the face area, suggesting that clear features can be identified. Low-quality ones are more scattered, suggesting that it is difficult to locate effective information to minimize the uncertainty.

\section{Conclusion}

In this work, we propose a robust probabilistic face embeddings (ProbFace) method to improve the recognition accuracy of PFE in unconstrained environments.
In order to speed up the matching of face pairs, we simplify the calculation of the MLS metric by correcting cosine metric based on uncertainty.
In order to solve the problem that the range of the uncertainty is too large, a constraint term is added to penalize the variance of the uncertainty.
An uncertainty-aware identification loss function is proposed to preserve the identity information by considering both the positive sample pairs and negative sample pairs.
Additionally, we use multi-layer fusion module to use both low-level and high-level features to enhance the ability of uncertainty predictions.
Comprehensive experiments demonstrate that the proposed ProbFace can achieve better and robust performance than PFE in large-pose, large-age and noisy benchmarks and risk-controlled face recognition settings.
In the future, we can further improve the probabilistic face embeddings method for cross-modal face recognition and retrieval, face adversarial attacks, face interpretability, etc.

\section*{Competing Interests}

The authors declare that there is no conflict of interests regarding the publication of this paper.

\section*{Acknowledgement}

This work was mainly supported by Natural Science Foundation of China (61906207,61803376).

\ifCLASSOPTIONcaptionsoff
  \newpage
\fi

\bibliographystyle{elsarticle-num}

\clearpage
\onecolumn

\appendix

\section{Appendices}

\subsection{ROC curves on IJB-B and IJB-C}
\label{sub:appendix_ijbb}

\begin{figure*}[htbp]
	\centering
	\subfloat[IJB-B,ResFace64(0.5)]{%
		\includegraphics[width=.30\textwidth]{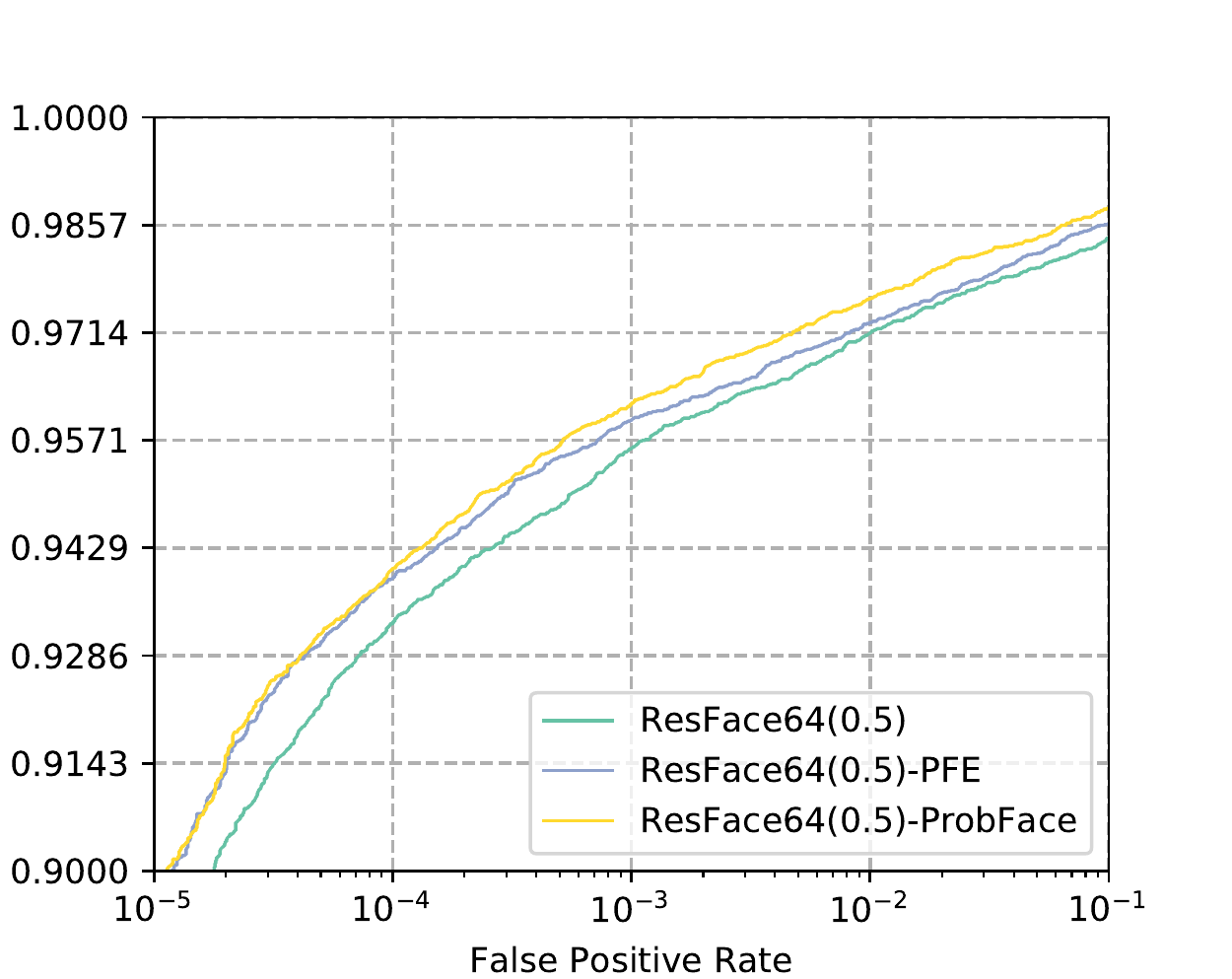}}\hfill
	\subfloat[IJB-B,ResFace64]{%
		\includegraphics[width=.30\textwidth]{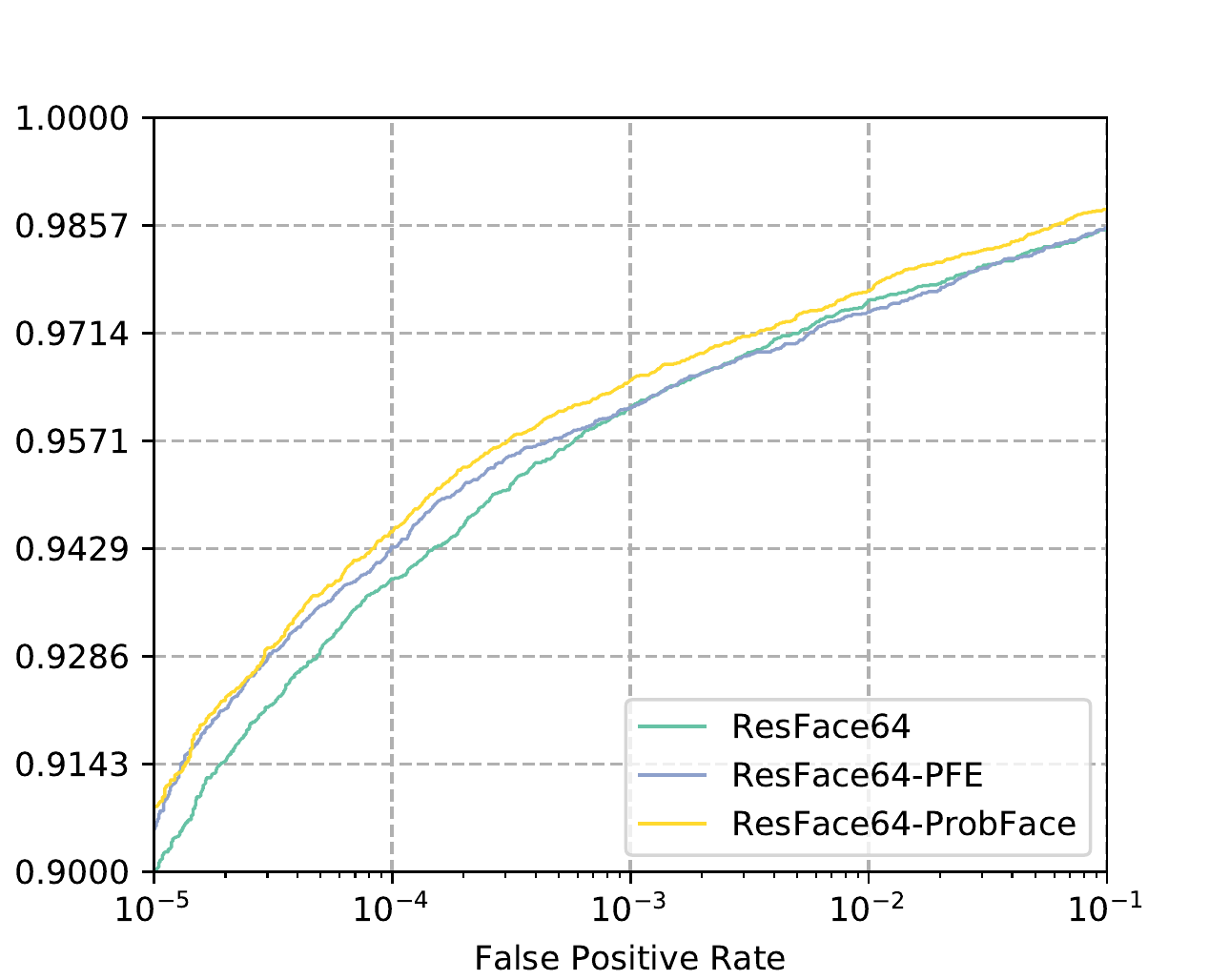}}\hfill
	\subfloat[IJB-B,ResFace100]{%
		\includegraphics[width=.30\textwidth]{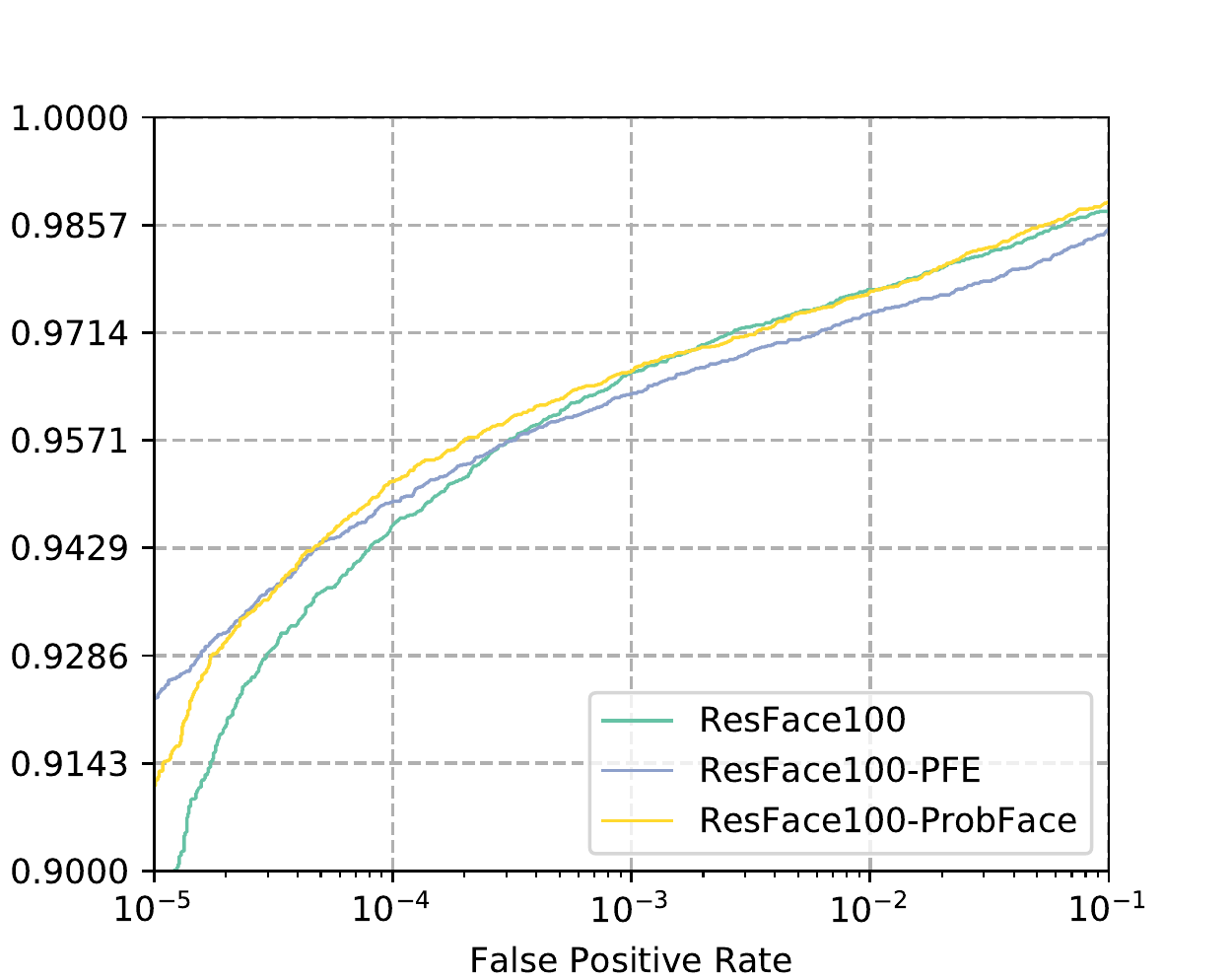}}\hfill
	\subfloat[IJB-C,ResFace64(0.5)]{%
		\includegraphics[width=.30\textwidth]{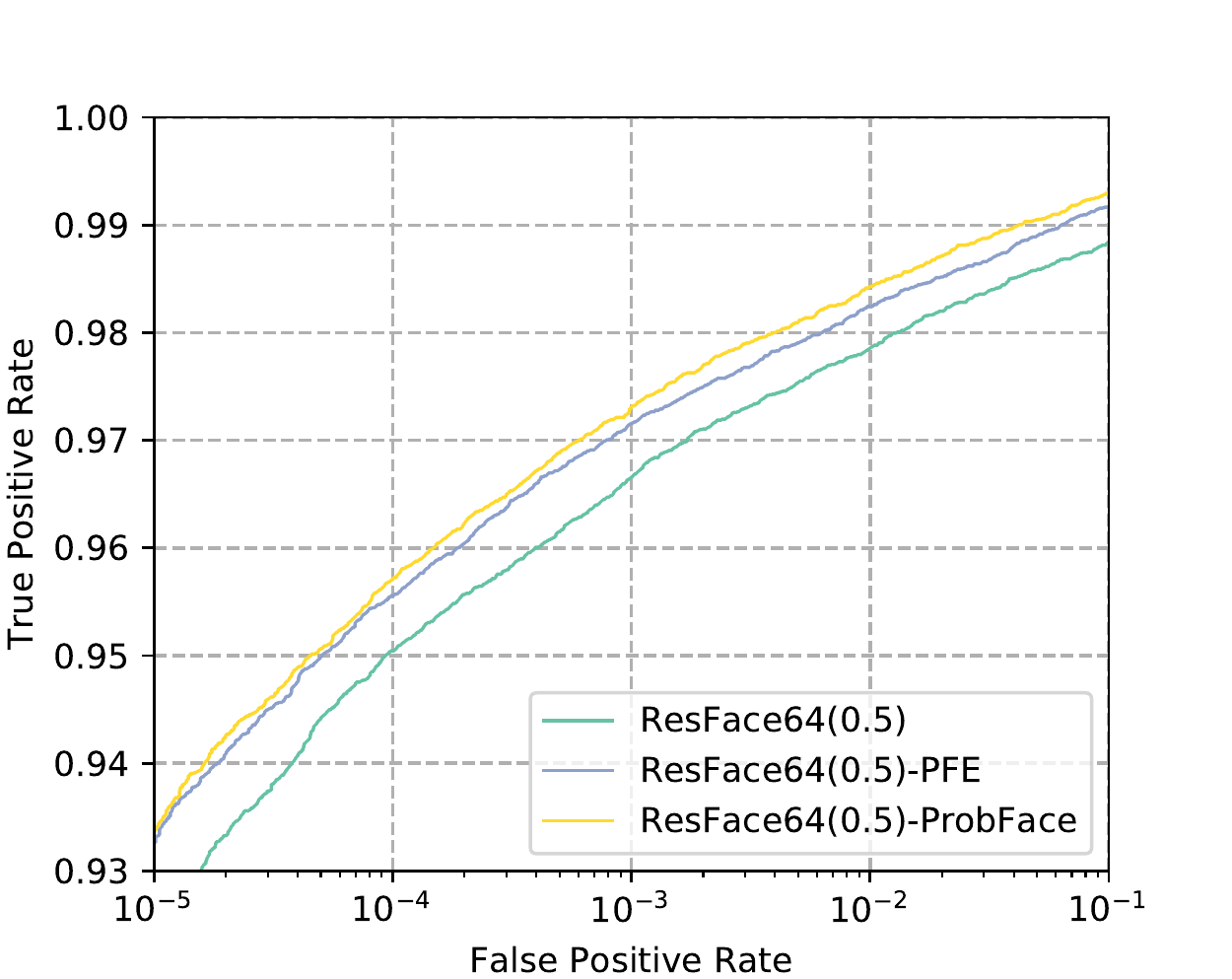}}\hfill
	\subfloat[IJB-C,ResFace64]{%
		\includegraphics[width=.30\textwidth]{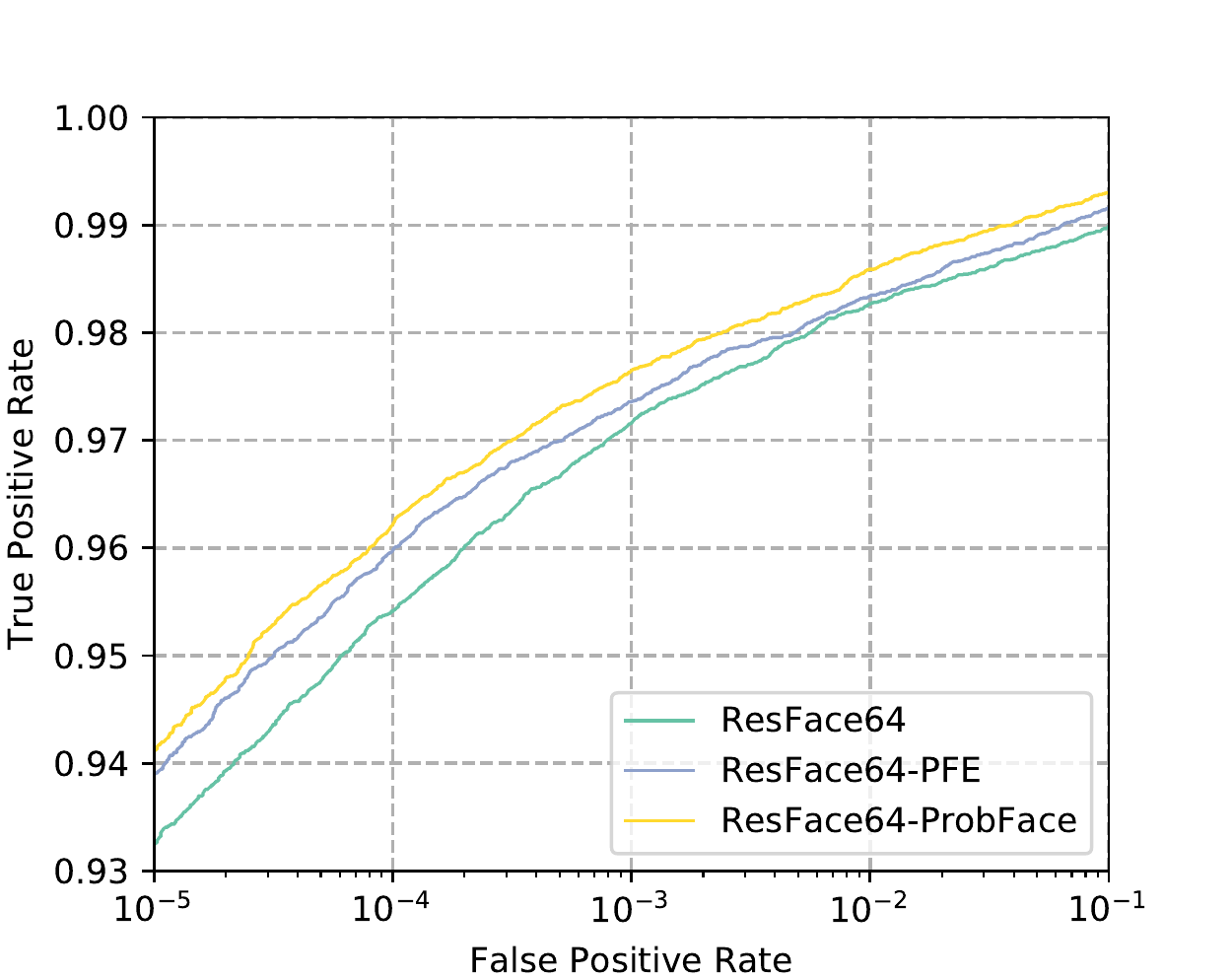}}\hfill
	\subfloat[IJB-C,ResFace100]{%
		\includegraphics[width=.30\textwidth]{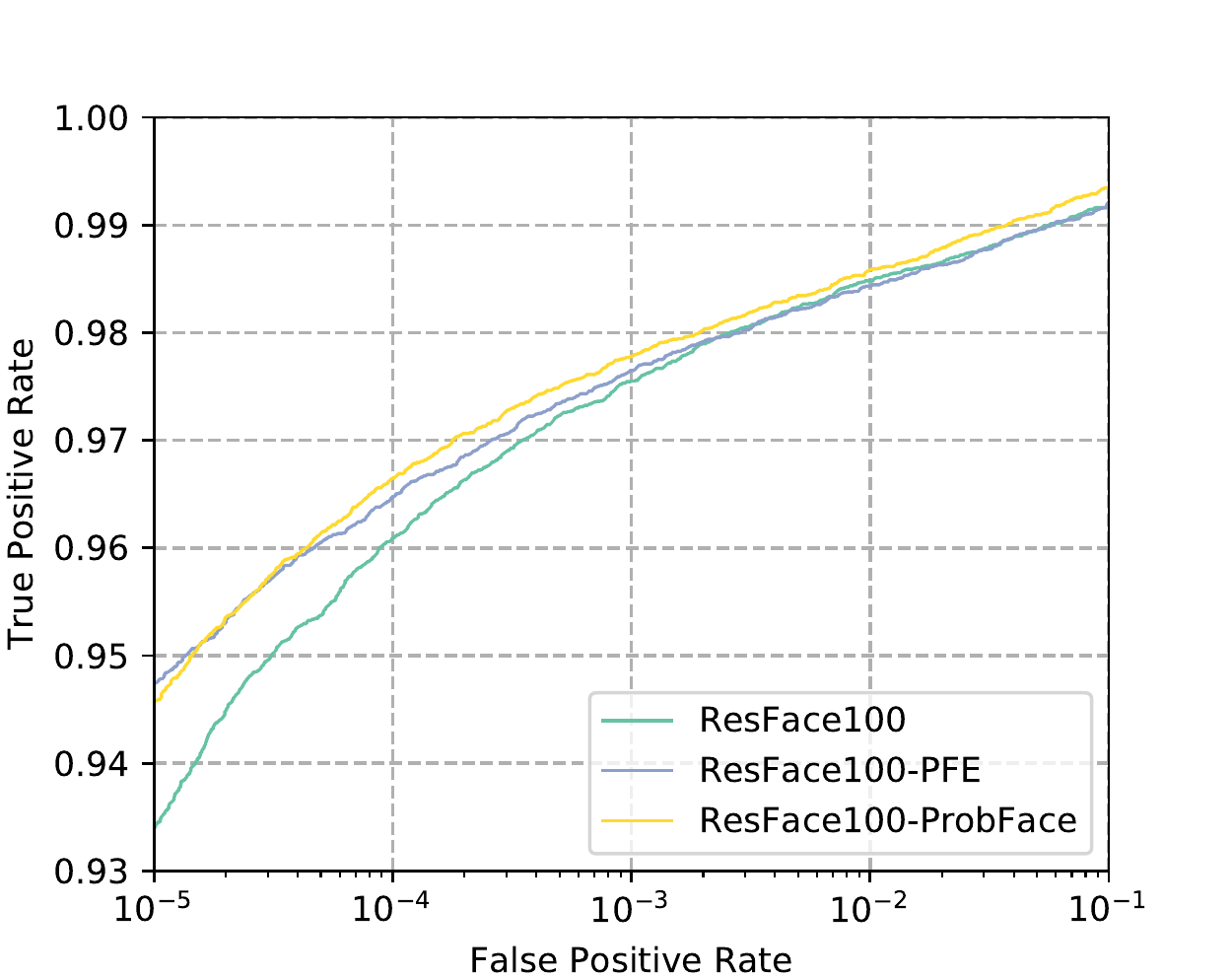}}\hfill
	\caption{~ROC curves of 1:1 verification protocol on the IJB-B and IJB-C. The accuracy of TPR@FAR=0.001 is given in brackets.}
	\label{fig:roc_ijb}
\end{figure*}

Figure \ref{fig:roc_ijb} shows the ROC curves of 1:1 verification protocol of 3 base model on the IJB-B and IJB-C.

\subsection{Performance Comparison on Noisy Datasets}
\label{sub:appendix_noise}

Table \ref{tab:noiseblur} and \ref{tab:noiseocc} give the detailed results of face recognition on datasets with blur and occlusion noise.

\begin{table*}[htbp]
  \centering
  \caption{Performance of face recognition on datasets with blur noise.}
    \resizebox{\textwidth}{!}{\begin{tabular}{l|cccc|cccc|cccc}
    \toprule
          & \multicolumn{4}{c|}{LFW with blur noise} & \multicolumn{4}{c|}{CFP-FP with  blur noise} & \multicolumn{4}{c}{AgeDB30 with  blur noise} \\
    Data Noise Rate & 0.05  & 0.1   & 0.15  & 0.2   & 0.05  & 0.1   & 0.15  & 0.2   & 0.05  & 0.1   & 0.15  & 0.2 \\
    \midrule
    Sphere64-Cosine & 95.22 & 91.37 & 88.03 & 85.10 & 92.74 & 88.00 & 84.80 & 80.87 & 93.03 & 88.83 & 84.97 & 80.08 \\
    Sphere64-PFE & 96.93 & 93.60 & 91.10 & 88.35 & 93.61 & 89.21 & 85.83 & 82.67 & 93.88 & 90.47 & 87.55 & 83.80 \\
    LResNet100E-IR,ArcFace & 95.50 & 92.13 & 89.58 & 86.77 & 94.41 & 85.62 & 89.36 & 82.30 & 93.53 & 89.47 & 85.17 & 80.66 \\
    \midrule
    ResFace64s2 & 96.63 & 94.40 & 92.15 & 90.35 & 93.61 & 89.93 & 87.13 & 83.64 & 93.20 & 89.98 & 86.15 & 82.48 \\
    ResFace64s2-PFE & \textbf{98.10} & 96.18 & 94.22 & 92.02 & 94.89 & 91.83 & \textbf{88.76} & 85.77 & \textbf{94.75} & 92.17 & \textbf{89.28} & 85.92 \\
    ResFace64s2-ProbFace & 98.02 & \textbf{96.32} & \textbf{94.32} & \textbf{92.50} & \textbf{94.90} & \textbf{92.30} & 88.53 & \textbf{86.16} & 94.60 & \textbf{92.23} & 89.20 & \textbf{85.95} \\
    \midrule
    ResFace64 & 96.95 & 94.47 & 92.72 & 91.77 & 93.99 & 90.16 & 87.74 & 84.43 & 93.42 & 89.90 & 86.77 & 83.05 \\
    ResFace64-PFE & 97.85 & 95.85 & 94.28 & 92.70 & 94.80 & 91.57 & 88.77 & 85.26 & 95.02 & 92.22 & 89.85 & 86.33 \\
    ResFace64-ProbFace & \textbf{98.30} & \textbf{96.42} & \textbf{94.78} & \textbf{92.85} & \textbf{94.99} & \textbf{91.99} & \textbf{89.20} & \textbf{85.73} & \textbf{95.15} & \textbf{92.45} & \textbf{90.50} & \textbf{86.75} \\
    \midrule
    ResFace100 & 97.27 & 95.45 & 93.68 & 92.20 & 95.09 & 91.31 & 88.69 & 85.41 & 93.68 & 90.07 & 87.50 & 84.33 \\
    ResFace100-PFE & 98.28 & 96.62 & 94.97 & 92.93 & 95.50 & 92.34 & 89.37 & 86.00 & 95.78 & 92.87 & 90.13 & 87.23 \\
    ResFace100-ProbFace & \textbf{98.33} & \textbf{96.70} & \textbf{95.18} & \textbf{93.03} & \textbf{95.64} & \textbf{92.43} & \textbf{89.89} & \textbf{86.41} & \textbf{95.97} & \textbf{93.10} & \textbf{90.48} & \textbf{87.72} \\
    \bottomrule
    \end{tabular}}
  \label{tab:noiseblur}%
\end{table*}%

\begin{table*}[htbp]
  \centering
  \caption{Performance of face recognition on datasets with occlusion noise.}
    \resizebox{\textwidth}{!}{\begin{tabular}{l|cccc|cccc|cccc}
    \toprule
          & \multicolumn{4}{c|}{LFW with occlusion noise} & \multicolumn{4}{c|}{CFP-FP with  occlusion noise} & \multicolumn{4}{c}{AgeDB30 with  occlusion noise} \\
    Data Noise Rate & 0.05  & 0.1   & 0.15  & 0.2   & 0.05  & 0.1   & 0.15  & 0.2   & 0.05  & 0.1   & 0.15  & 0.2 \\
    \midrule
    Sphere64-Cosine & 98.82 & 98.07 & 97.60 & 97.20 & 94.67 & 93.53 & 92.67 & 91.44 & 96.08 & 94.80 & 93.48 & 92.48 \\
    Sphere64-PFE & 99.37 & 98.97 & 98.38 & 98.12 & 95.59 & 93.96 & 93.33 & 92.57 & 96.15 & 95.28 & 94.57 & 93.65 \\
    LResNet100E-IR,ArcFace & 99.47 & 98.93 & 98.82 & 98.60 & 97.09 & 96.00 & 95.56 & 94.53 & 97.02 & 96.18 & 94.83 & 94.75 \\
    \midrule
    Resface64s2 & 99.48 & 99.15 & 98.73 & 98.67 & 96.01 & 95.14 & 94.61 & 93.99 & 96.73 & 96.00 & 94.62 & 94.05 \\
    Resface64s2-PFE & 99.60 & \textbf{99.45} & 99.37 & 99.20 & 96.51 & 95.87 & 95.40 & 94.46 & 96.95 & 96.43 & 95.57 & 95.43 \\
    Resface64s2-ProbFace & \textbf{99.62} & \textbf{99.45} & \textbf{99.40} & \textbf{99.35} & \textbf{96.53} & \textbf{95.93} & \textbf{95.59} & \textbf{94.54} & \textbf{97.25} & \textbf{96.47} & \textbf{95.63} & \textbf{95.63} \\
    \midrule
    Resface64 & 99.37 & 99.18 & 99.07 & 98.82 & 96.39 & 95.69 & 94.99 & 94.33 & 96.92 & 96.35 & 94.92 & 94.70 \\
    Resface64-PFE & 99.65 & 99.48 & \textbf{99.40} & 99.25 & 97.03 & 95.94 & 95.71 & 94.73 & 96.70 & 96.55 & 95.82 & \textbf{95.80} \\
    Resface64-ProbFace & \textbf{99.68} & \textbf{99.58} & 99.38 & \textbf{99.37} & \textbf{97.14} & \textbf{96.50} & \textbf{96.01} & \textbf{95.09} & \textbf{97.28} & \textbf{96.70} & \textbf{96.35} & 95.78 \\
    \midrule
    Resface100 & 99.52 & 99.20 & 99.13 & 98.97 & 97.40 & 96.74 & 96.20 & 95.26 & 96.98 & 96.38 & 95.48 & 95.53 \\
    Resface100-PFE & 99.61 & 99.57 & 99.43 & 99.37 & 97.52 & 96.80 & 96.63 & 95.60 & \textbf{97.65} & 96.85 & \textbf{96.83} & 95.82 \\
    Resface100-ProbFace & \textbf{99.67} & \textbf{99.62} & \textbf{99.55} & \textbf{99.48} & \textbf{97.79} & \textbf{97.04} & \textbf{96.73} & \textbf{95.77} & \textbf{97.65} & \textbf{96.92} & \textbf{96.83} & \textbf{95.97} \\
    \bottomrule
    \end{tabular}}
  \label{tab:noiseocc}%
\end{table*}%

\subsection{Additional Results on Risk-Controlled Face Recognition}
\label{sub:appendix_risk}

Table \ref{tab:risk-resface64} and \ref{tab:risk-resface64s2} report the results of 1:1 verification with rejection for CFP-FP and Vgg2FP based on ResFace64 and ResFace64(0.2) base model respectively.
In the experiment, we use two methods to calculate the filtering score:
1) \emph{Add}: $\sigma_1^2 + \sigma_2^2$;
2) \emph{Max}: $\max(\sigma_1^2, \sigma_2^2)$.
Among them, $\sigma_1$ and $\sigma_2$ are the uncertainties of the images to be compared.
We remove the image pairs with higher uncertainty scores dependent on the rejection ratio.
As can be seen from the table, ProbFace can achieve better results than PFE.
In most experimental results, the results of the Add method are slightly better than the Max method, since the optimization in the MLS loss is the addition of the two uncertainties.
In actual deployment, it is more convenient to use the Max method. You can directly use the score of a single image to select whether to reject, as opposed to the Add method, which requires two images to determine the rejection score.
In addition, it can be seen from the results that the recognition accuracy of MLS is still higher than that of Cosine under risk-controlled settings.
As the low-quality face images are gradually filtered out, the gap between the two gradually decreases.

\begin{table*}[htbp]
  \footnotesize
	\centering
	\caption{Results of 1:1 verification with rejection on CFP-FP and Vgg2FP based on ResFace64 model.}
	\begin{tabular}{ccclrrrrrr}
		\toprule
		\multicolumn{1}{l}{Dataset} & Method & Match Type & Filter Type & \multicolumn{1}{l}{r=0.0} & \multicolumn{1}{l}{r=0.1} & \multicolumn{1}{l}{r=0.2} & \multicolumn{1}{l}{r=0.3} & \multicolumn{1}{l}{r=0.4} & \multicolumn{1}{l}{r=0.5} \\
		\midrule
		\multirow{8}[4]{*}{CFP-FP} & \multirow{4}[2]{*}{PFE} & \multirow{2}[1]{*}{Cosine} & Add   & 98.04 & 98.51 & 98.96 & 99.12 & 99.60 & 99.57 \\
		&       &       & Max   & 98.04 & 98.56 & 98.98 & 99.24 & 99.48 & 99.60 \\
		&       & \multirow{2}[1]{*}{MLS} & Add   & 98.28 & 98.44 & 98.96 & 99.22 & 99.45 & 99.51 \\
		&       &       & Max   & 98.28 & 98.46 & 98.89 & 99.12 & 99.52 & 99.63 \\
		\cmidrule{2-10}          & \multirow{4}[2]{*}{ProbFace} & \multirow{2}[1]{*}{Cosine} & Add   & 98.04 & 99.05 & 99.34 & 99.61 & 99.79 & 99.83 \\
		&       &       & Max   & 98.04 & 98.87 & 99.41 & 99.65 & 99.74 & 99.80 \\
		&       & \multirow{2}[1]{*}{FastMLS} & Add   & \textbf{98.41} & \textbf{99.33} & 99.50 & 99.63 & \textbf{99.81} & \textbf{99.83} \\
		&       &       & Max   & \textbf{98.41} & 99.32 & \textbf{99.61} & \textbf{99.69} & 99.74 & 99.77 \\
		\midrule
		\multirow{8}[4]{*}{Vgg2FP} & \multirow{4}[2]{*}{PFE} & \multirow{2}[1]{*}{Cosine} & Add   & 94.94 & 95.84 & 96.25 & 96.77 & 96.77 & 97.00 \\
		&       &       & Max   & 94.94 & 95.76 & 95.98 & 96.40 & 96.63 & 96.72 \\
		&       & \multirow{2}[1]{*}{MLS} & Add   & 95.38 & 95.96 & 96.50 & 96.66 & 96.90 & 96.88 \\
		&       &       & Max   & 95.38 & 95.98 & 95.88 & 96.37 & 96.63 & 96.72 \\
		\cmidrule{2-10}          & \multirow{4}[2]{*}{ProbFace} & \multirow{2}[1]{*}{Cosine} & Add   & 94.94 & 96.04 & 96.48 & 96.91 & 97.00 & 97.08 \\
		&       &       & Max   & 94.94 & 95.95 & 96.08 & 96.57 & 96.70 & 96.56 \\
		&       & \multirow{2}[1]{*}{FastMLS} & Add   & \textbf{95.38} & 96.18 & \textbf{96.68} & \textbf{96.89} & \textbf{97.07} & \textbf{97.16} \\
		&       &       & Max   & \textbf{95.38} & \textbf{96.22} & 96.35 & 96.66 & 96.73 & 96.80 \\
		\bottomrule
	\end{tabular}
	\label{tab:risk-resface64}%
\end{table*}%

\begin{table*}[htbp]
  \footnotesize
	\centering
	\caption{Results of 1:1 verification with rejection on CFP-FP and Vgg2FP based on ResFace64(0.5) model}
	\begin{tabular}{ccclrrrrrr}
		\toprule
		\multicolumn{1}{l}{Dataset} & Method & Match Type & Filter Type & \multicolumn{1}{l}{r=0.0} & \multicolumn{1}{l}{r=0.1} & \multicolumn{1}{l}{r=0.2} & \multicolumn{1}{l}{r=0.3} & \multicolumn{1}{l}{r=0.4} & \multicolumn{1}{l}{r=0.5} \\
		\midrule
		\multirow{8}[4]{*}{CFP-FP} & \multirow{4}[2]{*}{PFE} & \multirow{2}[1]{*}{Cosine} & Add   & 97.50 & 98.35 & 98.68 & 98.98 & 99.38 & 99.51 \\
		&       &       & Max   & 97.50 & 98.21 & 98.79 & 99.12 & 99.40 & 99.51 \\
		&       & \multirow{2}[1]{*}{MLS} & Add   & 97.90 & 98.41 & 98.84 & 98.94 & 99.36 & 99.49 \\
		&       &       & Max   & 97.90 & 98.40 & 98.96 & 99.24 & 99.48 & 99.51 \\
		\cmidrule{2-10}          & \multirow{4}[2]{*}{ProbFace} & \multirow{2}[1]{*}{Cosine} & Add   & 97.50 & 98.76 & 99.02 & 99.29 & 99.48 & 99.49 \\
		&       &       & Max   & 97.50 & 98.60 & 99.00 & \textbf{99.33} & 99.48 & 99.54 \\
		&       & \multirow{2}[1]{*}{FastMLS} & Add   & \textbf{97.93} & \textbf{99.06} & \textbf{99.25} & 99.29 & 99.43 & 99.46 \\
		&       &       & Max   & \textbf{97.93} & 99.00 & \textbf{99.25} & \textbf{99.33} & \textbf{99.50} & \textbf{99.57} \\
		\midrule
		\multirow{8}[4]{*}{Vgg2FP} & \multirow{4}[2]{*}{PFE} & \multirow{2}[1]{*}{Cosine} & Add   & 94.24 & 95.29 & 95.80 & 96.29 & 96.67 & 97.00 \\
		&       &       & Max   & 94.24 & 95.09 & 95.60 & 96.23 & 96.90 & 96.88 \\
		&       & \multirow{2}[1]{*}{MLS} & Add   & 94.84 & 95.64 & 96.13 & 96.46 & 96.93 & 97.04 \\
		&       &       & Max   & 94.84 & 95.29 & 95.80 & 96.29 & 96.67 & 97.00 \\
		\cmidrule{2-10}          & \multirow{4}[2]{*}{ProbFace} & \multirow{2}[1]{*}{Cosine} & Add   & 94.24 & 95.42 & 96.05 & 96.43 & 96.53 & 96.88 \\
		&       &       & Max   & 94.24 & 95.24 & 95.73 & 96.20 & 96.73 & 96.68 \\
		&       & \multirow{2}[1]{*}{FastMLS} & Add   & \textbf{95.00} & \textbf{96.22} & \textbf{96.38} & \textbf{96.86} & \textbf{97.03} & \textbf{97.08} \\
		&       &       & Max   & \textbf{95.00} & 95.87 & 96.18 & 96.37 & 96.97 & 96.88 \\
		\bottomrule
	\end{tabular}
	\label{tab:risk-resface64s2}%
\end{table*}

\subsection{Visualization}
\label{sub:appendix_visu}

\begin{figure*}
	\centering
	\includegraphics[width=0.99\textwidth]{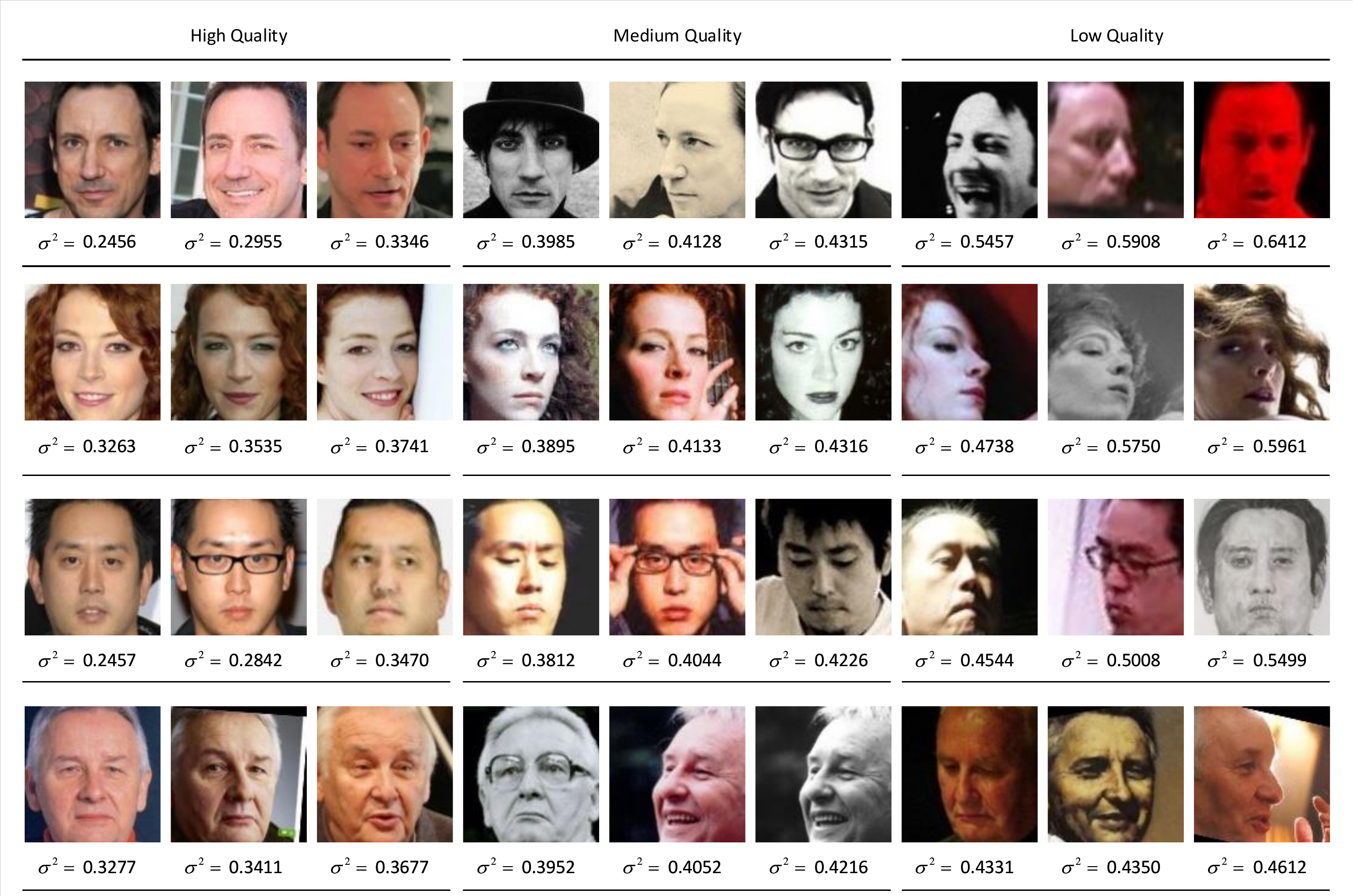}
	\caption{~Examples visualization of 4 individuals in MS-Celeb-1M dataset}
	\label{fig:ms1mexamples}
\end{figure*}

Figure \ref{fig:ms1mexamples} displays the data of 4 individuals in the MS-Celeb-1M dataset.
The images are sorted in ascending order of uncertainty from left to right. It can be seen from the figure that good-quality images are close to frontal images, medium-quality pictures have more angles and changes, and poor-quality images have the largest changes and are harder to distinguish.

\subsection{Distribution of the MLS score}

\begin{figure*}
	\centering
	\subfloat[cosine (99.80\%)]{%
		 \includegraphics[width=.33\textwidth]{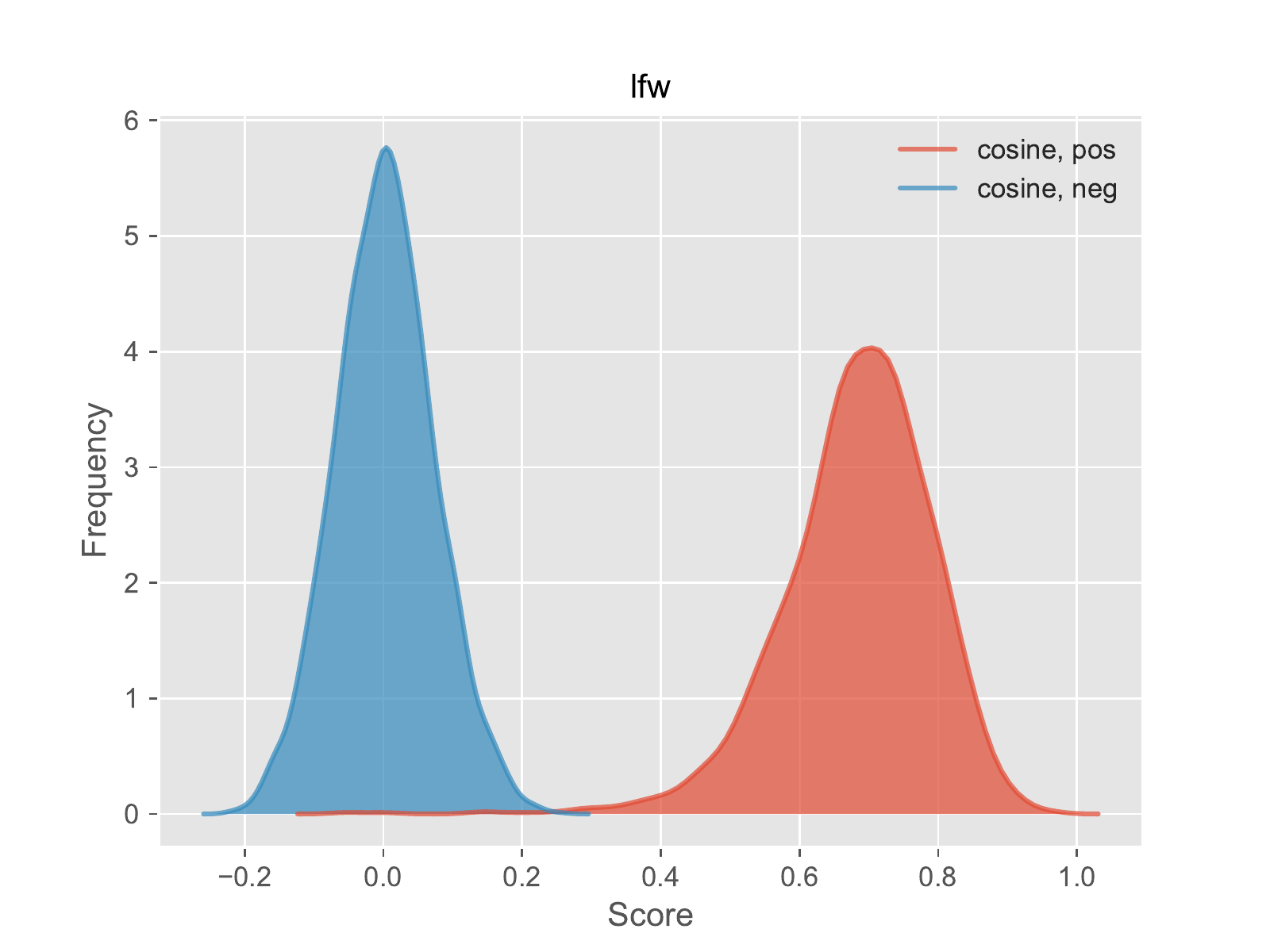}}\hfill
	\subfloat[MLS of PFE (99.82\%)]{%
		 \includegraphics[width=.33\textwidth]{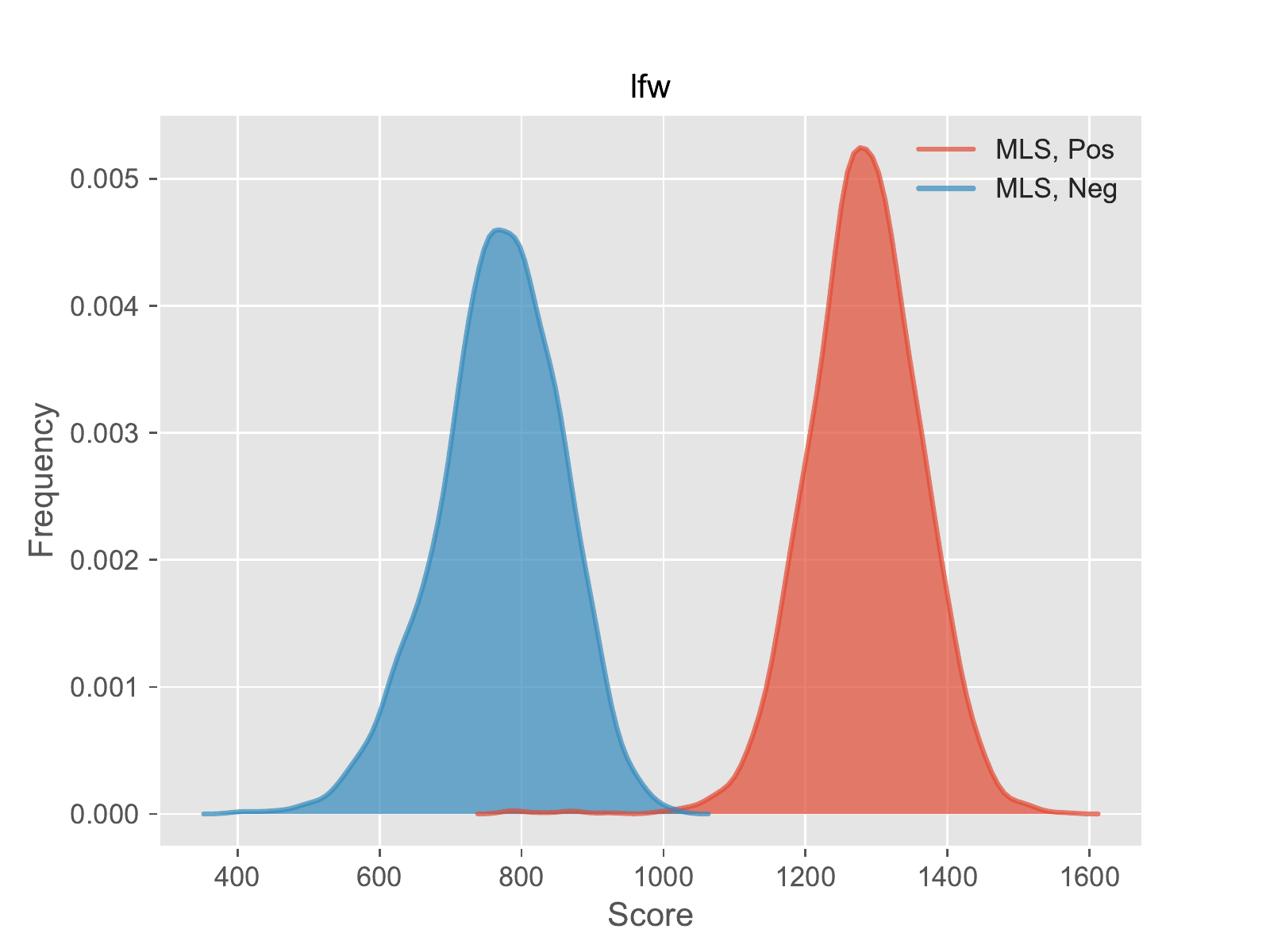}}\hfill
	\subfloat[FastMLS of ProbFace (99.85\%)]{%
		 \includegraphics[width=.33\textwidth]{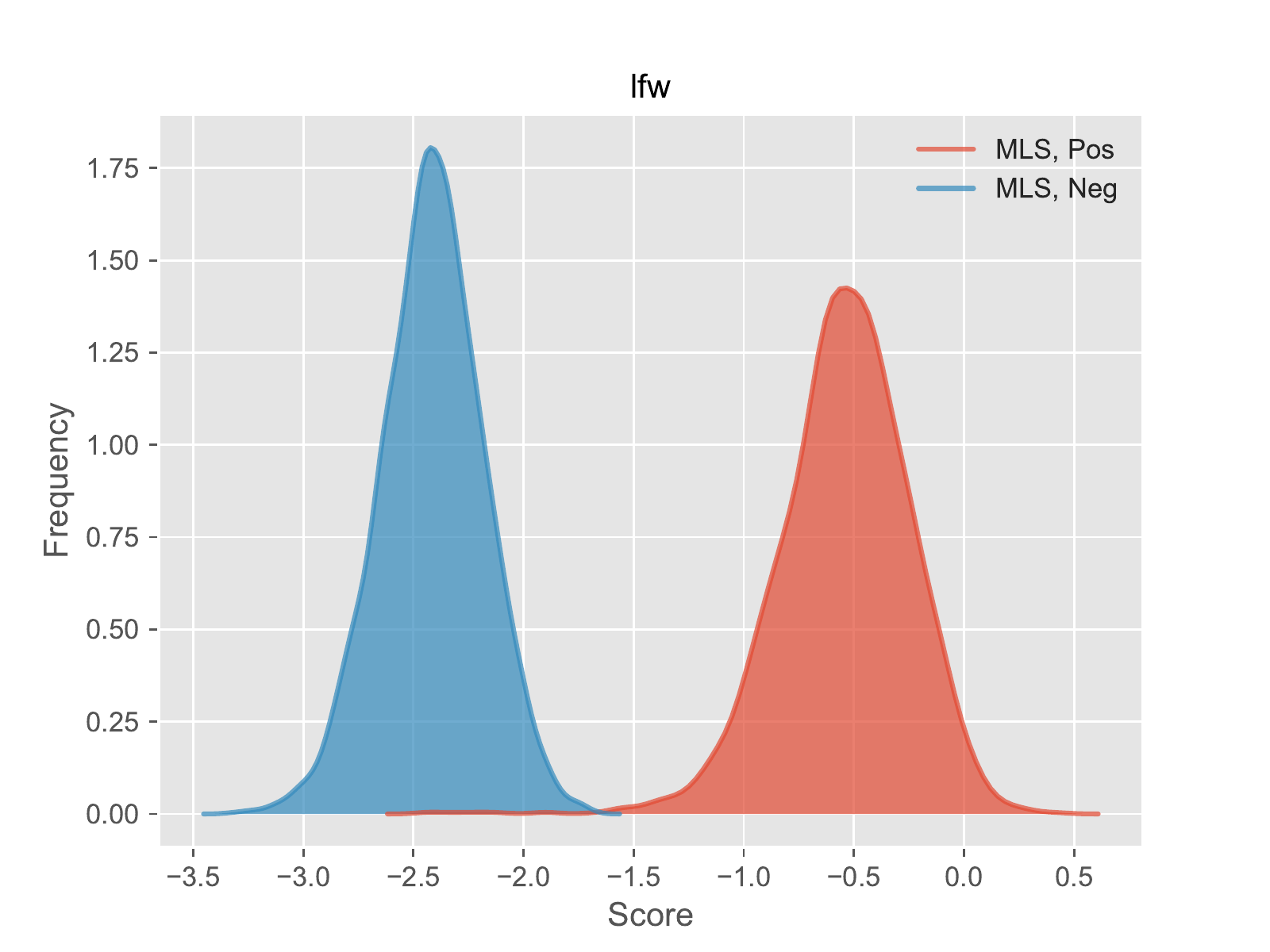}}\hfill
	\caption{~Distribution of the cosine and MLS score on LFW datasets.}
	\label{fig:HistMLSDatasetsLFW}
\end{figure*}

\begin{figure*}
	\centering
	\subfloat[cosine (95.93\%)]{%
		 \includegraphics[width=.33\textwidth]{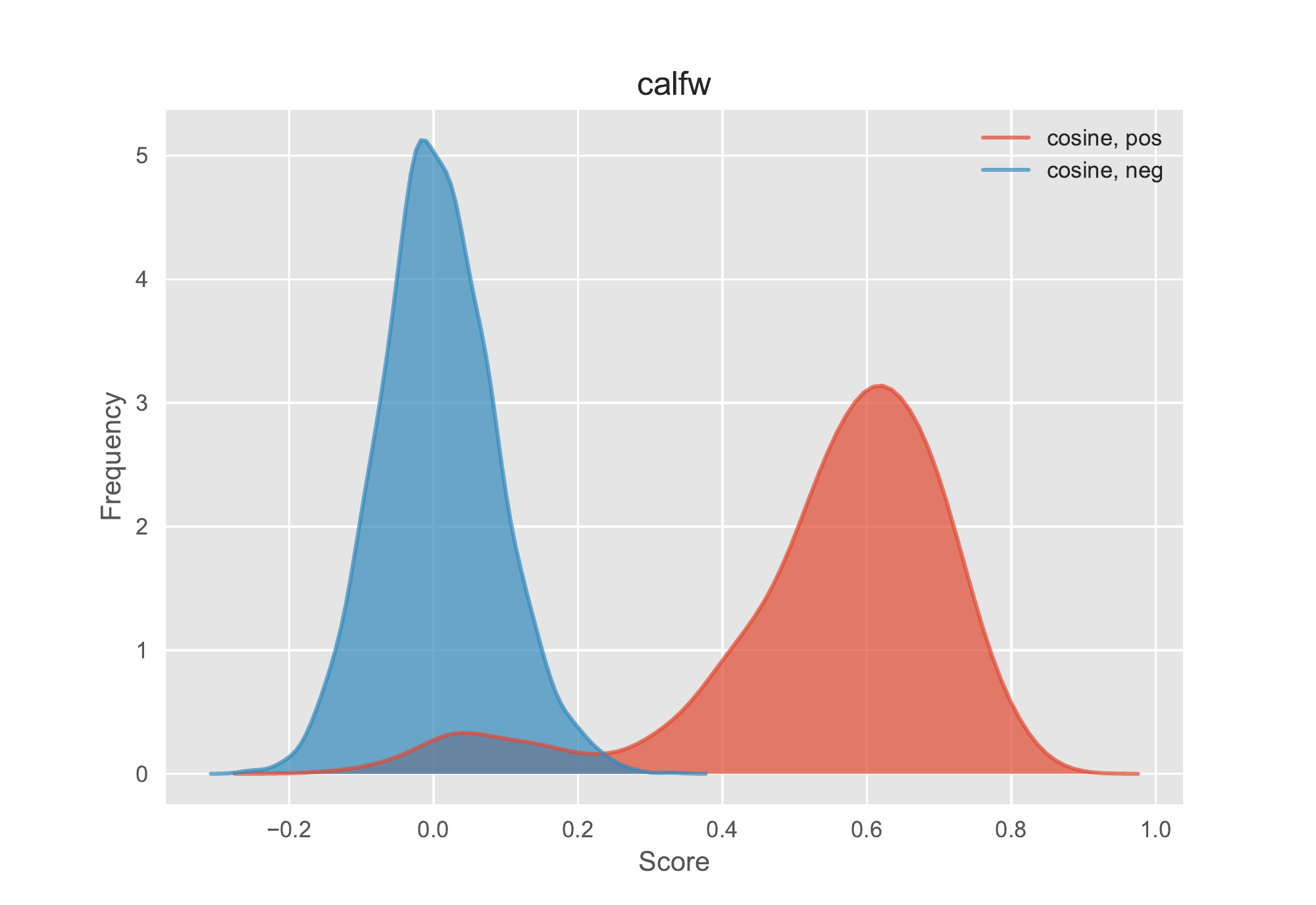}}\hfill
	\subfloat[MLS of PFE (95.85\%)]{%
		 \includegraphics[width=.33\textwidth]{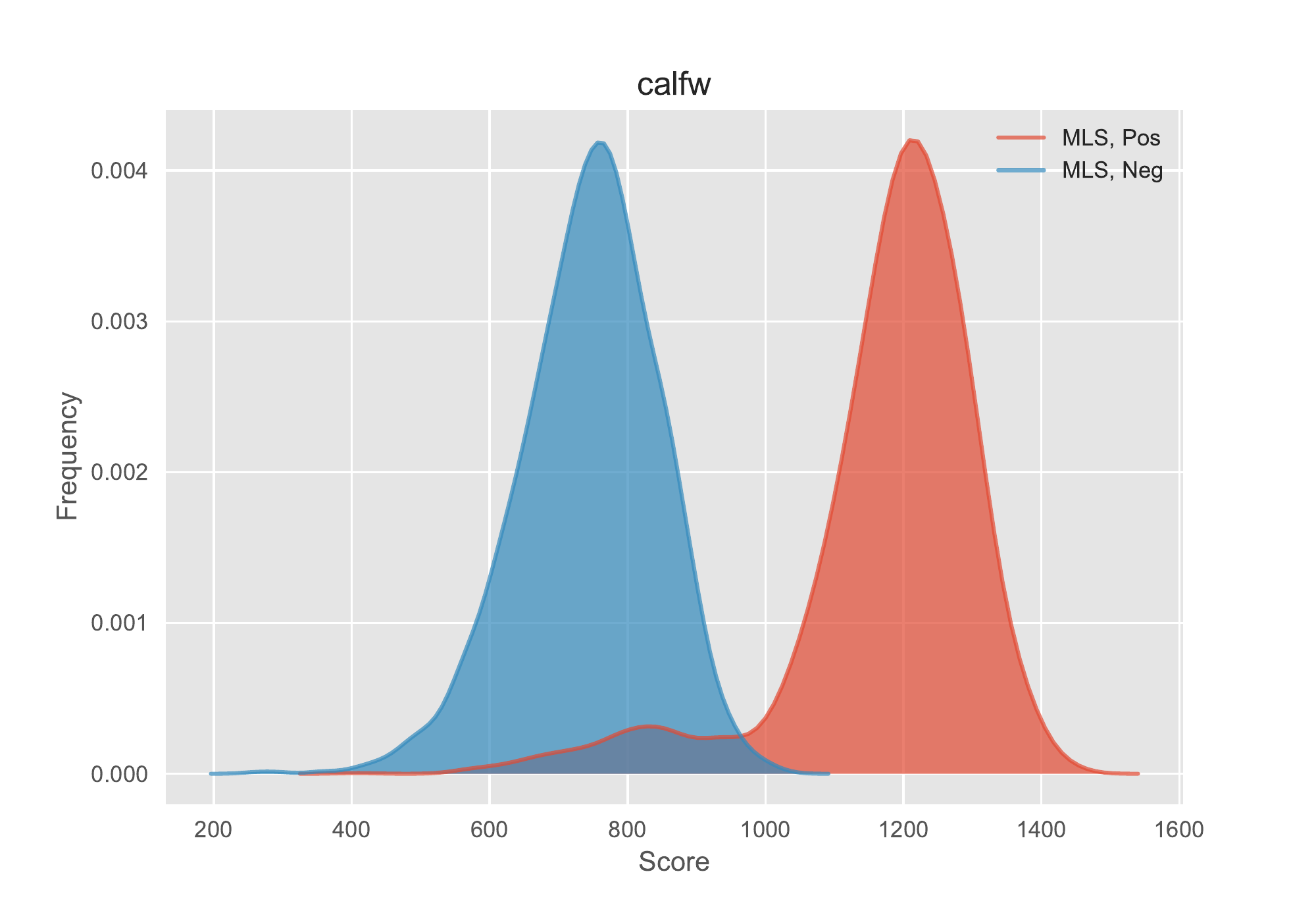}}\hfill
	\subfloat[FastMLS of ProbFace (99.85\%)]{%
		 \includegraphics[width=.33\textwidth]{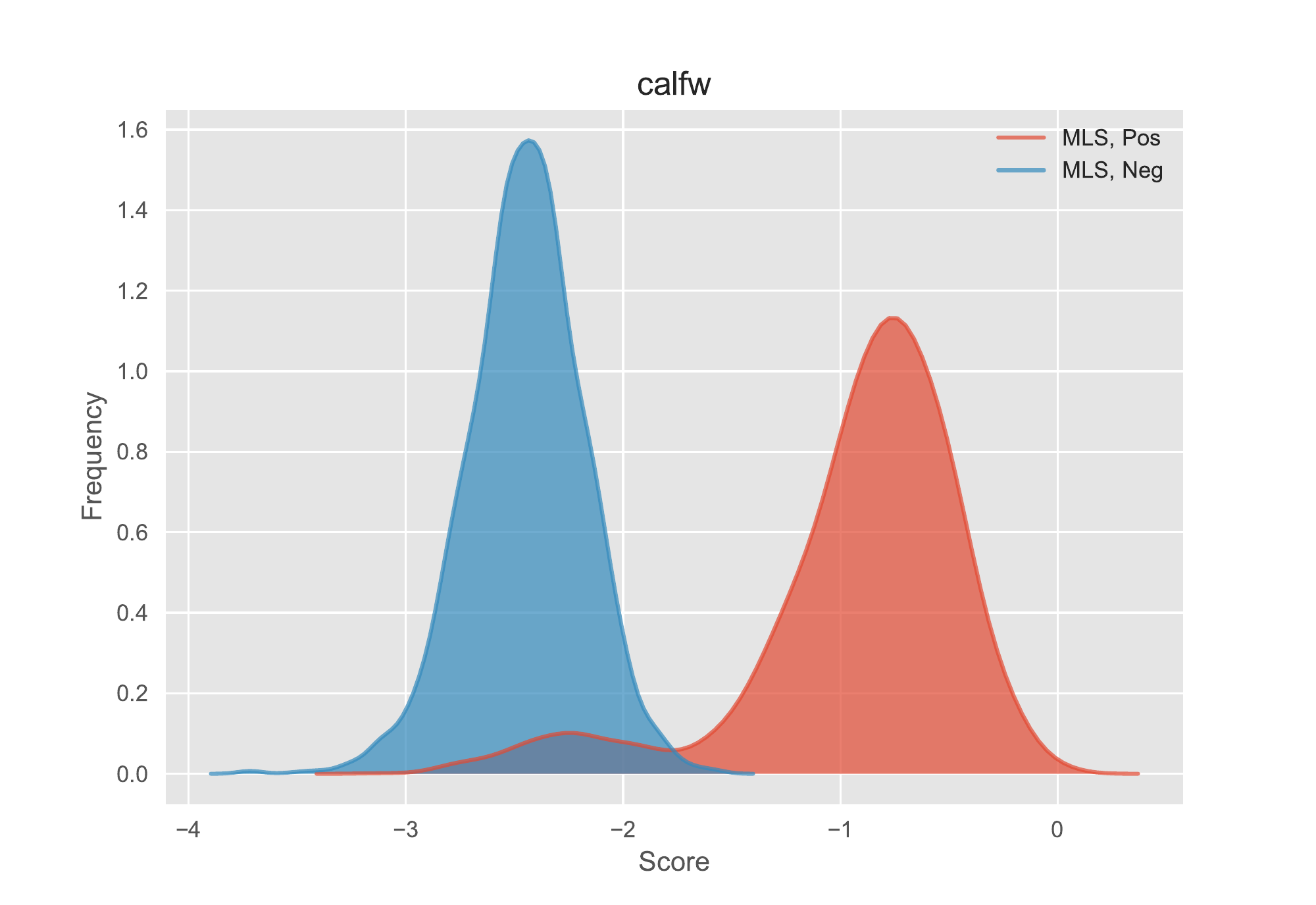}}\hfill
	\caption{~Distribution of the cosine and MLS score on CALFW datasets.}
	\label{fig:HistMLSDatasetsCALFW}
\end{figure*}

\begin{figure*}
	\centering
	\subfloat[cosine (92.53\%)]{%
		 \includegraphics[width=.33\textwidth]{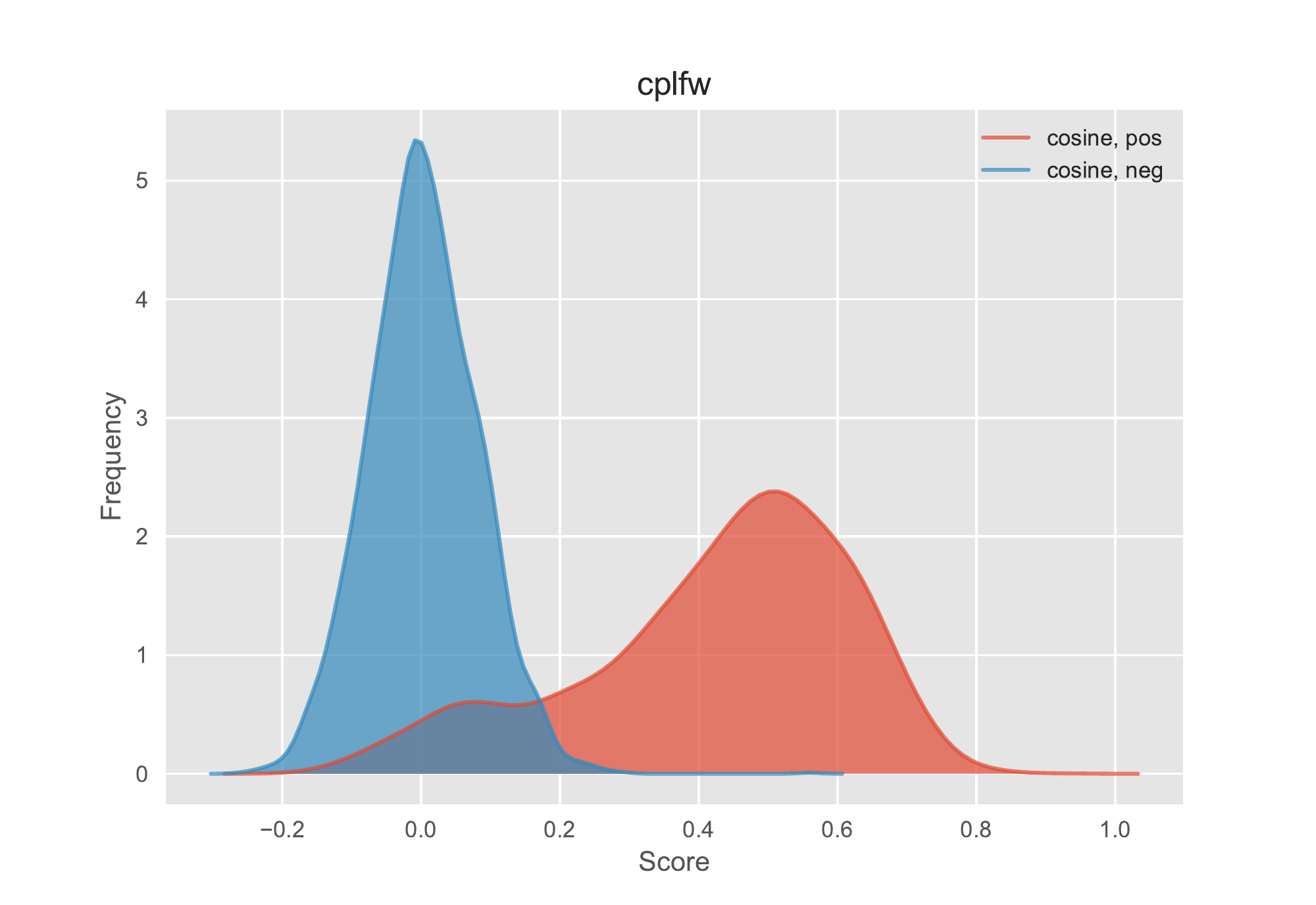}}\hfill
	\subfloat[MLS of PFE (92.80\%)]{%
		 \includegraphics[width=.33\textwidth]{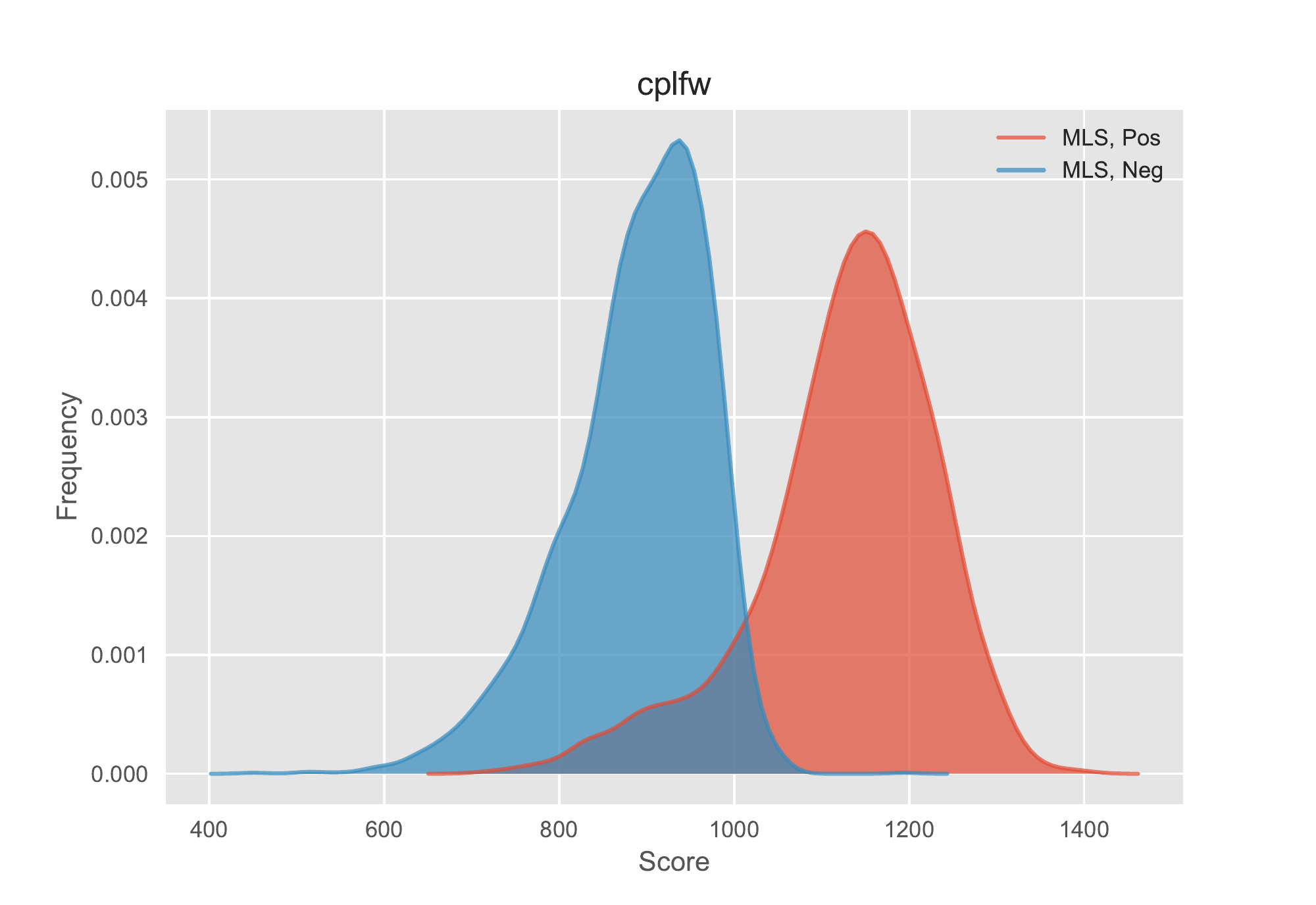}}\hfill
	\subfloat[FastMLS of ProbFace (93.53\%)]{%
		 \includegraphics[width=.33\textwidth]{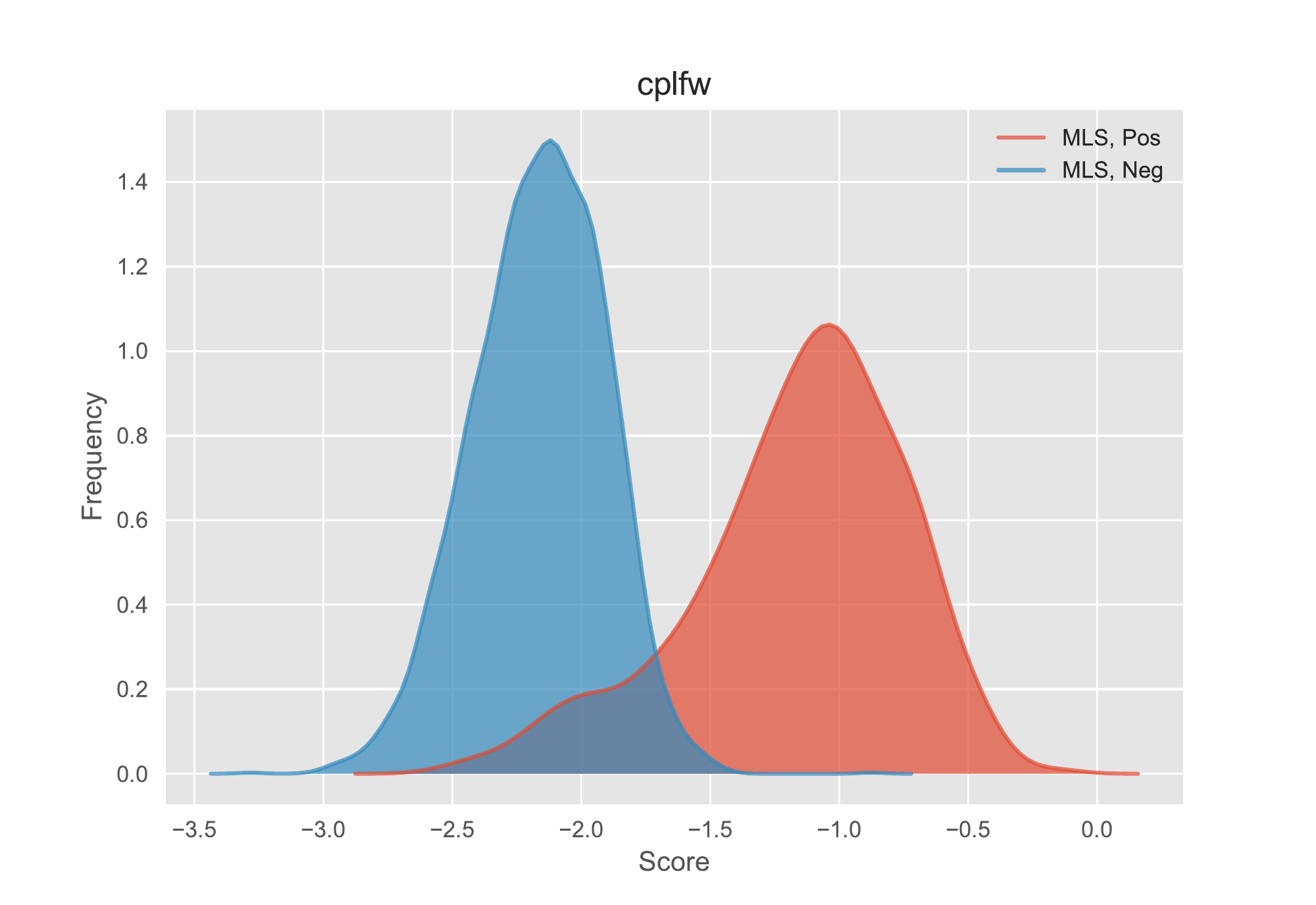}}\hfill
	\caption{~Distribution of the cosine and MLS score on CPLFW datasets.}
	\label{fig:HistMLSDatasetsCPLFW}
\end{figure*}

Figure \ref{fig:HistMLSDatasetsLFW}, Figure \ref{fig:HistMLSDatasetsCALFW} and Figure \ref{fig:HistMLSDatasetsCPLFW} display the distribution of the cosine and MLS score on LFW, CALFW and CPLFW datasets.
Red and blue denote positive matching score and negative matching score respectively.
Compared with PFE, ProbFace can distinguish positive and negative samples more clearly.

\end{document}